%% file: main_fci.tex
\documentclass[a4paper, 11pt]{article}

\usepackage{kotex}

\usepackage[margin=1in]{geometry}
\usepackage{setspace}
\usepackage{microtype}

\usepackage{amsmath} \allowdisplaybreaks
\usepackage{amssymb}
\usepackage{amsthm} \theoremstyle{definition}

\newtheorem{proposition}{Proposition}

\newtheorem{remark}{Remark}

\usepackage[numbers,square,sort&compress]{natbib}
\usepackage{bibentry}

\usepackage{booktabs}
\usepackage{array}

\usepackage{graphicx}
\usepackage{caption}

\newcounter{mycounter}
\usepackage{appendix}

\usepackage{soul} %

\usepackage[hyphens]{url}
\usepackage[bookmarks=false,hidelinks]{hyperref}

\usepackage{subcaption}
\usepackage{multirow}
\usepackage{algorithm}
\usepackage{algpseudocode}

\usepackage{float}
\newfloat{algorithm}{t}{lop}

\algrenewcommand\algorithmicrequire{\textbf{Input:}}
\algrenewcommand\algorithmicensure{\textbf{Output:}}

\algnewcommand\algorithmicforeach{\textbf{for each}}
\algdef{S}[FOR]{ForEach}[1]{\algorithmicforeach\ #1\ \algorithmicdo}

\usepackage{tikz}
\usepackage{pgfplots}
\usetikzlibrary{positioning}
\usetikzlibrary{arrows,shapes}
\usetikzlibrary{snakes}
\usetikzlibrary{calc, backgrounds}

\usepackage{authblk}

\usepackage[shortlabels]{enumitem}

\usepackage[makeroom]{cancel}

\usepackage{adjustbox}

\newcommand{\email}[1]{{\href{mailto:#1}{\nolinkurl{#1}}}}

\usepackage{bm}

\usepackage{xcolor}

\usepackage{etoolbox}

\usepackage[normalem]{ulem}

\usepackage{threeparttable}

\makeatletter
\def\do#1{\@namedef{#1c}{\ensuremath{\mathcal{#1}}}}
\docsvlist{A,B,C,D,E,F,G,H,I,J,K,L,M,N,O,P,Q,R,S,T,U,V,W,X,Y,Z}
\makeatother

\renewcommand{\bar}[1]{\mkern 1.5mu\overline{\mkern-1.5mu#1\mkern-1.5mu}\mkern 1.5mu}

\renewcommand{\tilde}{\widetilde}

\DeclareMathOperator*{\minimize}{minimize}

\renewcommand{\epsilon}{\varepsilon} %

\newif\ifshowauthors
\showauthorstrue

\title{Test-Time Search in Neural Graph Coarsening Procedures for the Capacitated Vehicle Routing Problem}

\ifshowauthors
    \author[a]{Yoonju Sim\thanks{심윤주; \email{syj5268@kaist.ac.kr}}}
    \author[b]{Hyeonah Kim\thanks{김현아; \email{hyeonah.kim@unist.ac.kr}}}
    \author[a,c]{Changhyun Kwon\thanks{권창현; Corresponding author: \email{chkwon@kaist.ac.kr}}}
    \affil[a]{Department of Industrial and Systems Engineering, KAIST, Daejeon, 34141, Republic of Korea}
    \affil[b]{Ulsan National Institute of Science and Technology (UNIST), Ulsan, 44919, Republic of Korea}
    \affil[c]{Omelet, Inc., Daejeon, 34051, Republic of Korea}
    \date{October 20, 2026}
\else
    \author{Anonymous Author(s)}
    \date{}
\fi

\begin{document}
\maketitle

\begin{abstract}
The identification of valid inequalities, such as the rounded capacity inequalities (RCIs), is a key component of cutting plane methods for the Capacitated Vehicle Routing Problem (CVRP). While a deep learning-based separation method can learn to find high-quality cuts, our analysis shows that the model produces fewer cuts than expected and that its predictions vary only slightly with its input features, which limits the diversity of its candidate subsets. This paper proposes an alternative: enhancing the performance of a trained model at inference time through a novel test-time search with stochasticity. First, we introduce stochastic edge selection into the graph coarsening procedure, replacing the previously proposed greedy approach. Second, we propose the Graph Coarsening History-based Partitioning (GraphCHiP) algorithm, which reuses the coarsening history to identify additional RCIs and, without retraining, to construct candidate partitions for generalized capacity inequalities, in particular framed capacity inequalities (FCIs). On randomly generated CVRP instances, our approach reduces the root-node optimality gap of the existing neural separator at all sizes and outperforms the widely used CVRPSEP library on instances with 400 or more customers, though not on smaller ones. GraphCHiP also finds violated generalized capacity inequalities beyond those found by CVRPSEP on a few benchmark instances.

\end{abstract}

\section{Introduction}
\label{sec:introduction}

The Capacitated Vehicle Routing Problem (CVRP) has been a critical combinatorial optimization problem that drives the advances of the modern large-scale optimization algorithms for solving a variety of vehicle routing problems. 
Exact methods such as branch-and-cut (BC) and branch-and-price-and-cut (BPC) algorithms systematically explore the solution space to identify optimal solutions.
These methods build upon the branch-and-bound framework, whose effectiveness is fundamentally determined by the strength of bounds achieved at each node of the search tree.
To enhance bound quality, BC algorithms integrate the cutting plane method with the branching process.
The current state-of-the-art approach, BPC algorithms, extends this framework by incorporating column generation techniques \citep{pecin2017improved, costa2019exact, you2025routeopt}.
The \emph{cutting plane method} plays a crucial role in these algorithms by iteratively refining the relaxation of the problem through the addition of valid inequalities, known as \emph{cuts}.
Each cut reduces the relaxed solution space, eliminating the current fractional solution while preserving all feasible integer solutions.

There are various types of cuts for CVRP, such as rounded capacity inequalities (RCIs), framed capacity inequalities (FCIs), strengthened comb inequalities, and so on.
RCIs are the most widely used capacity inequalities in CVRP.
These inequalities consider the demand of a single subset of customers to determine the minimum number of vehicles required to serve that subset.
Additionally, FCIs extend RCIs by considering the demands of multiple subsets simultaneously, resulting in stronger inequalities.
Identifying high-quality cuts is essential for the cutting plane method to be effective, which is achieved by solving the \emph{separation problem}.
Solving the separation problems for RCIs and FCIs is known to be $\mathcal{NP}$-hard \citep{diarrassouba2017complexity, semet2014chapter}. 

Traditionally, RCI identification is approached in two ways: exact algorithms and heuristic methods.
Exact separation algorithms guarantee optimal cuts with the largest violation but incur substantial computational overhead; thus, they are hard to use for large-scale problems.
A heuristic method, CVRPSEP \citep{cvrpsep}, finds violated cuts quickly, without guaranteeing the most violated ones.
For FCI separation, heuristic methods such as CVRPSEP are also employed \citep{augerat1995computational, fukasawa2006robust}.
Most existing BC and BPC algorithms, if not all, for CVRP and other routing problems utilize CVRPSEP to generate cuts.
It is known that RCIs identified by CVRPSEP, when integrated within the BC framework, produce a dual bound that is as good as the bound by the exact RCI identification, for problem instances with fewer than about 150 customers \citep{Wenger2003, fukasawa2006robust}, despite the fact that the violations of CVRPSEP's cuts are less than those of exact cuts.

However, it may not be the case for large-scale problems.
\citet{kim2024neural} devised a deep learning method, called NeuralSEP, that trains a graph neural network from the exact separation algorithm and found that NeuralSEP outperforms CVRPSEP for problem sizes between 500 and 1,000 customers.
The core idea of NeuralSEP is to learn a separation strategy directly from data using \emph{neural graph coarsening procedures}, a process that involves a sequence of node merging steps guided by predicted probabilities from a graph neural network (GNN) \citep{gilmer2017neural}.
We notice significant opportunities for new separation algorithms that are suitable for very large-scale problems, especially for deep learning approaches.

\begin{figure}
    \centering
    \includegraphics[width=0.8\textwidth]{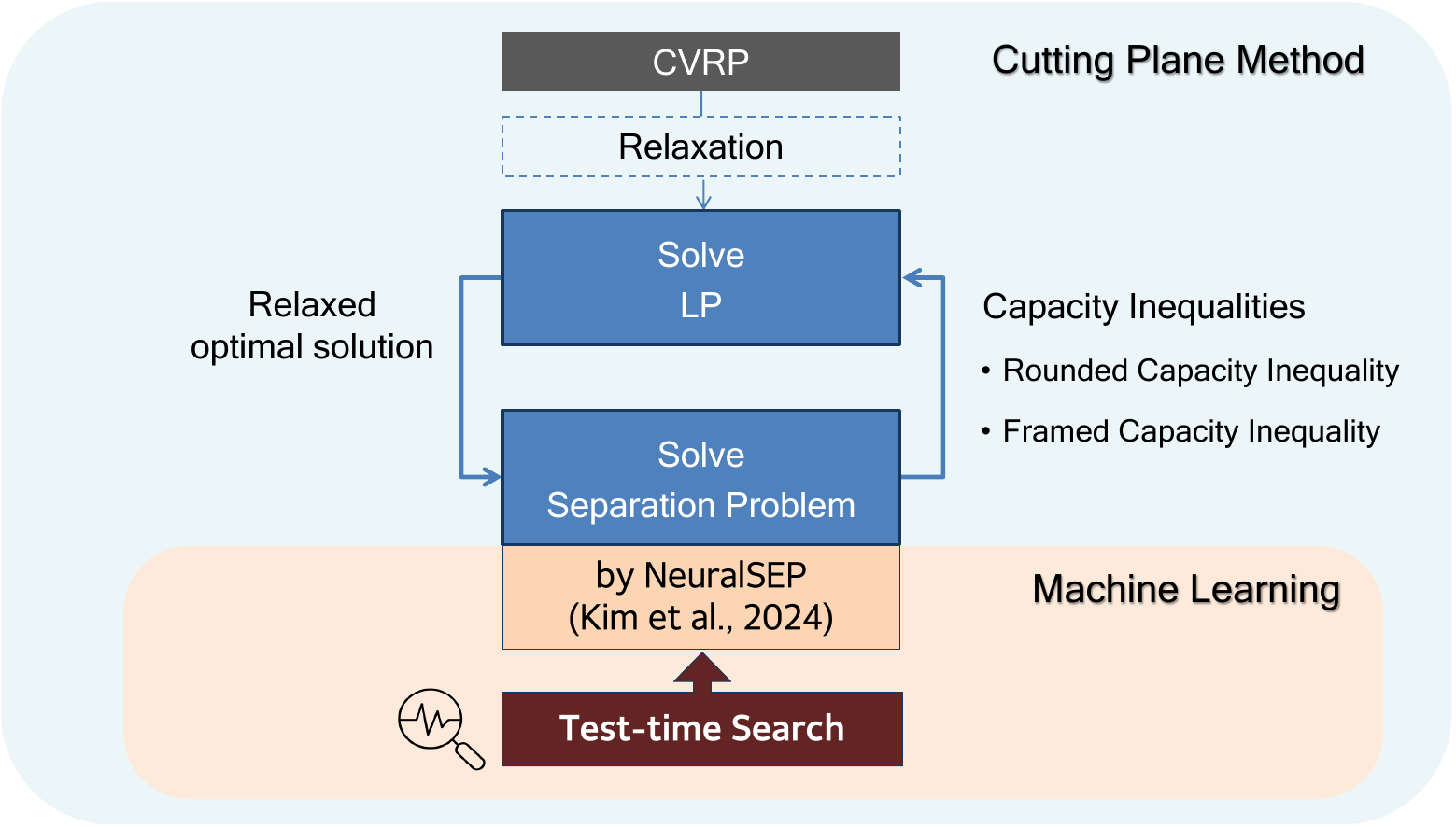}
    \caption{A hybrid approach for CVRP using cutting planes with NeuralSEP and test-time search.
    The cutting plane method alternates between solving the current LP relaxation and employing NeuralSEP as a separation algorithm to generate valid cuts.
    Our proposed test-time search technique enhances NeuralSEP's performance.}
    \label{fig:cuttingplane}
\end{figure}

In this paper, we propose to enhance the output of any neural graph coarsening procedure through a \emph{test-time search} technique.
Test-time search improves a trained model’s performance during inference without retraining \citep{snell2024scaling,NEURIPS2023_91edff07, yao2023tree-of-thoughts}.
We adapt this idea to NeuralSEP within a cutting plane method for CVRP, as illustrated in Figure~\ref{fig:cuttingplane}. 
The starting point is the observation that NeuralSEP, despite learning from the exact separation algorithm, identifies fewer cuts than expected. 
Our analysis suggests that a major contributing factor is the model's tendency to generate a limited variety of candidate subsets—the pool of subsets awaiting violation checking for cut generation.
To address this, we inject stochasticity into its coarsening procedure at test time, and we further propose the Graph Coarsening History-based Partitioning (GraphCHiP) algorithm, which exploits coarsening history to generate additional RCI candidates and candidate partitions for FCIs.
We observe that the coarsening history forms a laminar family, i.e., a nested hierarchy of customer groups, from which GraphCHiP draws both RCI and FCI candidates.
Root-node experiments show that these approaches improve NeuralSEP and, on instances with 400 or more customers, also outperform CVRPSEP.

We emphasize that the proposed test-time search method is general enough to be applied to other learning-based approaches that may be developed in the future, not only NeuralSEP.
The technique of \emph{graph shrinking} has been a key ingredient for heuristic methods for identifying various cutting planes \citep{AUGERAT1998546, ralphs2003capacitated, cvrpsep}.
Since graph coarsening can be interpreted as a probabilistic graph shrinking procedure, we expect that graph coarsening will remain a key ingredient in future learning-based methods.

The remainder of this paper is organized as follows. 
Section~\ref{sec:preliminaries} defines the problem and introduces the necessary background, including the CVRP formulation, RCIs and FCIs, and the learning-based methods for VRPs.
Section~\ref{sec:revisit} provides a recap of NeuralSEP and analyzes its limitations.
Section~\ref{sec:tts} presents our proposed test-time search method, which first employs stochastic edge selection and then utilizes the GraphCHiP algorithm to generate both RCIs and FCIs.
Finally, Section~\ref{sec:experiments} presents computational experiments evaluating RCI separation across various instances, showing increased cut identification and reduced average optimality gap.
For FCI, we report results on instances where our method identifies such cuts.

\section{Preliminaries and Background}
\label{sec:preliminaries}

In this section, we provide the definitions, formulations, and other background information necessary to introduce our problem and discuss our methodology.
We begin by defining the Capacitated Vehicle Routing Problem (CVRP), followed by a presentation of both the formulation and separation algorithms for rounded and framed capacity inequalities (RCIs and FCIs). 
Finally, we discuss learning-based methods for VRPs and the concept of test-time search to provide relevant context for our methodology.

\subsection{Capacitated Vehicle Routing Problem} 
The Capacitated Vehicle Routing Problem (CVRP) is a classical vehicle routing problem where a fleet of vehicles with fixed capacity must serve a set of customers with known demands while minimizing the total travel cost. 
The problem can be defined on an undirected complete graph $G = (V, E)$, where the set of vertices $V$ consists of a depot (vertex $0$) and customer vertices ($V_C = V \setminus \{0\}$), and $E$ is the set of edges.
Each customer $i \in V_C$ has demand $q_i$, each edge $(i,j) \in E$ has a non-negative cost $c_{ij}$, and all vehicles have identical capacity $Q$.
Following \citet{cvrpsep}, we define the edge variable $x_{ij}$ as the number of times edge $(i,j)$ is traversed by the vehicles, and let $\delta(S)$ denote the set of edges with exactly one endpoint in $S$. 
The two-index formulation of CVRP is written as follows:
\begin{align}
    \minimize \quad & \sum_{(i, j) \in E} c_{ij} x_{ij}, \label{cvrp_obj} \\
    \text{subject to} \quad & x(\delta ( \{i \})) = 2 && \forall i \in V_C, \label{cvrp_c1} \\
     & x(\delta(S)) \geq 2 r(S) && \forall S \subseteq V_C, |S| \geq 2, \label{cvrp_c2} \\
     & x_{ij} \in \{0, 1\} && \forall (i,j) \in E,\ i, j \in V_C, \label{cvrp_c3} \\
     & x_{0j} \in \{0, 1, 2\} && \forall j \in V_C, \label{cvrp_c4}
\end{align}
where $r(S)$ denotes the minimum number of vehicles required to cover all demand in a customer set $S$.
The degree equations \eqref{cvrp_c1} ensure that each customer is visited exactly once.
The capacity inequalities \eqref{cvrp_c2} require that the number of vehicles entering any customer set $S$ must meet the minimum needed to serve all demand within that set. 
This minimum value equals the optimal solution of the bin-packing problem, which is $\mathcal{NP}$-hard. 
This computational complexity has motivated the development of various capacity-based cutting planes, including rounded capacity inequalities, framed capacity inequalities, and strengthened comb inequalities.
For vertices other than the depot, variables $x_{ij}$ are binary, as shown in \eqref{cvrp_c3}.
Variables associated with the depot, $x_{0j}$ in \eqref{cvrp_c4}, can take values up to two, as each vehicle must both depart from and return to the depot.

When applying the cutting plane method to solve this problem, we begin by solving the LP relaxation of the CVRP. 
This relaxation is obtained by removing the exponentially many capacity constraints \eqref{cvrp_c2} and converting the integrality constraints \eqref{cvrp_c3} and \eqref{cvrp_c4} to continuous variables.
Next, we identify violated capacity inequalities by applying separation algorithms. 
These violated inequalities, commonly referred to as cuts, are then added to the current LP relaxation before re-solving the problem. This process continues iteratively until no further violated inequalities can be found.
If fractional solutions still exist after completing the cutting plane procedure, we employ a branch-and-bound method to obtain an optimal integer solution.

\subsection{Capacity Inequalities}

We introduce two valid inequalities that involve the vehicle capacity: rounded capacity inequalities and framed capacity inequalities.
Computational methods for solving the separation problems for the two valid inequalities are also discussed.

\subsubsection{Rounded Capacity Inequalities (RCIs)}
Rounded capacity inequalities (RCIs) are the most widely used capacity inequalities in the CVRP.
The inequalities become RCIs when $r(S)$ is calculated as $\left\lceil \sum_{i \in S}q_i/Q \right\rceil$.
Thus, the form of RCIs is given by:
\begin{align}
    x(\delta(S)) \geq 2 \left\lceil \frac{\sum_{i \in S} q_i}{Q} \right\rceil && \forall S \subseteq V_C, |S| \geq 2. \label{rci}
\end{align}
The separation problem for RCIs involves finding the subset $S$ that violates the inequality for the current relaxed solution, which is non-trivial. 
This can be addressed using either heuristic or exact algorithms.
Heuristic algorithms such as CVRPSEP \citep{cvrpsep} quickly find possible RCIs, but they do not guarantee finding the most violated constraints.
Our focus here is on the exact separation algorithms for RCIs, which guarantee finding the most violated constraints.
This approach provides the basis for the learning-based method. %

\paragraph{Exact Separation Algorithm for RCIs}
We can find the most violated RCI by solving an exact separation problem proposed by \citet{fukasawa2006robust}.
From a relaxed solution $\bar{x}$ (which can have fractional values), we construct a support graph as $\bar{G}=(V, \bar{E})$ where $\bar{E} = \{(i, j) \in E:\bar{x}_{ij} > 0\}$.
We then introduce binary variables $y_i$ for every vertex $i \in V$, and continuous variables $w_{ij}$ for every edge $(i, j) \in \bar{E}$.
A value of $y_i = 1$ indicates that vertex $i$ is included in set $S$, 
while $w_{ij} = 1$ indicates that edge $(i, j)$ is in the boundary of set $S$, i.e., $(i, j) \in \delta(S)$.
To find the minimum weight of cuts for a given $\bar{x}$, we solve the following MILP formulation 
for each $m \in \{0, 1, \ldots, \lceil \sum_{i\in V_C} q_i / Q \rceil - 1 \}$:
\begin{align}
    z(m)\ = \
     \min \quad & \sum_{(i, j) \in \bar{E}} \bar{x}_{ij} w_{ij} \label{exact_rci_obj} \\
    \text{s.t.} \quad & w_{ij} \geq y_i - y_j && \forall (i, j) \in \bar{E}, \label{exact_rci_c1} \\
     & w_{ij} \geq y_j - y_i && \forall (i, j) \in \bar{E}, \label{exact_rci_c2} \\
     & \sum_{i \in V_C} q_i y_i \geq (m \cdot Q) + 1, \label{exact_rci_c3} \\
     & y_0 = 0, \label{exact_rci_c4} \\
     & y_i \in \{0, 1\} && \forall i \in V_C, \label{exact_rci_c5} \\
     & w_{ij} \geq 0 && \forall (i, j) \in \bar{E}. \label{exact_rci_c6}
\end{align}
The parameter $m$ controls the number of vehicles required, and the constraint \eqref{exact_rci_c3} ensures that the total demand of selected vertices requires at least $m+1$ vehicles.
For a single support graph, this MILP is solved up to $K = \lceil \sum_{i\in V_C} q_i / Q \rceil$ times, where $K$ is a lower bound on the number of vehicles required to cover all customers.
Thus, we can find the most violated RCI for each $m \in \{0, \ldots, K-1\}$.
If the optimal objective value satisfies $z(m) < 2 (m+1)$, then the violated RCI is found.
Later, another exact separation formulation is proposed by \citet{pavlikov2024exact}.
Since the exact separation problem is $\mathcal{NP}$-hard \citep{diarrassouba2017complexity}, solving it for large-scale problems is computationally expensive.
This is where the learning-based separation algorithm, NeuralSEP \citep{kim2024neural}, comes into play.
This method utilizes a neural network to learn an effective separation strategy directly from data.
The framework of NeuralSEP is described in detail in Section \ref{sec:neuralsep}.

\subsubsection{Framed Capacity Inequalities (FCIs)}
So far, we have focused on RCIs, which are one class of capacity inequalities.
There are other classes of capacity inequalities, such as framed capacity inequalities, strengthened comb inequalities, and so on.
In this subsection, we examine FCIs in detail. 
While RCIs provide valuable cuts, tighter bounds can be obtained by considering finer structure.
One such approach is the framed capacity inequalities (FCIs) introduced by \citet{augerat1995computational}.
FCIs define a structure composed of a larger customer subset $H$ partitioned into multiple smaller, mutually disjoint components.
The total demand of each component is treated as an individual item in a bin packing problem, and the minimum number of vehicles is determined collectively.
To illustrate, consider a customer subset with total demand 200 and capacity $Q=100$.
RCIs compute $\lceil 200/100 \rceil = 2$ vehicles, but if this subset is partitioned into components with demands (90, 90, 5, 15), solving the bin packing problem yields three vehicles, providing a tighter bound.

The formulation of FCIs is as follows. Let $\Omega = \{ S_i : i \in I \}$ be a partition of a subset $H \subseteq V_C$, and let $d(S) = \sum_{j \in S} q_j$ denote the total demand of a set $S$.
We define $r(\Omega)$ as the optimal value of a bin-packing problem with capacity $Q$, in which each $S_i$ with $d(S_i) \le Q$ forms a single item of weight $d(S_i)$, and each $S_i$ with $d(S_i) > Q$ is split into $\lfloor d(S_i)/Q \rfloor$ items of weight $Q$ and one item of weight $d(S_i) \bmod Q$.
Then the inequality is formulated as:
\begin{align}
    x(\delta(H)) + \sum_{i \in I} x(\delta(S_i)) &\geq 2r(\Omega) + 2 \sum_{i \in I} \left\lceil \frac{d(S_i)}{Q} \right\rceil \label{fci}.
\end{align}
Note that if $H$ equals $V_C$, then $x(\delta(H)) = x(\delta(\{0\}))$ equals twice the number of routes, and the corresponding inequality is referred to as the \emph{generalized capacity inequality}.
The identification of these FCIs is computationally expensive, since it requires solving the bin-packing problem, which is known to be $\mathcal{NP}$-hard.

FCIs are incorporated in several algorithmic frameworks. \citet{fukasawa2006robust} integrated FCIs into their branch-and-price-and-cut framework for CVRP, 
and \citet{letchford2007branch} incorporated them in their branch-and-cut approach for the open CVRP, which is a variant where vehicles do not return to the depot. 
Both works employ the separation heuristics from CVRPSEP.

\paragraph{Heuristic Separation Algorithms for FCIs} 
The separation of FCIs is addressed by two heuristic approaches. 
The first, proposed by \citet{augerat1995computational}, is a greedy heuristic.
The algorithm iteratively merges subsets of a partition, starting from subsets with a coboundary value of 2.
In each iteration, it performs a conditional check for violations by solving bin-packing problems or proceeds to merge the most strongly connected pair of subsets, as determined by the $\bar{x}$ values.
The second approach, implemented in CVRPSEP \citep{cvrpsep}, employs a depth-first tree search procedure to explore more potential partitions.
Each tree node encodes a distinct partition of the customer set.
The algorithm evaluates each resulting partition by solving a bin-packing problem to check for violations of FCIs. %
The greedy heuristic can be viewed as a simplified version of this tree search, exploring just one path of the search tree.
Both heuristic approaches employ FCIs when RCI violations become insignificant, indicating that FCIs function as auxiliary cuts.
Building on these works, we suggest a learning-assisted approach that generates candidate partitions for generalized capacity inequalities in Section~\ref{sec:graphchip}.
To the best of our knowledge, neural coarsening histories have not been used for this purpose before.

\subsection{Learning-based Methods for VRPs}

Applying ML to combinatorial optimization has emerged as a powerful paradigm for solving $\mathcal{NP}$-hard problems \citep{bengio2021machine}.
Vehicle Routing Problems (VRPs), such as the Traveling Salesman Problem (TSP) and CVRP, have been a central focus within this field.
Several surveys provide comprehensive overviews of learning-based methods for VRPs \citep{li2022overview, bogyrbayeva2024machine, ZHOU2025104278}.
Following the common taxonomy, we categorize these methods into two main streams: construction-based and improvement-based.
We will briefly review these categories and then introduce the concept of test-time search.

Construction-based approaches, often referred to as end-to-end approaches, directly construct solutions for routing problems without relying on handcrafted heuristics.
These methods typically construct solutions sequentially, with models trained via reinforcement learning (RL) or supervised learning to produce routes step by step \citep{bello2016neural, nazari2018reinforcement, kool2018attention, kwon2020pomo, kim2022sym, luo2023neural}. Others instead predict a probabilistic distribution over edges (e.g., a heatmap), from which solutions are decoded through sampling or search procedures \citep{joshi2019efficient, kool2022deep, dimes-mcts}. 
Another category, improvement-based approaches, starts with a complete solution and iteratively refines it.
Here, an ML model learns to apply powerful heuristic operators, such as 2-opt \citep{d2020learning}, decomposition \citep{li2021deletate,ouyang2025learning}, or Large Neighborhood Search \citep{hottung2019neural}, with some models achieving state-of-the-art results.

A growing line of research integrates ML into exact methods such as branch-and-cut and branch-and-price-and-cut algorithms.
For example, ML can be used to improve column generation by predicting promising columns \citep{morabit2021machine, zhang2022learning, xu2025enhancing}, or to learn branching strategies \citep{cappart2021combining}.
Especially, for the CVRP, \citet{you2023two} explored learning policies for branching decisions, and \citet{kim2024neural} developed a learning-based separation algorithm for RCIs, named NeuralSEP.

\paragraph{Test-time Search}
Test-time search improves the solution quality of a trained model during inference without retraining.
Common approaches include sampling, often referred to as best-of-$N$ \citep{bello2016neural, kool2018attention}, in which multiple solutions are generated and the best one is selected, and tree search methods such as beam search \citep{tsp-gen, choo2022simulation} and Monte Carlo tree search \citep{dimes-mcts}.
A variety of (meta-)heuristic search methods have also been developed, including Random Re-Construct \citep{luo2023neural}, ant colony optimization-inspired methods \citep{ye2023deepaco, pmlr-v258-kim25a}, and genetic algorithm-inspired methods such as Neural Genetic Search \citep{kim2025neural}.
Building on these ideas, we develop a test-time search technique that improves the cut generation of NeuralSEP to strengthen the dual bound.

\section{Revisiting NeuralSEP}
\label{sec:revisit}
This section provides a detailed examination of NeuralSEP, a learning-based separation algorithm for RCIs that serves as the foundation for our methodology. 
We first present the core framework of the algorithm, followed by a sensitivity analysis to investigate its behavior and identify its current limitations.

\subsection{The Design of NeuralSEP}
\label{sec:neuralsep}
\begin{figure}
    \centering
    \includegraphics[width=\textwidth]{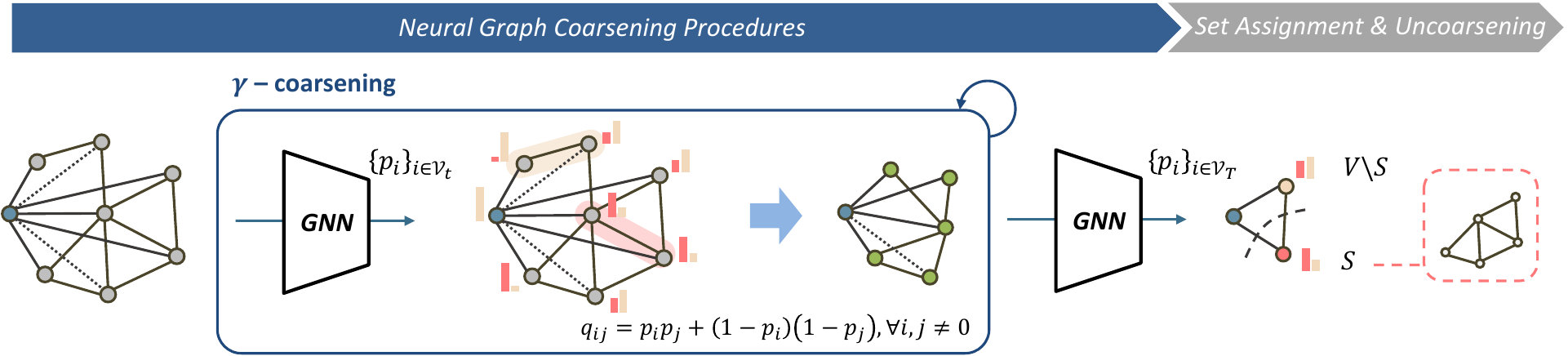}
    \caption{The details of the NeuralSEP framework.  $p_i$ is the predicted probability that vertex $i \in V_C$ is included in the subset $S$. $q_{ij}$ is the contraction probability that vertices $i \in V_C$ and $j \in V_C$ are contracted into a single vertex.}
    \label{fig:neuralsep}
\end{figure}
NeuralSEP, proposed by \citet{kim2024neural}, is a neural separation algorithm designed to generate RCIs for the CVRP.
In essence, NeuralSEP is a learned function that maps an input support graph to an output customer subset, $S$, which defines a violated inequality. 
The parameters of this function, embodied by a neural network, are trained in a supervised manner using labels derived from optimal solutions to the exact separation problem.
The overall framework of NeuralSEP consists of the following three key components, as illustrated in Figure~\ref{fig:neuralsep}.
\begin{enumerate}
    \item \textbf{Graph Embedding with Message Passing GNN}:
    NeuralSEP employs a GNN with a message passing scheme to learn relationships between vertices and computes the vertex selection probability $p_i$, the likelihood that vertex $i$ belongs to a violated subset.
    \item \textbf{Neural Graph Coarsening}: 
    The algorithm iteratively simplifies the graph by contracting vertices.
    This is guided by a contraction probability $q_{ij}$, calculated for pairs of nodes based on their individual selection probabilities $p_i$ and $p_j$, for a pair of nodes $i$ and $j$.
    The algorithm merges the pair of nodes with the highest $q_{ij}$ value first.
    This process, termed $\gamma$-coarsening, continues until a target graph size determined by a coarsening ratio $\gamma$ is reached, then repredicts probabilities for the coarsened graph.
    The overall process terminates when either three vertices remain or all contraction probabilities become zero.
    \item \textbf{Set Assignment and Graph Uncoarsening}: 
    The final subset $S$ is constructed by first making a decision on the coarsest graph based on the final probabilities, then projecting the selected vertices back to the original graph.
\end{enumerate}
The detailed design of GNN is presented in Appendix~\ref{sec:gnn}.
As illustrated in Figure~\ref{fig:neuralsep}, a key part is the \emph{neural graph coarsening} procedure, inspired by the shrinking procedure of several heuristics \citep{AUGERAT1998546,ralphs2003capacitated, cvrpsep}.
It works by iteratively merging vertices that the model predicts are likely to be on the same side of the cut—that is, either both inside the violated subset $S$ or both outside of it.
Note that this procedure is distinct from standard graph coarsening methods, such as \citet{loukas2018spectrally} and \citet{cai2021graph}.
Although trained only on instances with 50 to 100 customers, NeuralSEP successfully generalizes to large-scale problems, outperforming CVRPSEP on instances with more than 400 customers under a fixed iteration.
Further evaluation on benchmark datasets demonstrates scalability and effective generalization to out-of-distribution problems.

\subsection{The Sensitivity Analysis of NeuralSEP}
\label{sec:problem_neuralsep}
Although NeuralSEP performs effectively on large-scale CVRP instances, we observe that it finds substantially fewer valid inequalities than the exact method when solving the same separation problems.
In this section, we analyze a likely source of this gap.

\begin{figure}
    \centering
    \includegraphics[width=\textwidth]{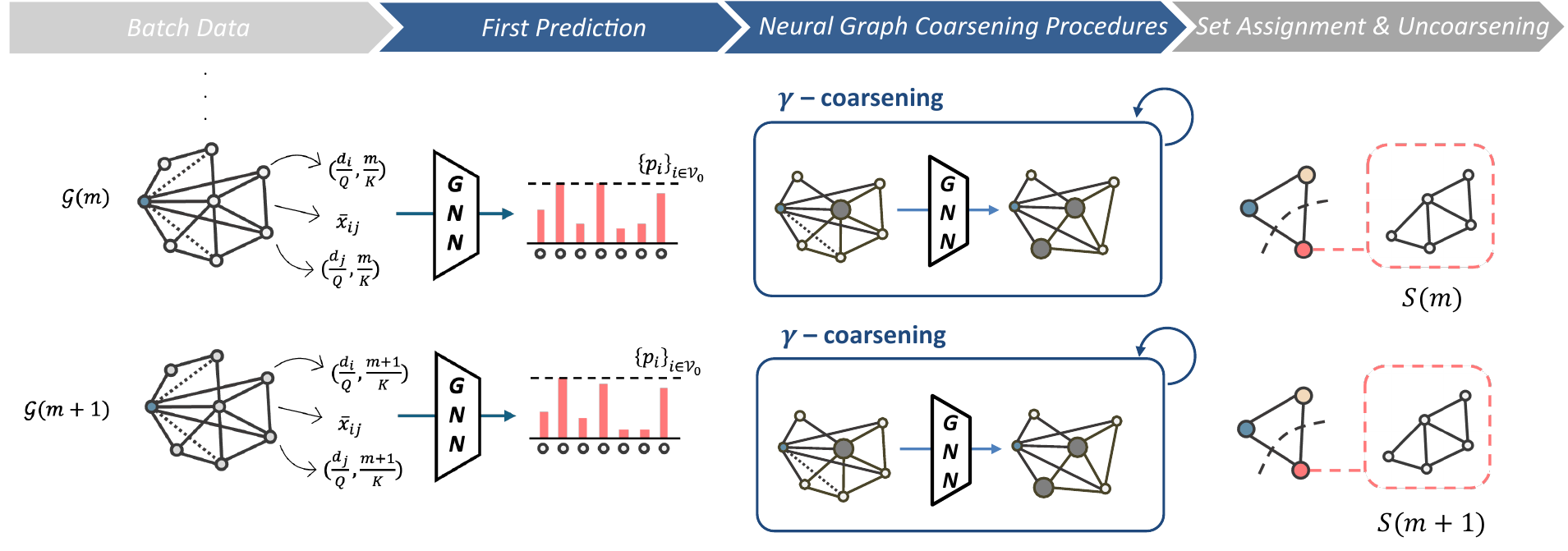}
    \caption{The comparison of the adjacent graphs in the NeuralSEP framework.}
    \label{fig:first_prediction}
\end{figure}

NeuralSEP is trained using the outputs of the exact separation algorithm as labels.
These labels depend on the integer parameter $m \in \{0, \ldots, K-1\}$ in \eqref{exact_rci_c3}, so the exact method can produce up to $K$ different subsets, and hence up to $K$ valid inequalities, for a single support graph.
As illustrated in Figure~\ref{fig:first_prediction}, NeuralSEP incorporates $m$ only through the vertex features: for each $m$, the feature of vertex $i \in V$ is $h_i(m) = (q_i/Q,\ m/K)$, whereas the edge feature $h_{ij} = \bar{x}_{ij}$ is the same for all $m$.
The resulting $K$ graphs are processed in parallel and differ only in the second vertex feature.

\begin{table}
    \centering
    \captionsetup{justification=centering}
    \caption{Comparison of unique valid inequalities found by the exact method and NeuralSEP}
    \label{table:main_sep}
    \input{tables/main_sep}

    \vspace{0.1cm}
\end{table}

Table~\ref{table:main_sep} compares the number of unique valid inequalities found by the two methods on identical separation problems, expressed as a percentage of the total number of problems.
The test set consists of 100 randomly selected graph instances for each size. 
Since one separation problem is solved for each $m$, the total number of problems equals the sum of $K$ over the instances.
NeuralSEP finds considerably fewer valid inequalities than the exact method, and the gap widens as the graph size increases.

To investigate this, we examine how the model responds to the second vertex feature $m/K$.
A closed-form expression of this sensitivity is intractable owing to the nonlinear activation functions as shown in Appendix~\ref{sec:math_analysis}, so we evaluate it numerically.
Let $G^\ell(m)$ denote instance $\ell \in \{1, \ldots, L\}$ with vertex features $h_i(m)$, and let $\Phi_i(G^\ell(m)) = p_i$ be the probability predicted by the model for vertex $i$.

\begin{figure}
    \centering
    \begin{subfigure}[b]{0.45\textwidth}
        \includegraphics[width=\textwidth]{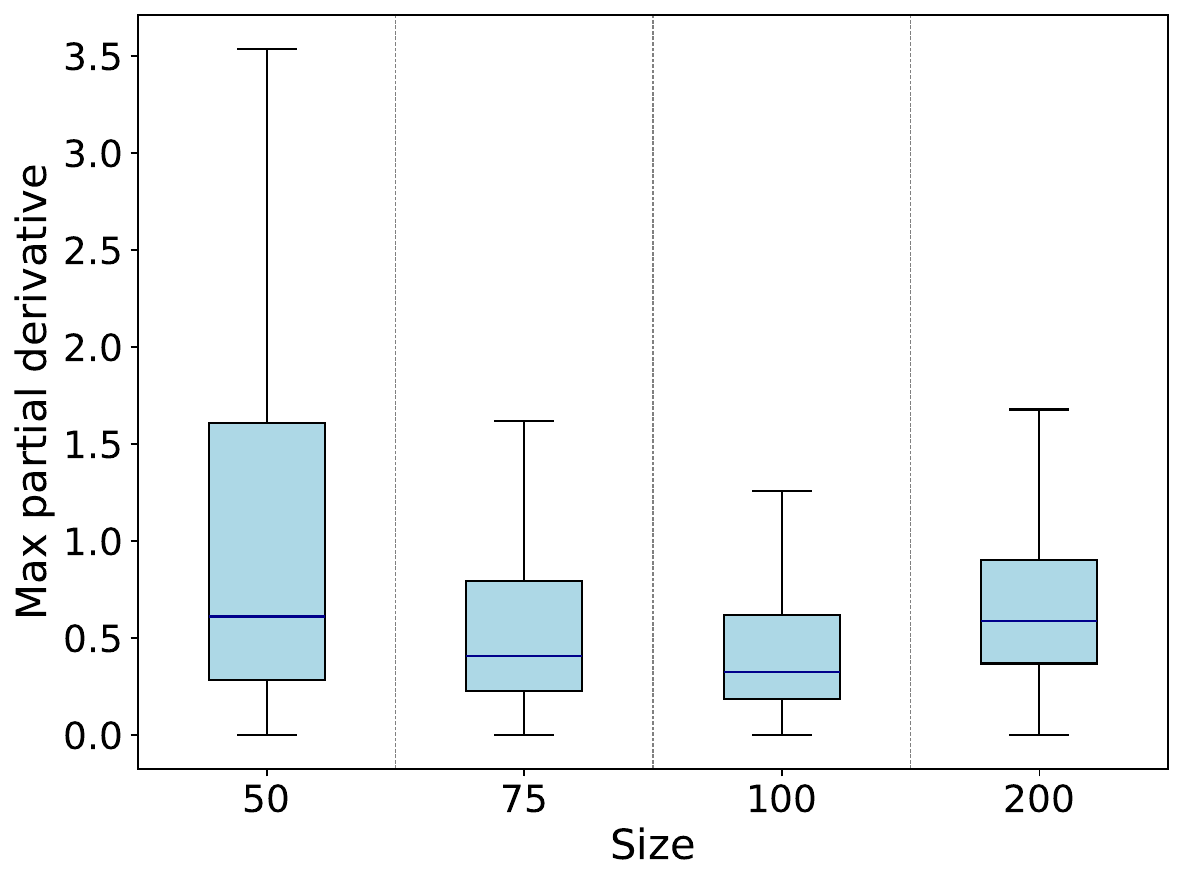}
        \caption{Metric $D_1$: Partial derivative}
        \label{fig:partial_derivative}
    \end{subfigure}
    \hfill
    \begin{subfigure}[b]{0.45\textwidth}
        \includegraphics[width=\textwidth]{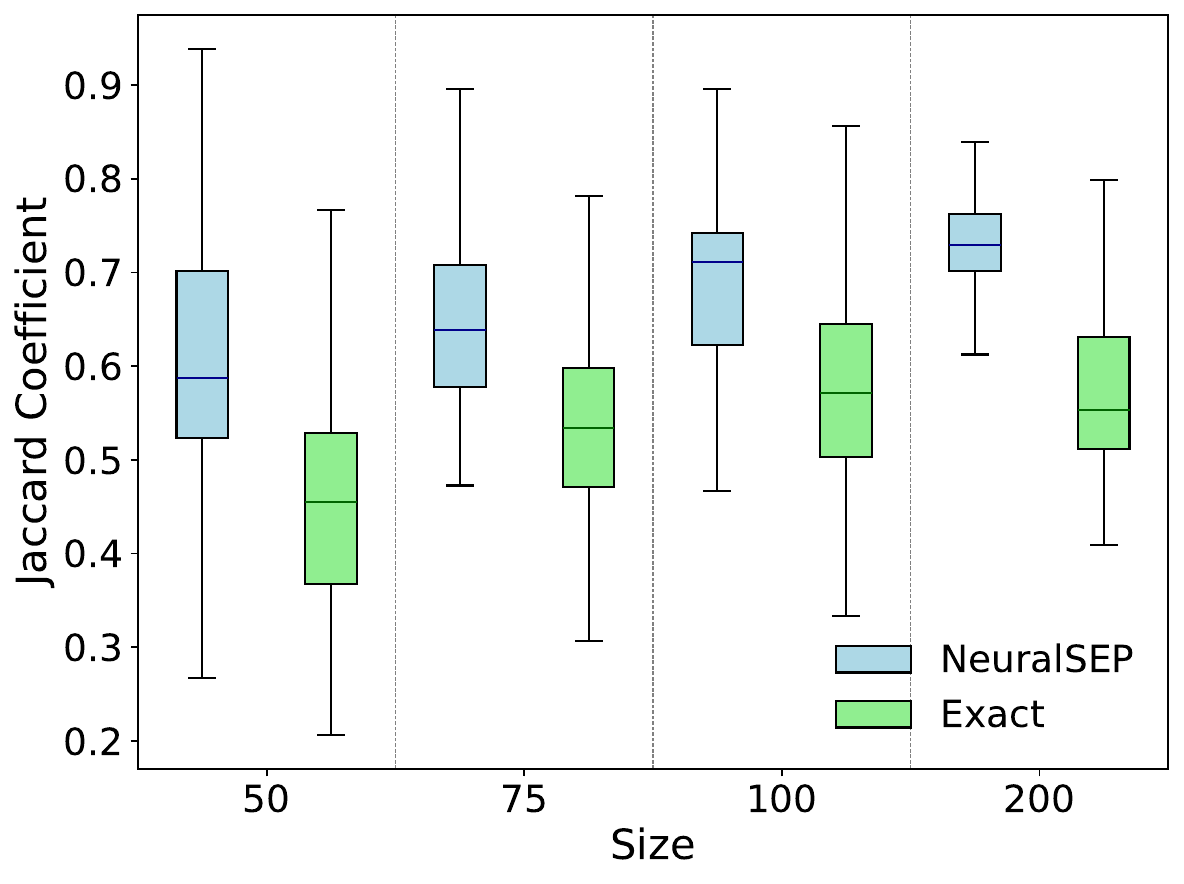}
        \caption{Metric $D_2$: Jaccard coefficient}
        \label{fig:jaccard}
    \end{subfigure}
    \caption{Box plots of the two metrics over graph sizes}
\end{figure}

\paragraph{Sensitivity of the initial prediction.}
We first measure the sensitivity of the initial predicted probabilities, before coarsening, using a finite-difference approximation of the derivative with respect to $m/K$:
\begin{equation}
    D_1 = \max_{i \in V^\ell} \left| \frac{\partial \Phi_i(G^\ell(m))}{\partial (m/K)} \right|
    \approx \max_{i \in V^\ell} \left| \frac{\Phi_i(G^\ell(m) + \epsilon e_2) - \Phi_i(G^\ell(m))}{\epsilon} \right|,
\end{equation}
where $G^\ell(m) + \epsilon e_2$ denotes the graph obtained by adding a small perturbation $\epsilon$ to the second feature $m/K$ of every vertex, and $D_1$ is computed for every $m$ and $\ell$.
As shown in Figure~\ref{fig:partial_derivative}, $D_1$ averages approximately 0.5.
Since consecutive values of $m$ change this feature by only $1/K$, the predicted probability of even the most sensitive vertex changes by only about $0.5/K$, indicating that the predictions vary only marginally with $m$.

\paragraph{Diversity of the final subsets.}
We then examine whether this low sensitivity carries over to the subsets returned after coarsening.
For adjacent values of $m$, we compare the final subsets $S^\ell(m)$ and $S^\ell(m+1)$ using the Jaccard coefficient
\begin{equation}
    D_2 = \frac{|S^\ell(m) \cap S^\ell(m+1)|}{|S^\ell(m) \cup S^\ell(m+1)|}, \quad m \in \{0, \ldots, K-2\},
\end{equation}
where a higher value indicates more similar subsets.
Figure~\ref{fig:jaccard} shows that NeuralSEP yields consistently higher values of $D_2$ than the exact method, i.e., its candidate subsets for different values of $m$ are less diverse.
Both methods are evaluated on all separation problems, including those without violated inequalities, on which NeuralSEP is also trained.
An additional analysis of the similarity between the predicted probabilities for adjacent values of $m$ is reported in Appendix~\ref{sec:cosine}.

In summary, despite being trained to produce different outputs for different values of $m$, the model is not sensitive enough to $m$ to generate the required diversity, and the coarsening procedure does not compensate for this.
This limitation is consistent with, and likely contributes to, the smaller number of valid cuts generated by NeuralSEP in Table~\ref{table:main_sep}, although the lower accuracy of individual predictions may also play a role.
Rather than retraining the model, the next section introduces test-time methods that increase the diversity of candidate subsets for the trained model.

\section{Test-Time Search in Neural Graph Coarsening}
\label{sec:tts}
In this section, we introduce two approaches to improve the performance of NeuralSEP in the neural graph coarsening at test time, without any additional training.
First, in Section~\ref{sec:stochastic}, we propose a stochastic edge selection method designed to enhance the diversity of the pool of candidate subsets.
Then, in Section~\ref{sec:graphchip}, we present a novel approach that discovers more RCIs and additionally identifies FCIs, another type of cut.
Both approaches operate on the coarsening history.
Stochastic edge selection diversifies the history generated by the trained model, and GraphCHiP exploits this history to construct candidate subsets and partitions. 

\subsection{Stochastic Edge Selection in Graph Coarsening}
\label{sec:stochastic}
An effective approach to addressing the challenge is to introduce stochastic elements into the \emph{neural graph coarsening} procedure.
Given that the original NeuralSEP generates highly similar outputs for different values of $m$, incorporating random perturbation to the coarsening process produces an effect similar to sampling multiple solutions from the trained neural network.

\begin{figure}
    \centering
    \includegraphics[width=\textwidth]{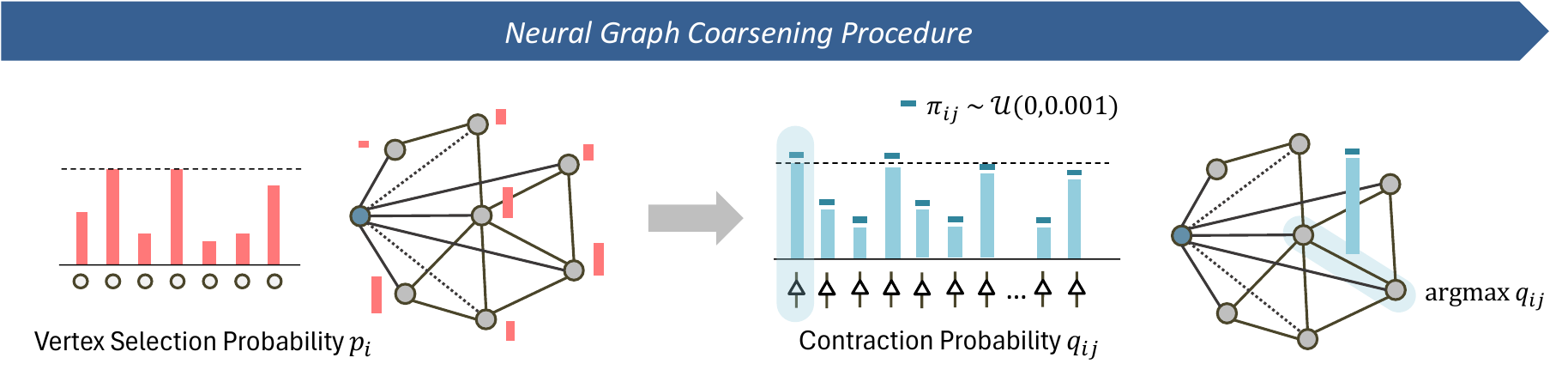}
    \caption{Illustration of $\pi$-greedy selection method in the \emph{neural graph coarsening} procedure}
    \label{fig:stochastic}
\end{figure}

We first present a simple sampling-based strategy to enhance the diversity of the pool of candidate subsets generated by NeuralSEP.
The original contraction probability $q_{ij}$ is defined as the probability of connecting vertices into the same set, i.e., both inside set $S$ or both outside $S$, as follows:
\begin{equation} \label{eq:edge_prob}
    q_{ij} = \begin{cases}
        p_i p_j + (1-p_i)(1-p_j) & \mbox{if } i, j \neq 0 \\
        0 & \mbox{otherwise.}
    \end{cases}
\end{equation}
We modify this probability by introducing a stochastic parameter:
\begin{equation} \label{eq:edge_prob2}
    \tilde{q}_{ij} = \begin{cases}
        p_i p_j + (1-p_i)(1-p_j) + \pi_{ij} & \mbox{if } i, j \neq 0 \\
        0 & \mbox{otherwise.}
    \end{cases}
\end{equation}
where $\pi_{ij}$ is a small random value drawn from a uniform distribution $\Uc(0, 0.001)$.
During the coarsening procedure, an edge $e$ whose perturbed probability $\tilde{q}_{ij}$ is the highest among the edges of the current coarsened graph is selected for contraction, i.e., 
$e = \arg\max_{(i,j) \in \bar{E}'} \, \tilde{q}_{ij}$, where $\bar{E}'$ denotes the edge set of the current coarsened graph and initially equals $\bar{E}$.
This approach is termed \emph{$\pi$-greedy selection}, and Figure~\ref{fig:stochastic} illustrates this process.
Adding randomness can alter which edges are chosen for contraction and create chain effects throughout the procedure.
Since the coarsening procedure iterates and generates predictions several times, early changes in contraction order spread through the entire process. 
As a result, the algorithm produces different cuts despite starting from the same initial conditions.
A more diverse candidate pool can expose additional violated cuts; whether this improves the bound after accounting for the extra computation is evaluated in Section~\ref{sec:pi_perf}.

For comparison, we also consider two alternative methods that sample an edge directly from a distribution over the contraction probabilities: stochastic roulette selection and softmax selection.
The \emph{stochastic roulette selection} method is a weighted random selection method that reflects the original edge importance, i.e., the contraction probabilities \(q_{ij}\).
The method first converts $q_{ij}$ into normalized probabilities \(q_{ij}^{'}\):
\begin{align}
    q_{ij}^{'} = \frac{q_{ij}}{\displaystyle \sum_{(u,v) \in \bar{E}'} q_{uv}}, \quad \forall (i,j) \in \bar{E}'.
\end{align}
An edge is then sampled according to the categorical distribution, 
$\text{Cat}\bigl(\{q_{ij}^{'}\}_{(i,j)\in \bar{E}'}\bigr). $

The other method---the \emph{softmax selection method}---is a more sophisticated version of the stochastic roulette selection.
It applies a softmax function to the contraction probabilities \(q_{ij}\) to obtain a distribution over the edges,
which gives more weight to edges with higher probabilities. Formally,
\begin{align}
    q_{ij}^{''} = \frac{\exp\bigl(q_{ij}/\tau\bigr)}{\displaystyle \sum_{(u,v) \in \bar{E}'} \exp\bigl(q_{uv}/\tau\bigr)}, \quad \forall (i,j) \in \bar{E}'.
\end{align}
where \(\tau\) is a temperature parameter that controls the randomness of the selection.
This edge selection follows $\text{Cat}\bigl(\{q_{ij}^{''}\}_{(i,j)\in \bar{E}'}\bigr).$
Note that, since $\pi_{ij} < 0.001$, $\pi$-greedy selection always chooses an edge whose contraction probability is within $0.001$ of the maximum, and it coincides with greedy selection whenever the largest contraction probability exceeds the second largest by more than $0.001$.
Hence, it diversifies the contraction order only among nearly tied edges while preserving the ranking produced by the trained model.
By contrast, the two sampling methods above may select edges with clearly lower contraction probabilities.

\subsection{GraphCHiP: Graph Coarsening History-based Partitioning Algorithm}
\label{sec:graphchip}
Stochastic edge selection diversifies the subset returned at the end of the coarsening procedure, but the intermediate graphs generated along the way are discarded.
Each intermediate graph groups the customers into supernodes, which provide further candidate subsets for RCIs and, taken together, candidate partitions for FCIs that NeuralSEP cannot produce since it returns a single subset.
We therefore propose GraphCHiP, which traces the coarsening history backward to construct candidate \emph{subsets} and \emph{partitions} for violated RCIs and FCIs.
The coarsening procedure is controlled by the coarsening ratio $\gamma$, which specifies the ratio of vertices retained at each step.
For a given support graph $\bar{G} = (V, \bar{E})$, the coarsening process generates a sequence of graphs $\{\mathcal{G}_t\}_{t=0}^{T}$, where $\mathcal{G}_0 = \bar{G}$ and $\mathcal{G}_T$ is the coarsest graph.
Each coarsening step $t \in \{0, \ldots, T-1\}$ merges vertices in $V_t$ into disjoint supernodes to form the next graph $\mathcal{G}_{t+1}$ with vertex set $V_{t+1}$.
Accordingly, each supernode is described by \emph{node maps} $\{\mathcal{M}_t\}_{t=1}^{T}$, where $\mathcal{M}_t: V_t \rightarrow 2^{V}$ maps each supernode $u \in V_t$ to the set of original vertices aggregated into it.
Here, $0$ denotes the depot, which is never contracted since $q_{0j} = 0$.
Let $S$ denote the subset returned by NeuralSEP from this coarsening sequence.
Since $S$ is obtained by assigning the supernodes of the coarsest graph $\mathcal{G}_T$ and projecting them back to the original graph, $S$ is a union of sets $\mathcal{M}_T(u)$, $u \in V_T$.
For each level $t \in \{1, \ldots, T\}$, let $S_t = \{u \in V_t : \mathcal{M}_t(u) \cap S \neq \emptyset\}$ and define
\begin{equation}\label{eq:omega_t}
    \Omega_t = \{\mathcal{M}_t(u) : u \in S_t\} \cup \big\{\{v\} : v \in V_C \setminus S\big\}.
\end{equation}
We write $\Omega \preceq \Omega'$ if every block of $\Omega$ is contained in some block of $\Omega'$, i.e., $\Omega$ is a refinement of $\Omega'$.
The following remark shows that the node maps record a hierarchical clustering of the original vertices and characterizes the candidates generated from it.
\begin{remark}\label{rem:structure}
For a coarsening sequence $\{\mathcal{G}_t\}_{t=0}^{T}$ and the subset $S$ obtained from it, the following hold:
\begin{enumerate}[(i)]
    \item For every $t$ and every $u \in V_t \setminus \{0\}$, either $\mathcal{M}_t(u) \subseteq S$ or $\mathcal{M}_t(u) \cap S = \emptyset$. Hence, $\{\mathcal{M}_t(u) : u \in S_t\}$ is a partition of $S$.
    \item The family $\mathcal{L} = \{\mathcal{M}_t(u) : t \in \{1, \ldots, T\},\ u \in V_t \setminus \{0\}\}$ is laminar, i.e., any two of its members are either disjoint or nested.
    \item Each $\Omega_t$ is a partition of $V_C$, and $\Omega_1 \preceq \Omega_2 \preceq \cdots \preceq \Omega_T$.
\end{enumerate}
\end{remark}

\begin{figure}
    \centering
    \includegraphics[width=\textwidth]{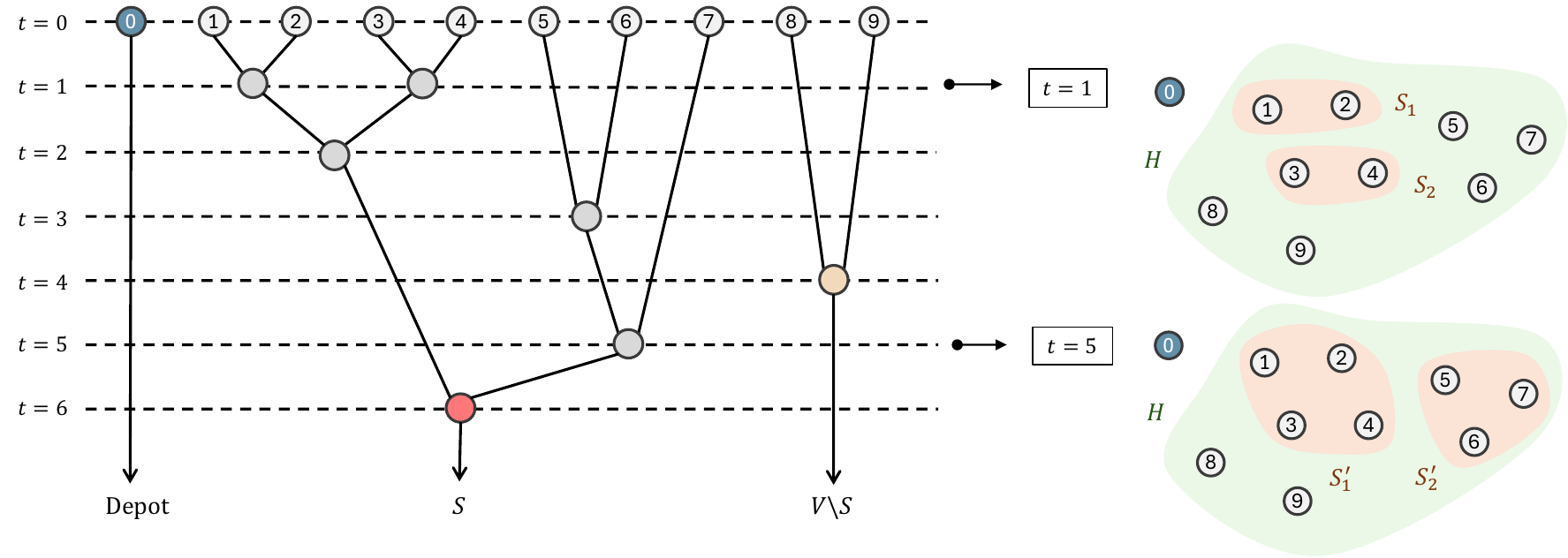}
    \caption{Illustration of how the GraphCHiP algorithm makes the subsets and partitions from the node map.
    The left diagram shows the neural graph coarsening process from $t=0$ (original graph with vertices 0-9) to $t=6$ (final coarsened graph with three supernodes: Depot, $S$, and $V_C \setminus S$). 
    The right diagrams demonstrate partition generation at steps $t=1$ and $t=5$ by backtracking the coarsening history. 
    At $t=5$, merged vertex groups (e.g., $S_1' = \{1,2,3,4\}$, $S_2' =\{5,6,7\}$) and individual vertices form partition $\Omega$ of $V_C$.
    }
    \label{fig:fci}
\end{figure}

All three statements follow from the fact that each coarsening step merges whole supernodes, as detailed in Appendix~\ref{sec:proof-1}.
Figure~\ref{fig:fci} illustrates these candidates, and Remark~\ref{rem:structure} makes the connection between the two types of cuts precise.
At each level $t$, the blocks $\mathcal{M}_t(u)$, $u \in S_t$, serve as candidate subsets for RCIs when examined individually, and the same blocks, completed by singletons, form the candidate partition $\Omega_t$ for FCIs when examined jointly.
Since any two sets in $\mathcal{L}$ are either disjoint or nested, $\mathcal{L}$ can be viewed as a collection of trees in which each set is the parent of the sets merged into it, and both searches move along these trees.
Thus, the RCI and FCI searches of GraphCHiP are two views of a single hierarchy, and the candidate partitions form one chain in the lattice of partitions of $V_C$.
This places GraphCHiP between the two FCI heuristics reviewed in Section~\ref{sec:preliminaries}: the greedy heuristic of \citet{augerat1995computational} follows a single chain generated by merging the most strongly connected pair of subsets, and the tree search of CVRPSEP explores a tree of such chains, whereas GraphCHiP evaluates the chain selected by the learned coarsening procedure.

\subsubsection{GraphCHiP for RCI Separation}
GraphCHiP performs a backward search over the coarsening hierarchy, starting from the coarsest level. 
At each level $t$, the set of original vertices of each supernode contained in $S$ is checked for RCI violation.
To prevent redundant cut discovery and improve computational efficiency, this search is performed only when the original subset $S$ satisfies the RCI (i.e., no violation).
Once any RCI-violating subset is found, the search terminates immediately. 
The pseudo-code is given in Algorithm~\ref{algo:graphchip_rci} in Appendix~\ref{sec:algo_rci}, and the number of checks is bounded as follows, with the proof in Appendix~\ref{sec:proof-2}.
\begin{proposition}\label{prop:rci}
The worst-case number of RCI checks performed by GraphCHiP is bounded by $\frac{\gamma}{1-\gamma}|V|$, where $\gamma$ is the coarsening ratio.
\end{proposition}
With $\gamma$ set to 0.75 as in our experiments, the number of checks is at most 3$|V|$.
In practice, far fewer checks are required because the search terminates as soon as a violating subset is found, so GraphCHiP does not become a bottleneck within the cutting plane method.

\subsubsection{GraphCHiP for FCI Separation}
For FCIs, since enumerating all partitions is intractable, GraphCHiP evaluates only the candidate partition $\Omega_t$ in \eqref{eq:omega_t} at each level $t$.
Hence, $H = V_C$, and the identified inequalities are generalized capacity inequalities, a subclass of FCIs.
To focus on cuts arising from the partition structure, partitions in which a block already violates an RCI are discarded.
For the remaining partitions, $r(\Omega)$ is replaced by $L_2(\Omega)$, the lower bound of \citet{martello1990knapsack}, which is computed in $O(n \log n)$ time for $n$ items and yields a weaker but still valid inequality.
This bound is computed only when $\text{RHS} - \text{LHS} > -2$ under the trivial bound $\lceil d(V_C)/Q \rceil$, since each unit increase of $r(\Omega)$ raises the RHS by two.
This screen is heuristic: a skipped partition can still yield a violated inequality if its $L_2$ bound exceeds the trivial bound by two or more.
The complete procedure is given in Algorithm~\ref{algo:graphchip}, and its worst-case time complexity is given below, with the proof in Appendix~\ref{sec:proof-3}.

\begin{algorithm}
    \caption{GraphCHiP for FCI} \label{algo:graphchip}
    \small
    \begin{algorithmic}[1]
    \Require
        Support Graph $\Gc$, \par
        Subset $S \subseteq V_C$, \par
        A node map $\{\mathcal{M}_t\}_{t=1}^{T}$ \par
    \Ensure A collection $\Fc$ of the partitions $\Omega$ of $V_C$ and their RHS values %
        \State $\Fc \gets \emptyset$, $t \gets T$ \Comment{Initialize. Iterate backwards from the coarsest graph}
        \While{$t > 0$}  
            \State $\Omega \gets \emptyset$  \Comment{Initialize a partition of $V_C$}
            \State $S_t \gets \{ u \in V_t \mid \mathcal{M}_t(u) \cap S \neq \emptyset \}$
            \Comment{Identify relevant supernodes at level $t$}
            \ForEach{$u \in S_t$} 
                \State $h \gets \mathcal{M}_t(u)$ \Comment{Original vertices aggregated into supernode $u$}
                \State Add $h$ to $\Omega$  \Comment{Add the subset to the partition}
            \EndFor            

            \ForEach{$v \in V_C \setminus \bigcup \Omega$}
                \State $h \gets \{v\}$ 
                \State Add $h$ to $\Omega$  \Comment{Add remaining vertices as singletons}
            \EndFor

            \If{$\exists h \in \Omega$ such that $2 \lceil d(h)/Q \rceil > \text{Calculate\_LHS}(\Gc, h)$} 
                \State $t \gets t - 1$   
                \State \textbf{continue} 
                \Comment{Skip if any subset in $\Omega$ violates an RCI}
            \EndIf

            \State LHS $\gets \text{Calculate\_LHS}(\Gc, \Omega)$
            \State RHS $\gets 2\left\lceil\frac{d(V_C)}{Q}\right\rceil + 2\sum_{h \in \Omega} \left\lceil\frac{d(h)}{Q}\right\rceil$
            \If{$\text{RHS} - \text{LHS} > -2$}  \Comment{Check tight RHS for the potential cuts}
                \State $\text{RHS} \gets 2L_2(\Omega) + 2\sum_{h \in \Omega} \left\lceil\frac{d(h)}{Q}\right\rceil$ \Comment{Update RHS with the $L_2$ lower bound of $r(\Omega)$}
                \If{$\text{RHS} - \text{LHS} > 0$}  \Comment{Check if the violation exists}
                    \State Add ($\Omega$, RHS) to $\Fc$ 
                \EndIf
            \EndIf
            \State $t \gets t - 1$  \Comment{Backtrack to the previous coarsening step}
        \EndWhile
        \\
        \Return $\Fc$
    \end{algorithmic}
\end{algorithm}

\begin{proposition}\label{prop:fci}
Given a coarsening history generated with a fixed coarsening ratio $\gamma \in (0,1)$, GraphCHiP for FCIs can be implemented in $O\big(|V|^2 \log |V|\big)$ worst-case time, where $|V|$ is the number of vertices of the original graph.
More precisely, its running time is $O\big((|\bar{E}| + |V| \log |V|) \log |V|\big)$.
\end{proposition}
Owing to the $L_2$ bound, GraphCHiP runs in polynomial time even though the underlying bin-packing problem is $\mathcal{NP}$-hard.

\section{Experiments}
\label{sec:experiments}
We evaluate the proposed methods within a cutting plane method at the root node of a branch-and-cut algorithm, in which they serve as the separation routine.
We first evaluate stochastic edge selection in terms of the optimality gap and the diversity of candidate subsets, and then evaluate GraphCHiP in terms of additional RCIs and its ability to identify violated FCIs on structured benchmark instances, including its combination with CVRPSEP.

For a fair comparison, we run all experiments under the same conditions.
The experiments are conducted on a single machine with an AMD Ryzen 9 5900X CPU and 62 GB DDR4 RAM @ 3200 MT/s.
All separation algorithms, including the learning-based ones, are run on the CPU during the cutting plane experiments.
The software environment is summarized in Table~\ref{tab:experiment_env} in Appendix~\ref{sec:environment}.

\subsection{Experimental Settings for RCI Separation}
\label{sec:setting}
\paragraph{Choice of Graph Neural Network Framework}
The original NeuralSEP is implemented with the Deep Graph Library (DGL) \citep{wang2019deep}. 
Since NeuralSEP relies only on simple graph operations, we migrate it to the lighter-weight PyTorch Geometric (PyG) \citep{fey2019fast}, which substantially reduces the inference time, as reported in Appendix~\ref{sec:dgl_pyg}.

\paragraph{Baseline and Compared Algorithms}
For the RCI separation experiments, we compare five methods that allow us to measure the gain from each of our contributions.
CVRPSEP is the widely used heuristic separation library and serves as the non-learning baseline.
NeuralSEP$_1$ is the original implementation of NeuralSEP in DGL.
NeuralSEP$_2$ is our retrained PyG re-implementation and serves as a controlled baseline.
$\pi$-NeuralSEP$_2$ applies $\pi$-greedy selection to NeuralSEP$_2$, and $\pi$-NeuralSEP$_2$ + GC further applies GraphCHiP and constitutes our full approach.

\paragraph{Metric and Dataset}
Our evaluation methodology follows the approach from \citet{kim2024neural}.
We measure performance using the optimality gap defined as $GAP = \frac{UB - LB}{UB} \times 100(\%)$, 
where $UB$ is the upper bound provided by the hybrid genetic search (HGS) algorithm from \citet{vidal2022hybrid} and $LB$ is the lower bound obtained from the cutting plane method at the root node of a branch-and-cut algorithm.
We conduct our evaluation on the same set of publicly available test instances available at \url{https://github.com/hyeonahkimm/neuralsep}.
These randomly generated CVRP instances were created following the guidelines of \citet{uchoa2017new} and \citet{queiroga202110}.
The test set includes instances with a varying number of customers, $|V_C| \in \{50, 75, 100, 200, 300, 400, 500, 750, 1000\}$.

\subsection{Experimental Results of RCI Separation}
\begin{figure}
    \centering
    \begin{subfigure}[b]{0.45\textwidth}
        \includegraphics[width=\textwidth, height=5cm]{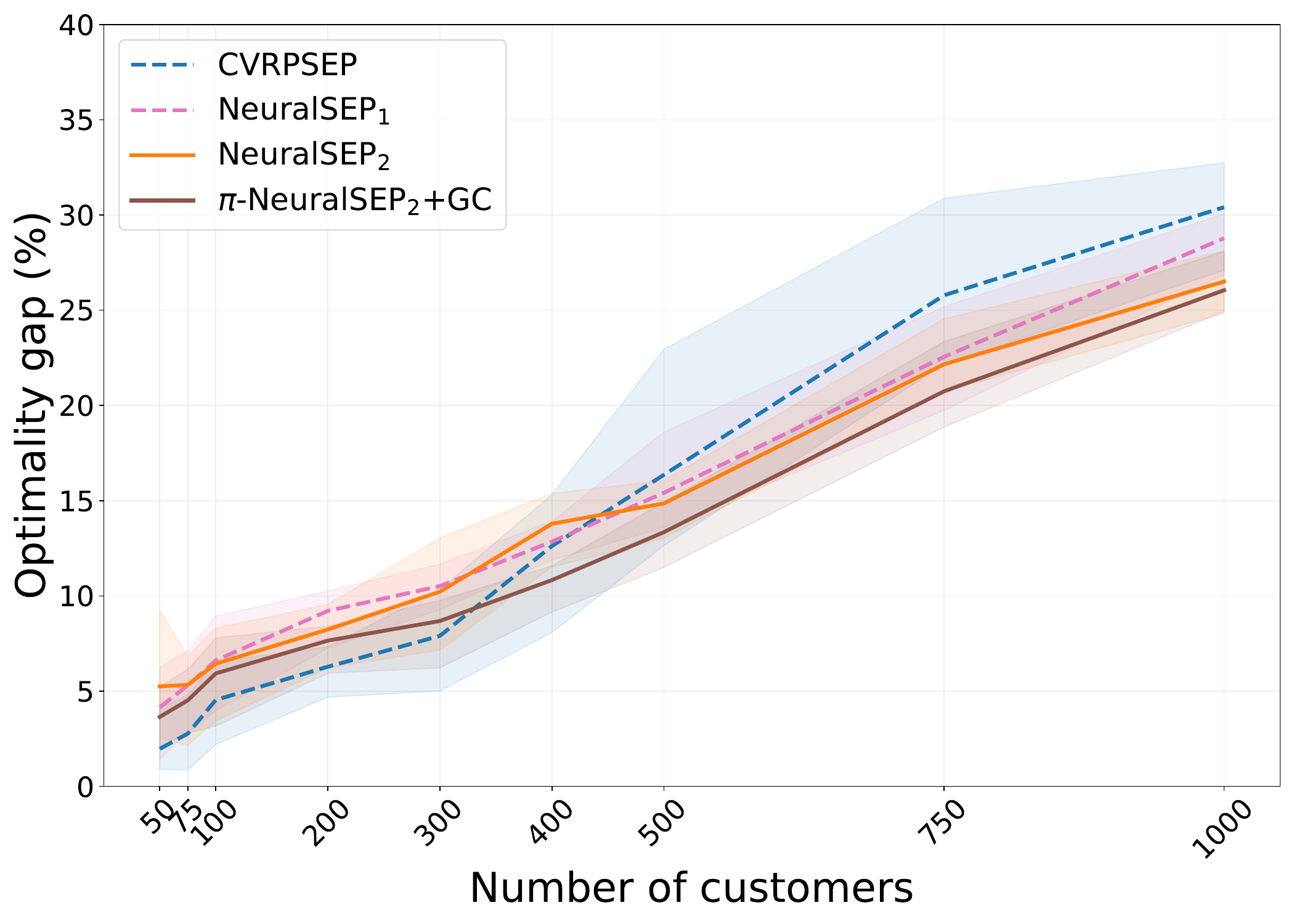}
        \caption{Optimality gap by customer size}
        \label{fig:gap_rci}
    \end{subfigure}
    \hfill
    \begin{subfigure}[b]{0.45\textwidth}
        \includegraphics[width=\textwidth, height=5cm]{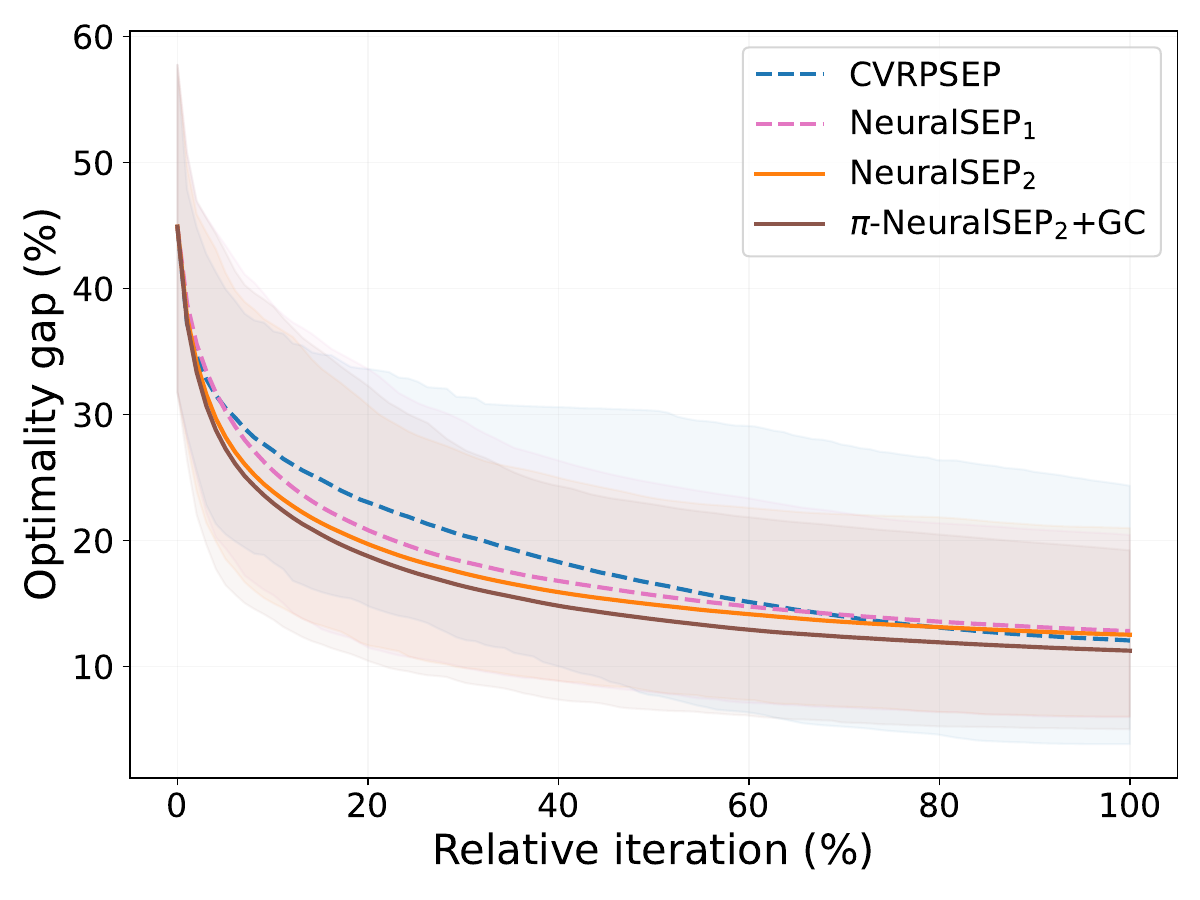}
        \caption{Optimality gap reduction over relative iterations}
        \label{fig:iter_rci}
    \end{subfigure}
    \caption{Comparison of the performance of RCI separation algorithms}
    \label{fig:cvrp_rci}
\end{figure}

We evaluate the performance of the proposed RCI separation methods on the randomly generated test instances under a uniform computational time limit of 3600 seconds for each run.
The performance of the RCI separation algorithms on the test set is summarized in Table~\ref{table:compare_rci} and visually represented in Figure~\ref{fig:cvrp_rci}.
Although the proposed test-time search increases the time per iteration, it still achieves lower optimality gaps than NeuralSEP within the same time limit.
As shown in Figure~\ref{fig:gap_rci}, our complete approach achieves lower optimality gaps than both the DGL and PyG implementations of NeuralSEP across all instance sizes, and outperforms CVRPSEP on instances with 400 or more customers.
Figure~\ref{fig:iter_rci} aggregates all instances and indicates that, despite relatively small margins, the proposed method maintains the most favorable gap reduction trajectory over the course of relative iterations, although this pooled average is dominated by the larger instances.
In the figures, the shaded regions represent instance-level variability, corresponding to the min–max range in Figure~\ref{fig:gap_rci} and the 20–80 percentile range in Figure~\ref{fig:iter_rci}.

\begin{table}
    \captionsetup{justification=centering}
    \caption{Summary of the performance of RCI separation algorithms}
    \label{table:compare_rci}
	\begin{adjustbox}{max width=\textwidth}

\input{tables/compare_rci}

    \end{adjustbox}
    \begin{tablenotes}
        \footnotesize
        \item \textit{Notes.} ``Time/Iter'' denotes the average time (in seconds) spent per separation iteration. All experiments are conducted with a time limit of 3600 seconds.
    \end{tablenotes}
\end{table}

\subsubsection{Performance of $\pi$-greedy Selection Method}
\label{sec:pi_perf}

We first isolate the effect of $\pi$-greedy selection by comparing NeuralSEP$_2$ and $\pi$-NeuralSEP$_2$, which share the same trained model and PyG implementation.
Table~\ref{table:compare_rci_pi} reports the number of RCIs generated by each method, its relative increase, the change in the optimality gap, and the winning ratio.
$\pi$-greedy selection reduces the average optimality gap for all instance sizes, with improvements of more than one percentage point for instances with 50 customers and with 300 to 750 customers.

\begin{table}
    \captionsetup{justification=centering}
    \caption{Comparison of the performance with and without $\pi$-greedy selection}
    \label{table:compare_rci_pi}
	\begin{adjustbox}{max width=\textwidth}
    	\input{tables/compare_rci_pi}

    \end{adjustbox}
    \begin{tablenotes}
        \footnotesize
        \item \textit{Notes.} \% Cut Increase is computed as the relative increase in the number of RCI cuts produced by $\pi$-NeuralSEP$_2$ compared to NeuralSEP$_2$. $\Delta$ Gap is measured at the root node (in percentage points), and $\downarrow$ indicates improvement. The winning ratio denotes the fraction of instances where $\pi$-NeuralSEP$_2$ outperforms NeuralSEP$_2$.
    \end{tablenotes}
\end{table}

\begin{figure}
    \centering
    \includegraphics[width=\textwidth]{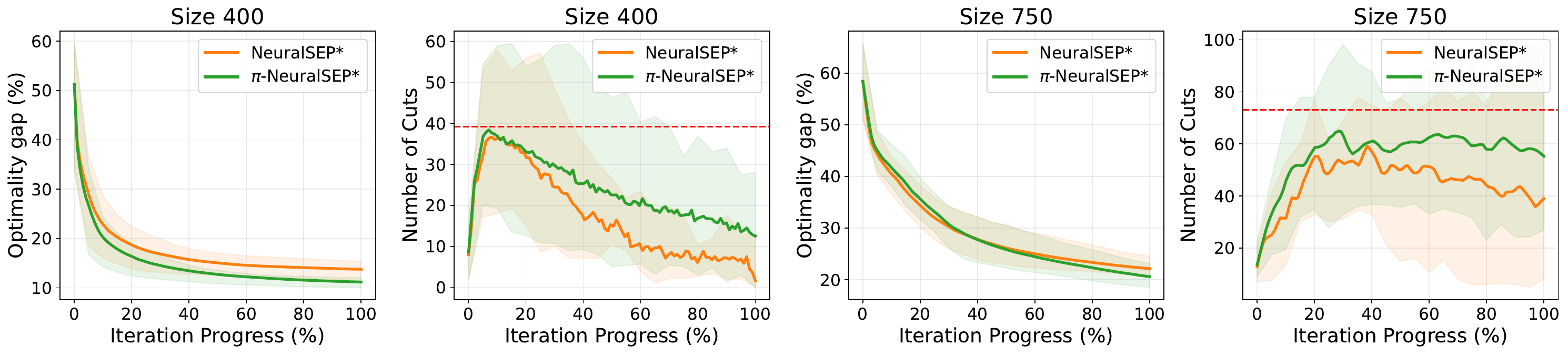}
    \caption{The optimality gap and number of cuts added over relative iterations on instances of size 400 and 750}
    \label{fig:gap_cut_summary_v2}
\end{figure}

To examine the effect of $\pi$-greedy selection on each iteration, Figure~\ref{fig:gap_cut_summary_v2} presents the average trajectories of the optimality gap and the number of RCI cuts over relative iterations for instances with 400 and 750 customers, with full results in Appendix~\ref{sec:pi_results}.
The red dashed horizontal line denotes the average number of vehicles for each instance size.
The $\pi$-greedy selection produces more cuts per iteration, which is consistent with its faster gap reduction.

\begin{figure}
    \centering 
    \includegraphics[width=0.6\textwidth]{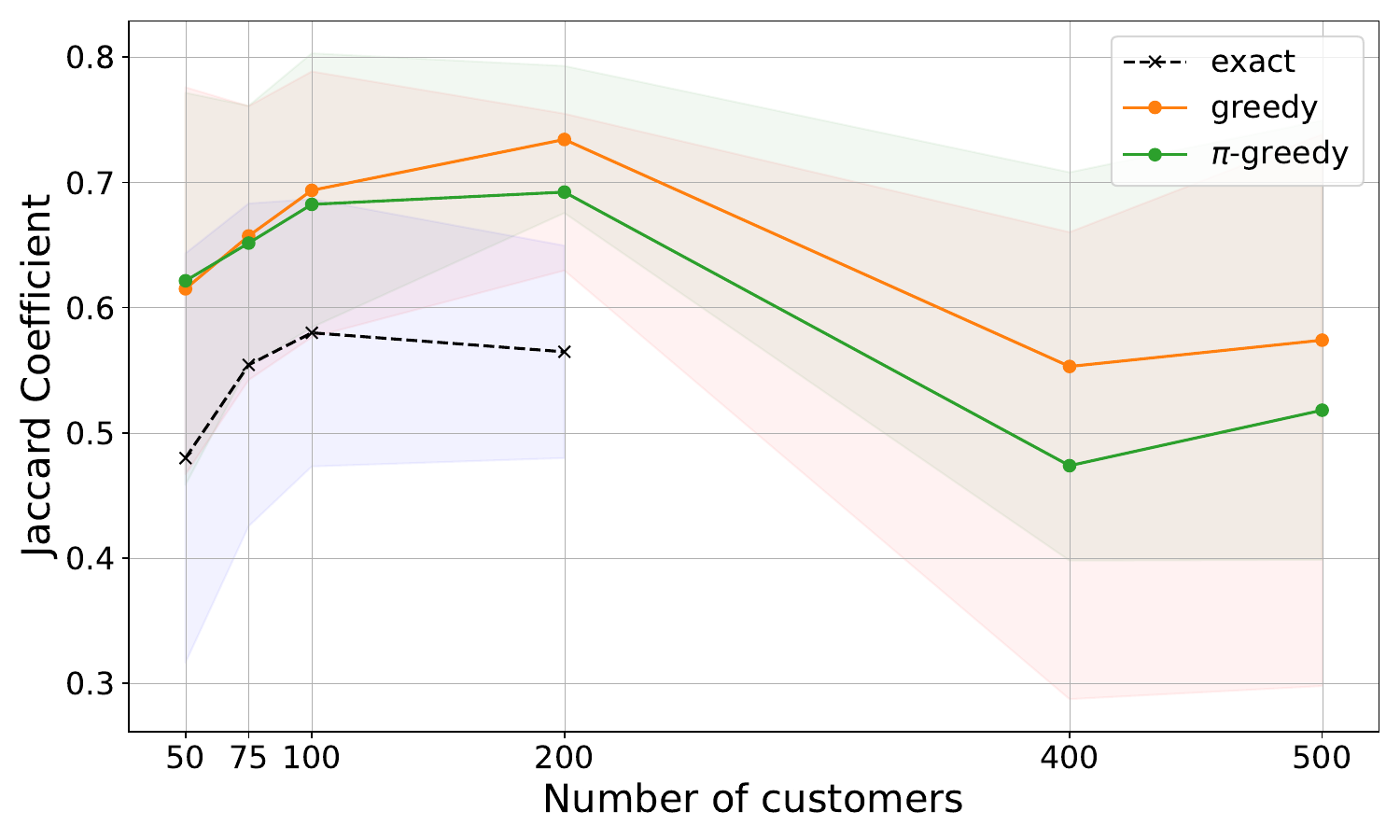}
    \caption{Comparison of Jaccard similarity. The lower the Jaccard coefficient, the more diverse the set of candidate subsets.}
    \label{fig:jaccard2}
\end{figure}

\paragraph{Diversity of Generated Subsets}
To evaluate the effect of $\pi$-greedy selection on the diversity of candidate subsets, we compute the Jaccard coefficient $D_2$ defined in Section~\ref{sec:problem_neuralsep} with and without $\pi$-greedy selection on 100 separation problems for each size between 50 and 500 customers, which extend those of Section~\ref{sec:problem_neuralsep} to 400 and 500 customers.
The exact method is shown only up to 200 customers, since its solutions could not be obtained for larger instances.
As shown in Figure~\ref{fig:jaccard2}, $\pi$-greedy selection yields lower Jaccard coefficients, i.e., more diverse subsets, for instances with 200 or more customers, whereas the difference is small for smaller instances.
This indicates that $\pi$-greedy selection mainly diversifies the candidate subsets for larger instances.

\paragraph{Comparison with Other Stochastic Edge Selection Methods}
We also compare $\pi$-greedy selection with the stochastic roulette and softmax selection methods of Section~\ref{sec:stochastic} on instances with up to 400 customers, using a softmax temperature of $\tau = 0.25$.
As shown in Table~\ref{table:compare_methods}, the differences among the three methods are small, but $\pi$-greedy selection tends to generate more cuts and a slightly lower optimality gap.
A possible reason is that $\pi$-greedy selection perturbs only near-ties and thus preserves the ranking learned by the model, whereas roulette and softmax selection may contract edges with clearly lower $q_{ij}$.

\begin{table}
    \captionsetup{justification=centering}
    \caption{Comparison of stochastic edge selection methods on randomly generated instances}
    \small
    \label{table:compare_methods}
    \input{tables/compare_methods.tex}

\end{table}

\subsubsection{Performance of GraphCHiP Algorithm for RCIs}
\begin{figure}
    \centering
    \begin{subfigure}[b]{0.45\textwidth}
        \includegraphics[width=\textwidth, height=5cm]{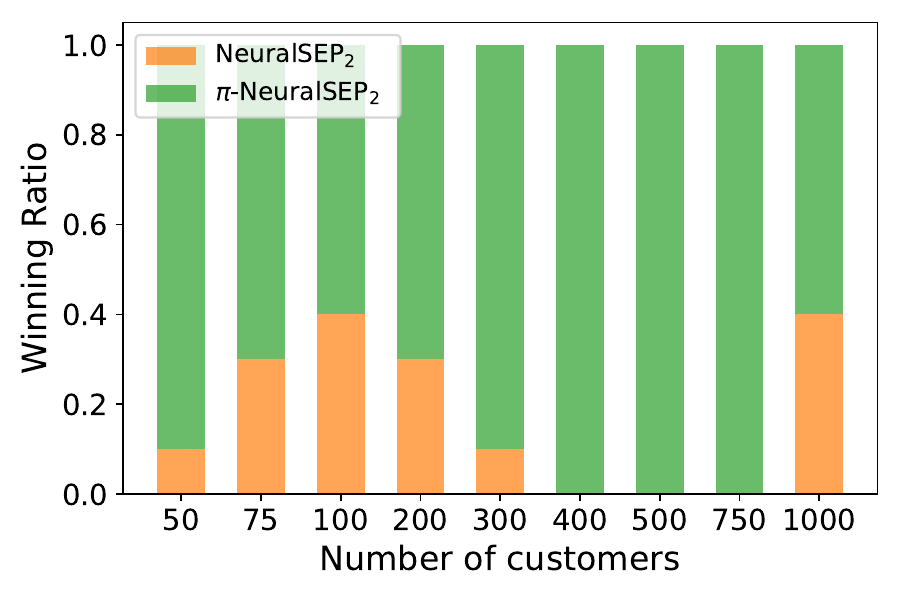}
        \caption{NeuralSEP$_2$ vs. $\pi$-NeuralSEP$_2$}
        \label{fig:pi_rci}
    \end{subfigure}
    \hfill
    \begin{subfigure}[b]{0.45\textwidth}
        \includegraphics[width=\textwidth, height=5cm]{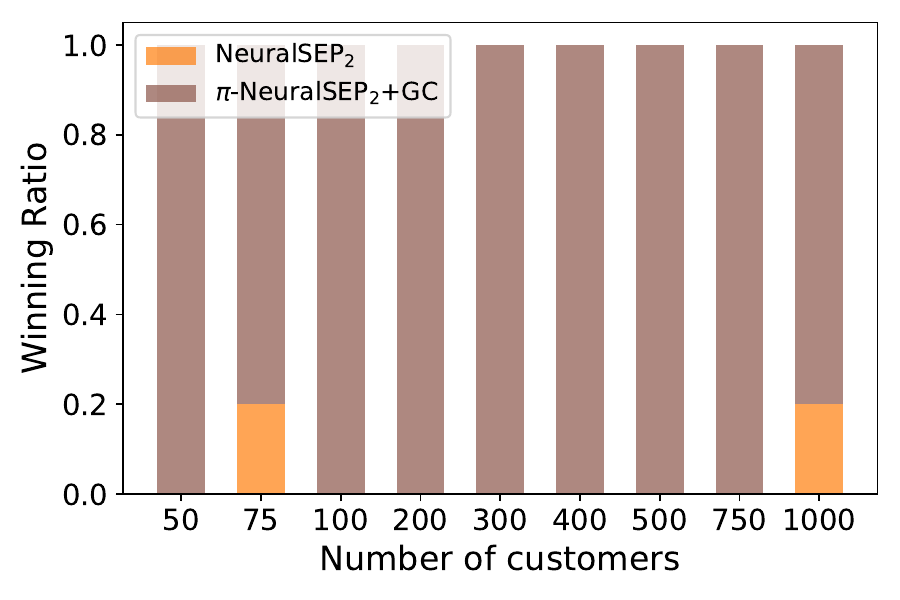}
        \caption{NeuralSEP$_2$ vs. $\pi$-NeuralSEP$_2$+GC}
        \label{fig:graphchip_rci}
    \end{subfigure}
    \caption{The winning ratio out of 10 instances}
    \label{fig:winning_ratio}
\end{figure}

Table~\ref{table:compare_rci_graphchip} shows that GraphCHiP increases the number of RCIs by 2.0--20.7\% and reduces the average optimality gap for all sizes except 750 customers. The improvement is consistent for instances with 75 to 400 customers, with gap reductions of 0.36--0.51 percentage points and winning ratios of 0.8--1.0, and diminishes for larger instances, presumably because the additional time per iteration reduces the number of iterations within the time limit.
Figure~\ref{fig:winning_ratio} shows that adding GraphCHiP further raises the winning ratio against NeuralSEP$_2$, reaching 1.0 for seven of the nine sizes.

\begin{table}
    \captionsetup{justification=centering}
    \caption{Comparison of the performance with the GraphCHiP algorithm for RCIs}
    \label{table:compare_rci_graphchip}
  	\begin{adjustbox}{max width=\textwidth}
    	\input{tables/compare_rci_graphchip}

    \end{adjustbox}
    \begin{tablenotes}
        \footnotesize
        \item \textit{Notes.} ``+GC'' denotes $\pi$-NeuralSEP$_2$ augmented with GraphCHiP. The winning ratio reports how often $\pi$-NeuralSEP$_2$+GC outperforms $\pi$-NeuralSEP$_2$, where a tie is counted as half a win.
    \end{tablenotes}
\end{table}

\subsection{Experimental Results of FCI Separation}
FCI separation is invoked only when the maximum violation found during RCI separation is at most 1.0, which typically occurs in the later stages of the cutting plane method. Therefore, the FCI experiments are conducted without a time limit, and the cutting plane method terminates when no violated inequality is found.

On the randomly generated instances with sizes ranging from 50 to 500, our method does not identify any violated FCIs.
This is not unexpected, since the existence of violated FCIs depends strongly on the demand structure of the instance.
On structured benchmark instances from CVRPLIB, we observe that violated FCIs are identified in five instances.
Table~\ref{table:x_instance} summarizes the performance of FCI separation on CVRPLIB instances with up to 300 customers for which GraphCHiP identifies violated FCIs.
Here, CVRPSEP FCI refers to the FCI separation algorithm implemented in the CVRPSEP package.
Adding GraphCHiP FCI to $\pi$-NeuralSEP$_2$ reduces the gap on all five instances, whereas adding CVRPSEP FCI to CVRPSEP RCI reduces the gap on two instances and finds no violated FCI on two others. When both FCI routines are added to CVRPSEP RCI, the gap is the lowest on every instance, with a tie on X-n233-k16. GraphCHiP finds FCIs on P-n23-k8 and X-n125-k30, where CVRPSEP FCI finds none, and on X-n153-k22 it reduces the gap from 5.82\% to 4.40\% by adding 279 FCIs. A detailed example of an FCI found on this instance is given in Appendix~\ref{sec:case_study}.

\begin{table}
    \captionsetup{justification=centering}
    \caption{Performance of FCI separation on benchmark instances}
    \label{table:x_instance}
    \input{tables/x_instance_fci.tex}
    \begin{tablenotes}
        \footnotesize
        \item \textit{Notes.} For the last block, we report the total number of FCI cuts,
        with the breakdown shown in parentheses as (CVRPSEP FCI + GraphCHiP FCI).
        A dash (--) indicates that FCI separation was not applied.
    \end{tablenotes}
\end{table}

\begin{figure}
    \centering
    \includegraphics[width=\textwidth]{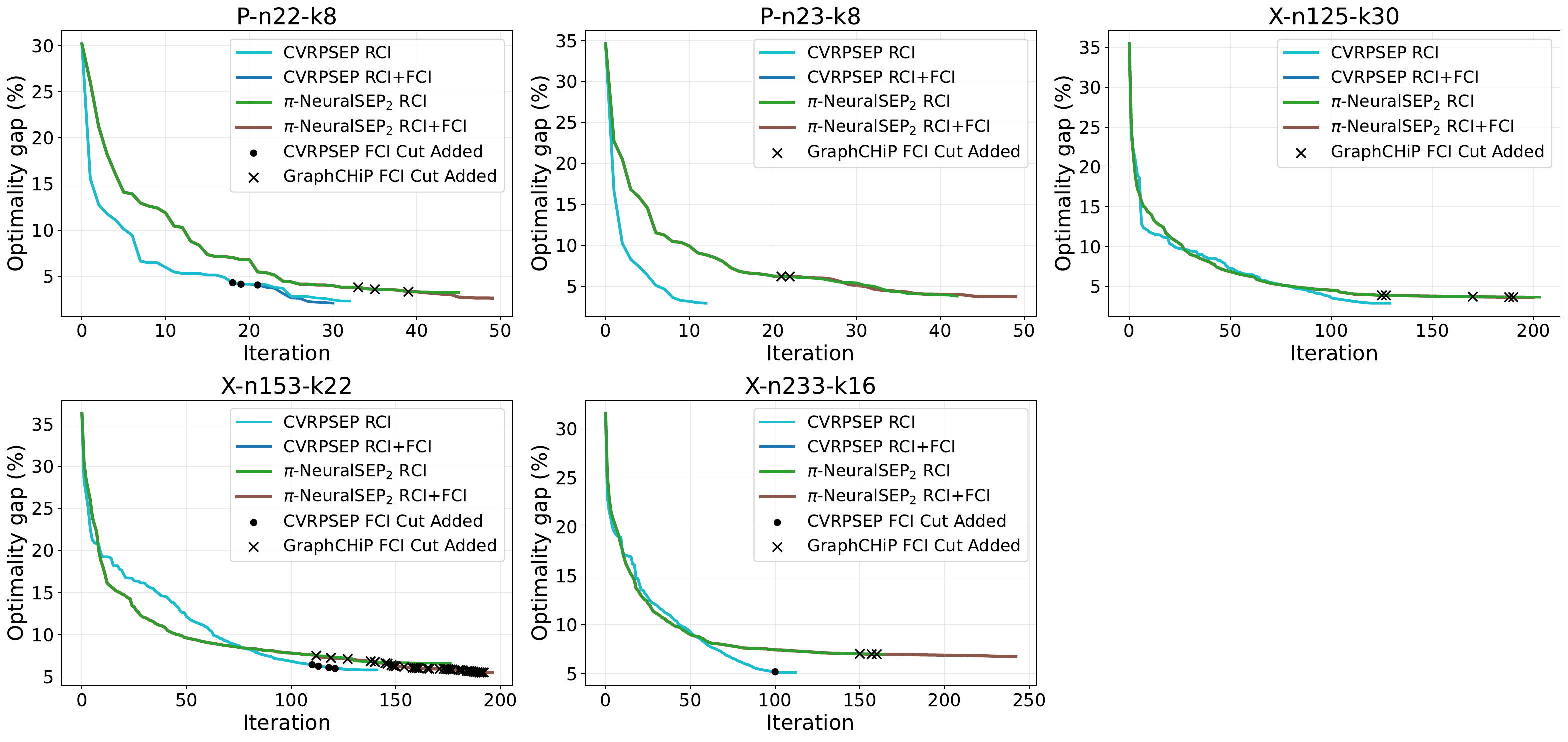}
    \caption{The optimality gap reduction and FCI cut addition points during the iterations}
    \label{fig:fci_gap_cut_summary}
\end{figure}

\section{Concluding Remarks}
This paper introduced a test-time search method that enhances a neural separation algorithm for capacity inequalities in CVRP.
We identified and addressed NeuralSEP's insufficient sensitivity in generating diverse candidate subsets through stochastic edge selection. %
Additionally, we proposed the GraphCHiP algorithm, which utilizes the node mapping history to identify additional RCIs and propose candidate partitions for generalized capacity inequalities, a subclass of FCIs.
We also observed that the coarsening history forms a laminar family, which allows GraphCHiP to obtain RCI and FCI candidates from a single hierarchy.
These results suggest that test-time search can improve a trained neural separator without architectural modifications or retraining.

However, several challenges and directions for future work remain. 
The FCI identification capability, while novel, has been demonstrated primarily on specific instances.
A broader evaluation is challenging because the presence and effectiveness of FCI cuts are highly instance-dependent, as many instances may inherently lack useful FCIs.
This instance-dependent effectiveness suggests a direction for future work: characterizing problem features through empirical analysis or developing methods to determine which problem instances would benefit from FCI separation.

Future work should also focus on integrating these techniques into a complete BC or BPC framework to validate their practical impact. 
This current work utilizes a BC framework as a base, focusing solely on solving the root node.
Integrating NeuralSEP within a BPC framework would require retraining, as the pattern of the separation problem could shift significantly.
As the test-time approach proposed in this paper is, however, independent of how the underlying neural coarsening procedure is trained, it can be applied unchanged to a retrained model, although whether its root-node gains carry over to a full BC or BPC algorithm remains to be validated.

\ifshowauthors
    \paragraph{Acknowledgements}
    This work was supported by the National Research Foundation of Korea (NRF) grant funded by the Korean government (MSIT) (RS-2023-00259550).
\fi

\section*{Data Availability Statement}
The randomly generated CVRP instances used in this study are publicly available at \url{https://github.com/hyeonahkimm/neuralsep}, and the benchmark instances are available from CVRPLIB at \url{http://vrp.galgos.inf.puc-rio.br/}.
The source code of the proposed methods is available at \url{https://github.com/syj5268/neuralsep-tts}.

\section*{Conflict of Interest Statement}
The authors declare no conflict of interest.

\bibliographystyle{abbrvnat}
\bibliography{reference}

\newpage
\appendix

\label{sec:appendix}
\input{appendix.tex}

\input{supplementary.tex}

\end{document}

%% file: tables/main_sep.tex
\begin{tabular}{lrrrr} 
    \toprule
    \multicolumn{1}{l}{Method} & \multicolumn{4}{c}{Size} \\
    \cmidrule(lr){2-5}
    &  50 & 75 & 100 & 200 \\
    \midrule
    Exact & 65.27\% & 69.58\% & 66.76\% & 60.41\% \\
    NeuralSEP & 37.11\% & 31.15\% & 27.38\% & 15.44\% \\
    \bottomrule
\end{tabular}

%% file: tables/compare_rci.tex
\centering
\begin{tabular}{r|rrrrrrrr}
    \toprule
    \multirow{2}{*}{\textbf{Size}} & \multicolumn{2}{c}{\textbf{CVRPSEP}} & \multicolumn{2}{c}{\textbf{NeuralSEP$_1$}} & \multicolumn{2}{c}{\textbf{NeuralSEP$_2$}} & \multicolumn{2}{c}{\textbf{$\boldsymbol{\pi}$-NeuralSEP$_2$ + GC}}\\
    \cmidrule(lr){2-3} \cmidrule(lr){4-5} \cmidrule(lr){6-7} \cmidrule(lr){8-9} 
    & Gap & Time/Iter & Gap  & Time/Iter & Gap  & Time/Iter & Gap  & Time/Iter \\
    \midrule
    50   & \textbf{1.970\%} & 0.009 & 4.151\% & 0.830 & 5.250\% & 0.120 & 3.647\% & 0.138\\
    75   & \textbf{2.769\%} & 0.054 & 5.305\% & 1.066 & 5.329\% & 0.159 & 4.518\% & 0.206\\
    100  & \textbf{4.539\%} & 0.145 & 6.611\% & 1.440 & 6.441\% & 0.289  & 5.930\% & 0.345\\
    200  & \textbf{6.280\%} & 2.001 & 9.214\% & 3.411 & 8.236\% & 1.161 & 7.653\% & 1.416\\
    300  & \textbf{7.903\%} & 10.431& 10.515\% & 12.006 & 10.219\% & 4.660 & 8.683\% & 6.461\\
    400  & 12.618\% & 16.936& 12.848\% & 26.714 & 13.786\% & 13.726 &  \textbf{10.830\%} & 23.953\\
    500  & 16.357\% & 16.947& 15.413\% & 41.227 & 14.856\% & 31.450 &  \textbf{13.349\%} & 44.739\\
    750  & 25.783\% & 16.603& 22.553\% & 102.623 & 22.159\% & 93.817 & \textbf{20.741\%} & 113.266\\
    1,000& 30.408\% & 23.321& 28.777\% & 161.183 & 26.505\% & 145.040 & \textbf{26.048\%} & 160.113\\
    \bottomrule 
\end{tabular}

%% file: tables/compare_rci_pi.tex
\centering
\begin{tabular}{r|rr|rrr}
    \toprule
    \multirow{2}{*}{\textbf{Size}} & \textbf{NeuralSEP$_2$} & \textbf{$\pi$-NeuralSEP$_2$} & \multicolumn{2}{c}{\textbf{Difference}} & \textbf{Winning Ratio}\\
    \cmidrule(lr){2-2} \cmidrule(lr){3-3} \cmidrule(lr){4-5} \cmidrule(lr){6-6}
    & RCI Cuts & RCI Cuts & \% Cut Increase & $\Delta$ Gap & $\pi$-NeuralSEP$_2$ Win \\
    \midrule
    50 & 130.9 & 165.8 & 26.7\% & $\downarrow$1.280\%p & 0.9 \\
    75 & 375.0 & 393.8 & 5.0\% & $\downarrow$0.299\%p & 0.7 \\
    100 & 731.3 & 733.5 & 0.3\% & $\downarrow$0.151\%p & 0.6 \\
    200 & 1440.5 & 1509.3 & 4.8\% & $\downarrow$0.134\%p & 0.7 \\
    300 & 2270.9 & 3180.2 & 40.0\% & $\downarrow$1.128\%p & 0.9 \\
    400 & 2297.6 & 3464.9 & 50.8\% & $\downarrow$2.594\%p & 1.0 \\
    500 & 2169.3 & 2657.6 & 22.5\% & $\downarrow$1.371\%p & 1.0 \\
    750 & 1678.2 & 1771.4 & 5.6\% & $\downarrow$1.550\%p & 1.0 \\
    1000 & 1289.6 & 1334.9 & 3.5\% & $\downarrow$0.363\%p & 0.7 \\
    \bottomrule
\end{tabular}

%% file: tables/compare_methods.tex
\centering
\small
\begin{tabular}{lrrr}
\toprule
\textbf{Methods} & \textbf{Size} & \textbf{Gap (\%)} & \textbf{\# of Cuts} \\
\midrule
\multirow{6}{*}{Stochastic Roulette}
 & 50  & 4.612  & 141.1 \\
 & 75  & 5.113  & 361.7 \\
 & 100 & \textbf{6.262}  & 662.9 \\
 & 200 & 8.289  & 1,358.6 \\
 & 300 & 9.227  & 2,905.0 \\
 & 400 & 11.303 & 3,290.5 \\
\midrule
\multirow{6}{*}{Softmax}
 & 50  & 4.526  & 142.3 \\
 & 75  & 5.180  & 393.3 \\
 & 100 & 6.463  & 688.9 \\
 & 200 & 8.156  & 1,435.2 \\
 & 300 & 9.137  & 3,089.2 \\
 & 400 & 11.312 & 3,407.6 \\
\midrule
\multirow{6}{*}{$\pi$-Greedy}
 & 50  & \textbf{3.970}  & \textbf{165.8} \\
 & 75  & \textbf{5.030}  & \textbf{393.8} \\
 & 100 & 6.290  & \textbf{733.5} \\
 & 200 & \textbf{8.102}  & \textbf{1,509.3} \\
 & 300 & \textbf{9.091}  & \textbf{3,180.2} \\
 & 400 & \textbf{11.192} & \textbf{3,464.9} \\
\bottomrule
\end{tabular}
\label{tab:three_methods_comparison}

%% file: tables/compare_rci_graphchip.tex
\centering
\begin{tabular}{r|rr|rrr}
    \toprule
    \multirow{2}{*}{\textbf{Size}} & \textbf{$\pi$-NeuralSEP$_2$} & \textbf{+ GC} & \multicolumn{2}{c}{\textbf{Difference}} & \textbf{Winning Ratio}\\
    \cmidrule(lr){2-2} \cmidrule(lr){3-3} \cmidrule(lr){4-5} \cmidrule(lr){6-6}
    & RCI Cuts & RCI Cuts & \% Cut Increase & $\Delta$ Gap & $\pi$-NeuralSEP$_2$+GC Win \\
    \midrule
    50 & 165.8 & 192.3 & 16.0\% & $\downarrow$0.323\%p & 0.5 \\
    75 & 393.8 & 475.2 & 20.7\% & $\downarrow$0.512\%p & 0.8 \\
    100 & 733.5 & 806.1 & 9.9\% & $\downarrow$0.360\%p & 0.8 \\
    200 & 1509.3 & 1752.6 & 16.1\% & $\downarrow$0.449\%p & 1.0 \\
    300 & 3180.2 & 3620.1 & 13.8\% & $\downarrow$0.408\%p & 1.0 \\
    400 & 3464.9 & 3824.0 & 10.4\% & $\downarrow$0.362\%p & 1.0 \\
    500 & 2657.6 & 2817.6 & 6.0\% & $\downarrow$0.136\%p & 0.6 \\
    750 & 1771.4 & 1806.7 & 2.0\% & $\uparrow$0.132\%p & 0.5 \\
    1000 & 1334.9 & 1373.9 & 2.9\% & $\downarrow$0.094\%p & 0.55 \\
    \bottomrule
\end{tabular}

%% file: tables/x_instance_fci.tex
\centering

\begin{tabular}{llccc}
\toprule
\textbf{Setting} & \textbf{Method} & \textbf{Instance} & \textbf{Gap (\%)} & \textbf{FCI cuts} \\
\midrule
\multirow{10}{*}{RCI}
& \multirow{5}{*}{CVRPSEP RCI} 
& P-n22-k8 & \textbf{2.32} & -- \\
& & P-n23-k8 & \textbf{2.93} & -- \\
& & X-n125-k30 & \textbf{2.91} & -- \\
& & X-n153-k22 & \textbf{5.83} & -- \\
& & X-n233-k16 & \textbf{5.14} & -- \\
\cmidrule(lr){2-5}
& \multirow{5}{*}{$\pi$-NeuralSEP$_2$}
& P-n22-k8 & 3.27 & -- \\
& & P-n23-k8 & 3.78 & -- \\
& & X-n125-k30 & 3.66 & -- \\
& & X-n153-k22 & 6.59 & -- \\
& & X-n233-k16 & 7.01 & -- \\
\midrule
\multirow{15}{*}{RCI+FCI}
& \multirow{5}{*}{\shortstack{CVRPSEP RCI \\+ CVRPSEP FCI}}
& P-n22-k8 & 2.09 & 13 \\
& & P-n23-k8 & 2.93 & 0 \\
& & X-n125-k30 & 2.91 & 0 \\
& & X-n153-k22 & 5.82 & 4 \\
& & X-n233-k16 & \textbf{5.14} & 1 \\
\cmidrule(lr){2-5}
& \multirow{5}{*}{\shortstack{$\pi$-NeuralSEP$_2$ \\+ GraphCHiP FCI}}
& P-n22-k8 & 2.64 & 3 \\
& & P-n23-k8 & 3.72 & 3 \\
& & X-n125-k30 & 3.64 & 5 \\
& & X-n153-k22 & 5.52 & 210 \\
& & X-n233-k16 & 6.77 & 3 \\
\cmidrule(lr){2-5}
& \multirow{5}{*}{\shortstack{CVRPSEP RCI \\+ CVRPSEP FCI \\+ GraphCHiP FCI}}
& P-n22-k8 & \textbf{2.00} & 13 (12 + 1) \\
& & P-n23-k8 & \textbf{2.80} & 1 (0 + 1) \\
& & X-n125-k30 & \textbf{2.89} & 3 (0 + 3) \\
& & X-n153-k22 & \textbf{4.40} & 295 (16 + 279) \\
& & X-n233-k16 & \textbf{5.14} & 1 (1 + 0) \\
\bottomrule
\end{tabular}

%% file: appendix.tex
\section{Proofs}
\label{sec:proof}

\subsection{Remark~\ref{rem:structure}}
\label{sec:proof-1}
Each coarsening step merges whole supernodes.
Hence, the sets at each level partition $V$, and for $t \le t'$ each set at level $t$ lies in exactly one set at level $t'$.
Since $S$ is a union of sets at level $T$, every set at a lower level lies either inside or outside $S$, which gives (i).
Two sets from levels $t \le t'$ are nested or disjoint because the sets at level $t'$ are pairwise disjoint, which gives (ii).
Statement (iii) follows from (i), since the singletons outside $S$ persist across levels and each block of $\Omega_t$ inside $S$ lies in a block of $\Omega_{t+1}$ inside $S$.

\subsection{Proposition~\ref{prop:rci}}
\label{sec:proof-2}
\begin{proof} 
    At each coarsening step $t\in\{1,\dots,T\}$, GraphCHiP considers supernodes $u\in V_t$ that are consistent with the subset $S$ identified by NeuralSEP and performs at most one RCI check per such supernode (the procedure terminates immediately if a violating subset is found).  
    Hence, the total number of inequality checks is upper bounded by
    \begin{equation}
        N_{\text{checks}} \ \le \ \sum_{t=1}^{T} |V_t|.
    \end{equation}
    By the construction of the $\gamma$-coarsening used in NeuralSEP, the number of vertices shrinks by a constant factor at each level: specifically, $|V_t|\le \gamma^t |V|$ for some $\gamma\in(0,1)$. If the coarsening terminates early because all contraction probabilities become zero, the number of levels only decreases, so the bound still holds. 
    Therefore,
    \begin{equation}
        \sum_{t=1}^{T} |V_t|
        \ \le\
        |V| \sum_{t=1}^{T} \gamma^t
        \ \le\
        |V| \sum_{t=1}^{\infty} \gamma^t
        \ =\
        |V|\cdot \frac{\gamma}{1-\gamma}.
    \end{equation}

\end{proof}

\subsection{Proposition~\ref{prop:fci}}
\label{sec:proof-3}
\begin{proof} 
The number of iterations in the GraphCHiP algorithm is determined by the depth of the neural graph coarsening process.
Since the number of vertices is reduced by a constant factor $\gamma < 1$ at each step, the total number of coarsening steps $T$ is bounded by $O(\log |V|)$.
The algorithm then iterates backward through these $T$ steps of the coarsening history.
At each step $t$, it constructs exactly one candidate partition $\Omega$ of the customer set $V_C$, which takes $O(|V|)$ time using the node map.
Labeling each vertex with its block, a single pass over $\bar{E}$ yields the cut values $\bar{x}(\delta(h))$ of all blocks $h \in \Omega$, which are needed for the RCI check and the left-hand side, in $O(|\bar{E}| + |V|)$ time.
Instead of solving a BPP to obtain $r(\Omega)$ exactly, we evaluate the FCI with a lower bound on $r(\Omega)$.
In particular, we use the $L_2$ bound from \citet{martello1990knapsack}, which has a time complexity of $O(n \log n)$ for $n$ items, as the calculation is dominated by sorting.
Since each customer demand is at most $Q$, a block $h$ yields at most $|h|+1$ items, so $n \le 2|V_C|$ and this step takes $O(|V| \log |V|)$ time.
Multiplying the number of iterations by the cost per iteration, $O(|\bar{E}| + |V| \log |V|)$, gives a total running-time bound of $O\big((|\bar{E}| + |V| \log |V|) \log |V|\big)$.
Since $\bar{E} \subseteq E$ and $|E| = \binom{|V|}{2} = O(|V|^2)$, this yields the worst-case bound $O\big(|V|^2 \log |V|\big)$ in terms of the original graph size.
If the support graph is sparse, with $|\bar{E}| = O(|V|)$, the more precise bound reduces to $O\big(|V| (\log |V|)^2\big)$.
\end{proof}

\section{Pseudo-code of GraphCHiP for RCI Separation}
\label{sec:algo_rci}
Algorithm~\ref{algo:graphchip_rci} presents the pseudo-code of GraphCHiP for RCI separation described in Section~\ref{sec:graphchip}.

\begin{algorithm}[!ht]
    \caption{GraphCHiP for RCI} \label{algo:graphchip_rci}
    \small
    \begin{algorithmic}[1]
    \Require
        Support Graph $\Gc$, \par
        Subset $S \subseteq V_C$ and its RCI violation $v$ \par
        A node map $\{\mathcal{M}_t\}_{t=1}^{T}$ \par
    \Ensure  A collection $\Rc$ of violated RCI subsets $h$ and their corresponding RHS values %
        \State $\Rc \gets \emptyset$, $t \gets T$ \Comment{Initialize. Iterate backwards from the coarsest graph.}
        \If{$v > 0$} \Comment{If the violation of the original subset $S$ is positive, do not search further.}
            \State \Return $\Rc$
        \EndIf
        \While{$t > 0$}  
            \State $S_t \gets \{ u \in V_t \mid \mathcal{M}_t(u) \cap S \neq \emptyset \}$
            \Comment{Lift $S$ to level $t$}
            \ForEach{$u \in S_t$}
                \State $h \gets \mathcal{M}_t(u)$
                \Comment{Original nodes aggregated into supernode $u$}
                \State LHS $\gets \text{Calculate\_LHS}(\Gc, h)$
                \State RHS $\gets 2 \left\lceil\frac{d(h)}{Q}\right\rceil$
                \If{$\text{RHS} - \text{LHS} > 0$}  \Comment{Check if the violation exists}
                    \State Add ($h$, RHS) to $\Rc$
                \EndIf
            \EndFor
            \If{$\Rc \neq \emptyset$}
                \State \Return $\Rc$
            \EndIf
            \State $t \gets t - 1$  \Comment{Backtrack to the previous coarsening step}
        \EndWhile
        \\
        \Return $\Rc$
    \end{algorithmic}
\end{algorithm}

\section{Additional Analysis of NeuralSEP}
\subsection{Details of GNN in NeuralSEP}
\label{sec:gnn}
\begin{figure}[h!]
    \centering
    \captionsetup{justification=centering}
    \includegraphics[width=\textwidth]{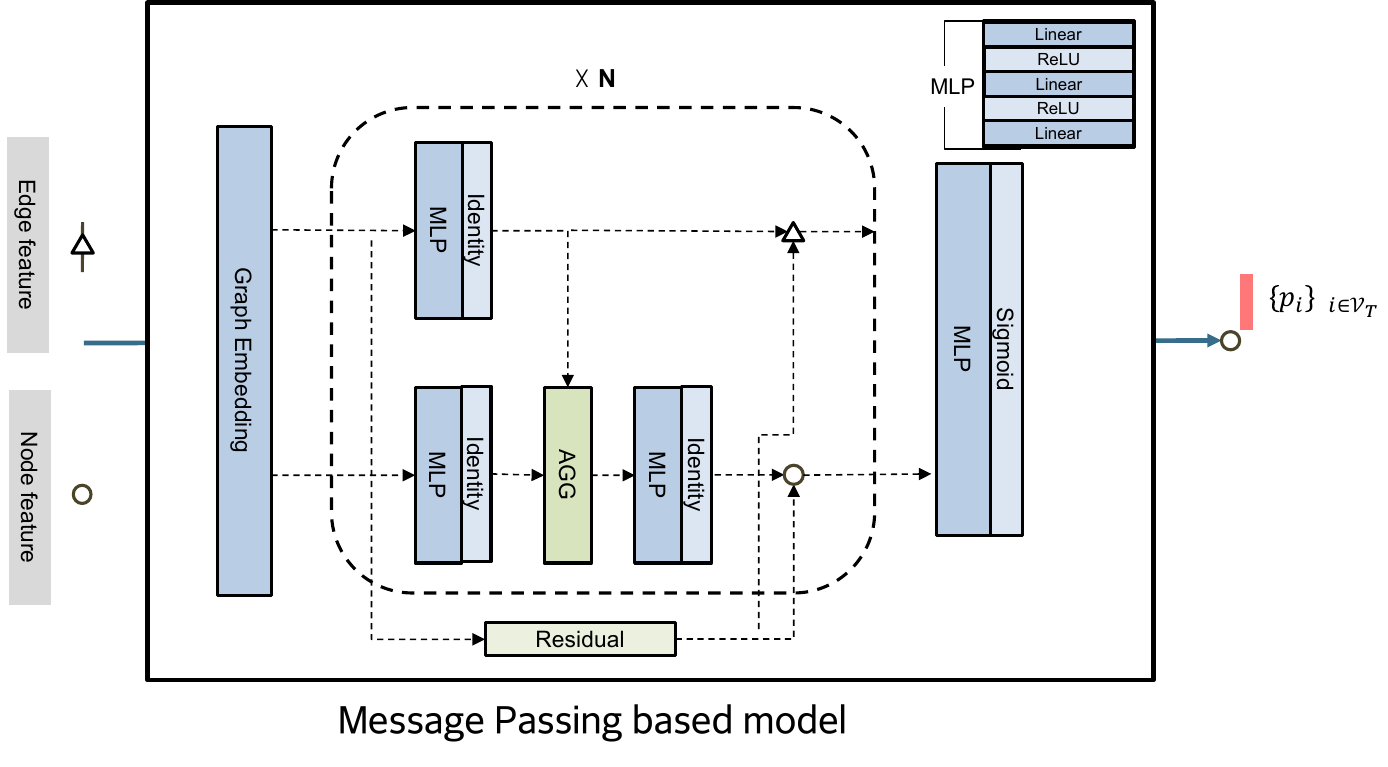}
    \caption{The GNN-based model of NeuralSEP}
    \label{fig:gnn}
\end{figure}

Figure~\ref{fig:gnn} shows the GNN of NeuralSEP, which is identical in the original work and ours because modifying the model would not be a test-time approach.
The input features are embedded into a higher-dimensional space and then processed by message-passing layers with multi-layer perceptrons.

\subsection{Mathematical Analysis of the Intractability of the Output Difference}
\label{sec:math_analysis}
We explain why the sensitivity of the GNN-based model $\Phi$ to the input parameter $m$ admits no closed-form expression.
Let $\mathbf{x}_m$ denote the input derived from $G^\ell(m)$, so that $\Phi_i(\mathbf{x}_m) = p_i$ is the predicted probability for vertex $i$, and define the sensitivity as
\begin{equation}
D(m) = \|\Phi(\mathbf{x}_{m+1}) - \Phi(\mathbf{x}_m)\|.
\end{equation}
The model is a composition $\Phi = f_C \circ \dots \circ f_1 \circ f_{\text{enc}}$ of a linear encoder and $C$ layers of the form $f_c(\mathbf{h}) = \sigma(\mathcal{A}_c(\mathbf{h}))$, where $\mathcal{A}_c$ is a linear transformation including aggregation and $\sigma$ is a non-linear activation.
Since only the second input feature changes by $1/K$, the embedding difference $\Delta \mathbf{h}^{(0)} = W_{\text{enc}}(\mathbf{x}_{m+1} - \mathbf{x}_m)$ is constant in $m$.
A first-order Taylor expansion of each layer around $\mathbf{a}^{(c)}_m = \mathcal{A}_c(\mathbf{h}^{(c-1)}_m)$ then gives
\begin{equation}
\Delta \mathbf{h}^{(C)} \approx \left( \prod_{c=C}^{1} J_{\sigma}(\mathbf{a}^{(c)}_m)\, \mathcal{A}_c \right) \Delta \mathbf{h}^{(0)},
\end{equation}
where $J_{\sigma}(\mathbf{a}^{(c)}_m)$ is the Jacobian of $\sigma$ at $\mathbf{a}^{(c)}_m$.
Although $\Delta \mathbf{h}^{(0)}$ is constant, every Jacobian is evaluated at a different point that depends on $m$, so the product does not reduce to an interpretable function of $m$.
This motivates the empirical analysis in Section~\ref{sec:problem_neuralsep}.

\clearpage
\subsection{Similarity of Predicted Probabilities for Adjacent Values of $m$}
\label{sec:cosine}
\begin{figure}[!ht]
    \centering
    \includegraphics[width=0.45\textwidth]{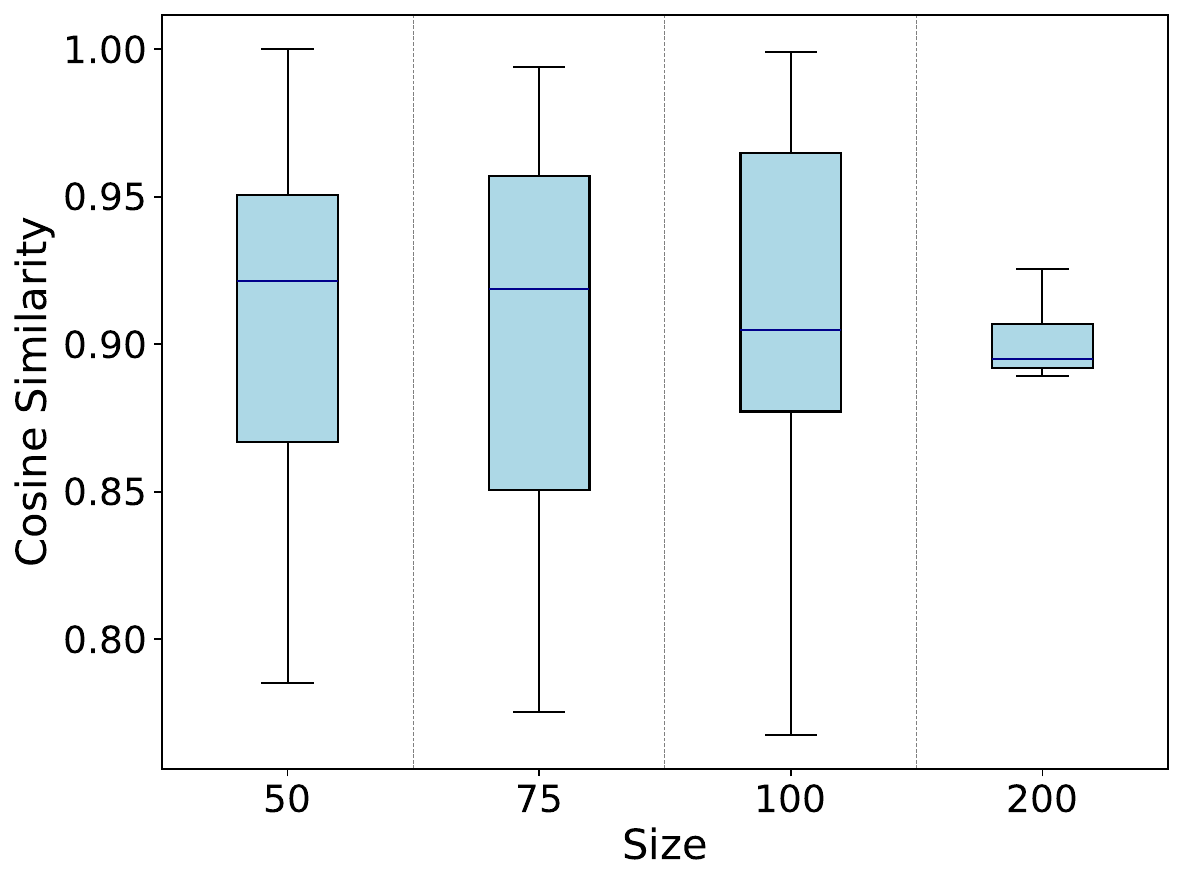}
    \caption{Cosine similarity between the predicted probabilities for adjacent values of $m$}
    \label{fig:cosine_sim}
\end{figure}

To complement the analysis in Section~\ref{sec:problem_neuralsep}, we measure the cosine similarity between the initial predicted probability vectors for adjacent values of $m$,
\begin{equation}
    \frac{\Phi(G^\ell(m)) \cdot \Phi(G^\ell(m+1))}{\|\Phi(G^\ell(m))\| \, \|\Phi(G^\ell(m+1))\|}, \quad m \in \{0, \ldots, K-2\},
\end{equation}
where $\Phi(G^\ell(m)) = (p_i)_{i \in V^\ell}$.
As shown in Figure~\ref{fig:cosine_sim}, the similarity is consistently around 0.9 for all graph sizes, indicating that the predictions for adjacent values of $m$ are highly similar.
Since the cosine similarity between nonnegative vectors tends to be high, we use it only as supplementary evidence to the metrics in Section~\ref{sec:problem_neuralsep}.

\section{Experimental Environment}
\label{sec:environment}
Table~\ref{tab:experiment_env} summarizes the hardware and software environment used in our experiments.
\begin{table}[!ht]
    \centering
    \captionsetup{justification=centering}
    \caption{Experimental environment}
    \label{tab:experiment_env}
    \begin{tabular}{>{\raggedright\arraybackslash}m{1.5in} l}
    \toprule
    \textbf{Component} & \textbf{Details} \\ \midrule
    CPU & AMD Ryzen 9 5900X \\
    Memory & 62 GB DDR4 RAM @ 3200 MT/s \\
    GPU (training only) & NVIDIA GeForce RTX 4070 \\ \midrule
    Framework & PyTorch 2.4.0 \\
              & Python 3.9.19 \\
              & Julia 1.9.4 \\ \midrule
    Solver and library & IBM ILOG CPLEX 22.1.1 (via CPLEX.jl v1.0.3) \\
            & PyCall v1.96.4 \\
            & JuMP v1.23.0 \\ \midrule
    Operating System & Ubuntu 22.04 \\
    \bottomrule
    \end{tabular}
\end{table}

\section{Experimental Results on Separation Problems}
In this section, we provide additional experimental results on the separation problems used to evaluate original NeuralSEP.
The number of separation problems is 100 for each customer size, ranging from 50 to 200.
The test sets are the same as those used in Section~\ref{sec:problem_neuralsep}.

\subsection{Implementation: DGL vs. PyG}
\label{sec:dgl_pyg}
In Table~\ref{table:sep_time}, we compare the average evaluation time of NeuralSEP implemented in DGL and PyG.
Both implementations are evaluated on the CPU, and the GPU is used only for training.
We observe that the PyG implementation is significantly faster than the DGL implementation across all customer sizes.
This speedup stems from the \emph{neural graph coarsening} step, where DGL introduces overhead by updating its internal graph index after each structural modification. 
Conversely, PyG operates directly on tensor representations (i.e., the edge list), bypassing these indexing steps for greater efficiency.

\begin{table} 
    \captionsetup{justification=centering}
    \caption{The average evaluation time comparison between DGL and PyG}
    \centering
    \label{table:sep_time}
    \input{tables/sep_time}
\end{table}

\subsection{Performance of NeuralSEPs for RCI}
We evaluate the separation algorithms using three metrics, namely the average maximum violation per problem, the total number of unique violated inequalities, and the success rate of finding at least one violated RCI.
Table~\ref{table:separation2} summarizes the results, where NeuralSEP$_1$ denotes the original DGL implementation of NeuralSEP.
CVRPSEP finds cuts with smaller maximum violations than the NeuralSEP variants, but finds a violated RCI more often for instances with 75 or more customers.
The NeuralSEP variants find cuts with larger violations, but their success rates decline as the problem size grows.
Adding $\pi$-greedy selection and GraphCHiP to NeuralSEP$_2$ increases the number of unique inequalities for all sizes, mainly due to GraphCHiP, whereas the success rate of the full approach improves slightly, by at most 0.06.

\begin{table}
    \captionsetup{justification=centering}
    \caption{Separation results for RCI on various sizes of problem}
    \centering
    \label{table:separation2}
        \input{tables/sep_violation}

\end{table}

\clearpage

\section{Additional Results for RCI Separation Algorithms}
\subsection{Detailed Results for RCI Separation Algorithms}
Table~\ref{table:random_runtime_rci} summarizes the detailed results of Table~\ref{table:compare_rci}.
\begin{table}[!ht]
    \captionsetup{justification=centering}
    \caption{Performance comparison of RCIs on randomly generated CVRP instances}
    \label{table:random_runtime_rci}
    \resizebox{\textwidth}{!}{
    \input{tables/random_runtime}

    }
    \vspace{0.1cm}
\end{table}

\subsection{Complete Results on the Impact of the Number of Cuts per Iteration}
\label{sec:pi_results}
\begin{figure}
    \centering
    \includegraphics[width=1.0\textwidth]{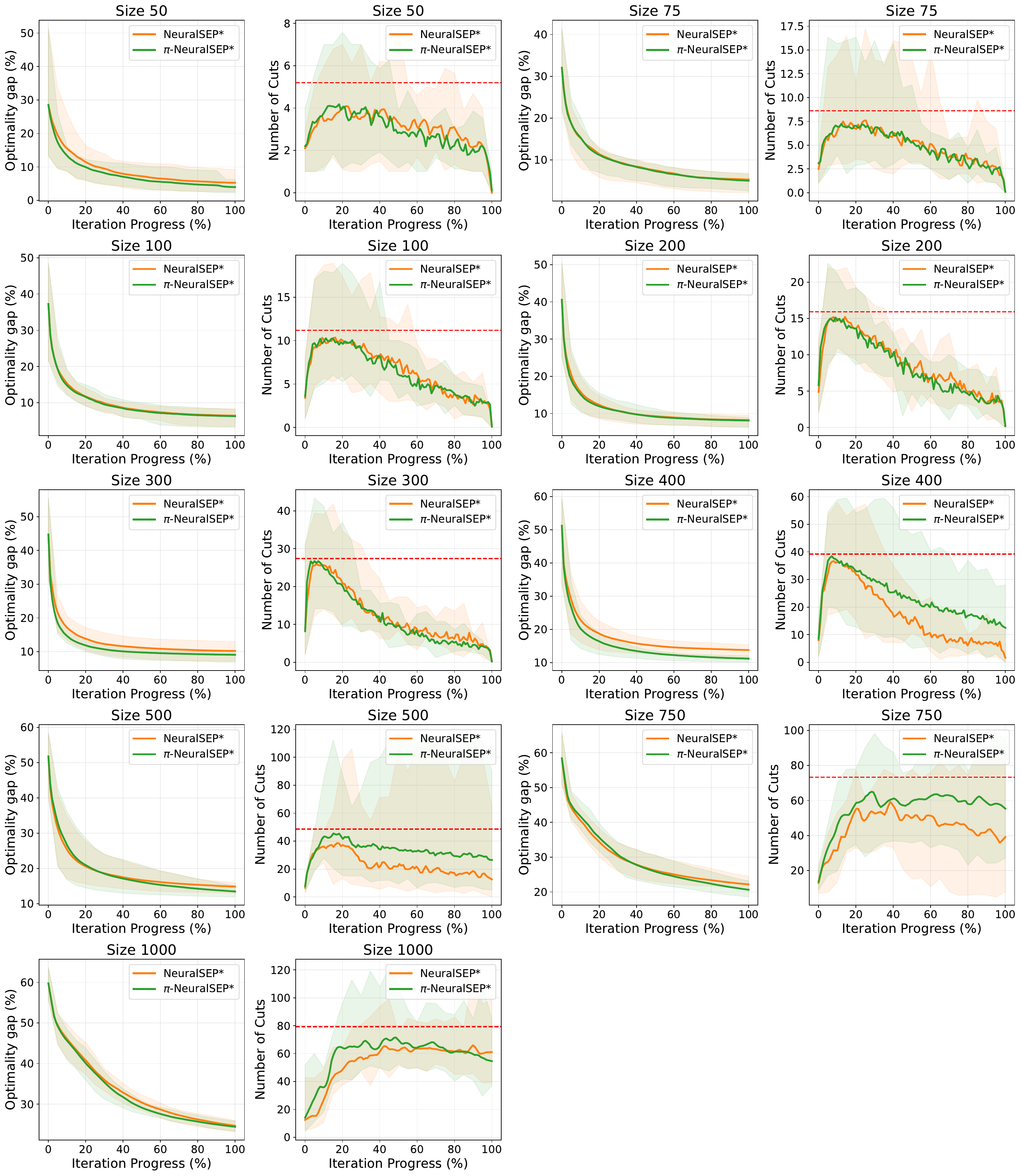}
    \caption{The optimality gap and number of cuts added over relative iterations for different customer sizes}
    \label{fig:gap_cut_summary}
\end{figure}

Additionally, Figure~\ref{fig:gap_cut_summary} presents the full set of results, comparing the performance of the greedy selection and $\pi$-greedy methods across all customer sizes.
The two methods perform similarly for instances with 75 to 200 customers, whereas $\pi$-greedy selection yields clearly more cuts and lower gaps for instances with 50 customers and with 300 to 750 customers.
In the figure, the shaded bands indicate variability across instances, showing the min–max range over all runs.

\subsection{Results on Size-1000 Instances with a 2-Hour Time Limit}
Size 1000 instances are evaluated within a 2-hour time limit.
For the instances with 1000 customers, the impact of GraphCHiP becomes more evident when additional computation time is available.

\begin{table}[!ht]
    \captionsetup{justification=centering}
    \caption{Performance comparison of RCIs on size-1000 instances with a 2-hour time limit}
    \label{table:random_runtime_rci_1000}
    \resizebox{\textwidth}{!}{
    \input{tables/random_runtime_1000} 
    }
    \begin{tablenotes}
    \footnotesize
    \item \textit{Note.} The winning ratio in each row denotes the fraction of instances for which the corresponding method
    achieves a lower optimality gap than the method listed in the immediately preceding row.
    \end{tablenotes}
\end{table}

\section{Case Study: Illustrative Example of FCI Identification}
\label{sec:case_study}
\begin{figure}
    \centering
    \includegraphics[width=\textwidth]{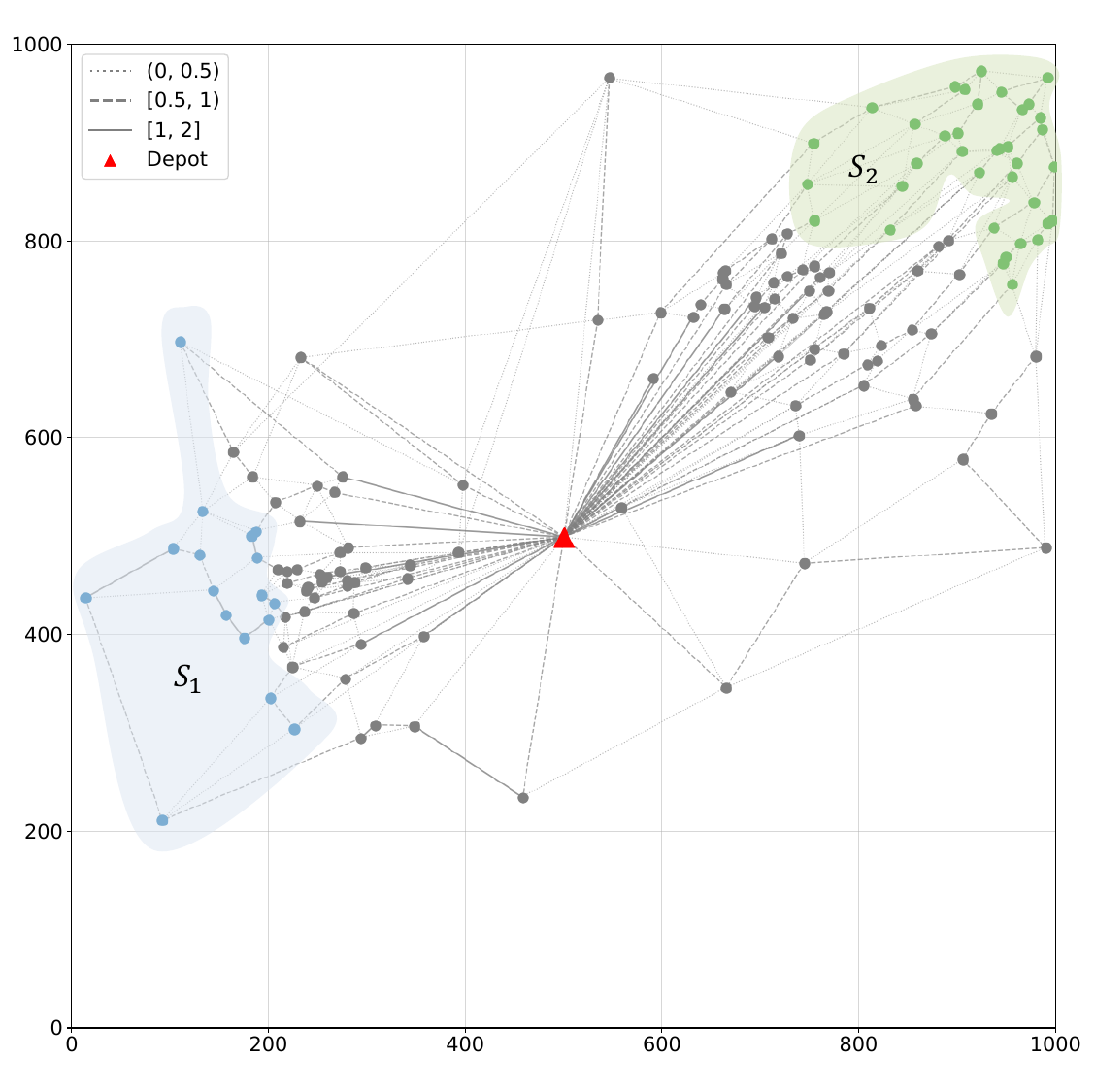}
    \caption{An example of an FCI found by GraphCHiP on \texttt{X-n153-k22} instance. 
    The blue and green nodes represent the two non-singleton subsets $S_1$ and $S_2$ in the partition $\Omega$, respectively, while the remaining nodes are singleton subsets. 
    The total demands of $S_1$ and $S_2$ are 442 and 811, respectively, and the total customer demand and the vehicle capacity are 3068 and 144.
    The set $H$ for the partition is the entire set of customers $V_C$, and the corresponding cut values are $\bar{x}(\delta(H))=44.0$, $\bar{x}(\delta(S_1))=8.15$, $\bar{x}(\delta(S_2))=13.05$.
    The $L_2$ lower bound on the bin-packing value is $L_2(\Omega) = 23$, larger than $\lceil \frac{d(V_C)}{Q} \rceil = 22$.
    }
    \label{fig:fractional_subset_X}
\end{figure}

To illustrate the mechanism behind the identified FCIs, we next examine a representative CVRPLIB instance in detail.
The \texttt{X-n153-k22} instance is characterized by a skewed demand distribution, with a large number of low-demand customers and a smaller group of high-demand customers.
Figure~\ref{fig:fractional_subset_X} illustrates the specific FCI found by GraphCHiP on this instance, which has a violation of 0.8.
As detailed in the caption, none of $V_C$, $S_1$, and $S_2$ violates its own RCI, so the partition passes the filtering described in Section~\ref{sec:graphchip}.
The left-hand side (LHS) of the inequality, which is the sum of the relevant cut values, is $\bar{x}(\delta(V_C)) + \bar{x}(\delta(S_1)) + \bar{x}(\delta(S_2))=65.20$.
The right-hand side (RHS) is $2 L_2(\Omega) + 2 \lceil \frac{d(S_1)}{Q} \rceil + 2 \lceil \frac{d(S_2)}{Q} \rceil = 46 + 8 + 12 = 66$.
Since the LHS is less than the RHS, the inequality is violated by $0.8$.
The terms for singleton subsets cancel out by the degree equations and are therefore omitted.
The violation of 0.8 equals the gain from bin packing, $2(L_2(\Omega) - \lceil d(V_C)/Q \rceil) = 2$, minus the total slack of the RCIs for $V_C$, $S_1$, and $S_2$, $(44.0 - 44) + (8.15 - 8) + (13.05 - 12) = 1.2$, that is, $2 - 1.2 = 0.8$.
Thus, an FCI can be violated even when all of its blocks satisfy their RCIs, provided that the bin-packing gain exceeds their total slack.

%% file: tables/sep_time.tex
\begin{tabular}{rrr} 
    \toprule
    \small
    Size & \multicolumn{2}{c}{Avg. Evaluation Time} \\ \cmidrule{2-3}
    & NeuralSEP$_1$ & NeuralSEP$_2$ \\
    \midrule
    50 & 45.13 & 8.45  \\
    75 & 76.34 & 17.47 \\
    100 & 104.94 & 26.83 \\
    200 & 225.29 & 76.74 \\
    \bottomrule
\end{tabular}

%% file: tables/sep_violation.tex
\begin{tabular}{llrrrr} 
    \toprule
    \multicolumn{1}{l}{Metric} & \multicolumn{1}{l}{Method} & \multicolumn{4}{c}{Size} \\
    \cmidrule(lr){3-6}
    &  &  50 & 75 & 100 & 200 \\
    \midrule
    \multirow{6}{*}{Avg. Max Violations}
    & Exact & 2.2426 & 3.0065 & 2.9619 & 2.1373 \\
    & CVRPSEP & 1.4096 & 1.3535 & 1.4556 & 0.6022 \\
    & NeuralSEP$_1$ & 1.8923 & 2.6155 & 2.5240 & 1.5827 \\
    & NeuralSEP$_2$ &  1.8611 & 2.6852 & 2.5983 & 1.6373 \\
    & $\pi$-NeuralSEP$_2$ &  1.9359 & 2.6908 & 2.5719 & 1.6289 \\
    & $\pi$-NeuralSEP$_2$ + GC &  1.9480 & 2.6996 & 2.5866 & 1.6357 \\
    \midrule
    \multirow{6}{*}{Num of Inequalities} 
    & Exact & 357 & 670 & 707 & 763 \\
    & CVRPSEP & 451 & 364 & 410 & 302 \\
    & NeuralSEP$_1$ & 203 & 300 & 290 & 195 \\
    & NeuralSEP$_2$ & 190 & 352 & 299 & 207 \\
    & $\pi$-NeuralSEP$_2$ &  199 & 364 & 305 & 213 \\
    & $\pi$-NeuralSEP$_2$ + GC & 239 & 389 & 332 & 245 \\
    \midrule
    \multirow{6}{*}{Success Rate} 
    & Exact & 1.00 & 0.96 & 0.99 & 1.00 \\
    & CVRPSEP & 0.73 & 0.72 & 0.83 & 0.83 \\
    & NeuralSEP$_1$ & 0.71 & 0.57 & 0.49 & 0.42 \\
    & NeuralSEP$_2$ & 0.70 & 0.63 & 0.55 & 0.34 \\
    & $\pi$-NeuralSEP$_2$ &  0.71 & 0.68 & 0.52 & 0.36 \\
    & $\pi$-NeuralSEP$_2$ + GC &  0.75 & 0.69 & 0.56 & 0.38 \\
    \bottomrule
\end{tabular}

%% file: tables/random_runtime.tex
\centering
\begin{tabular}{m{0.9in}|r|rrrrrrr}
	\toprule
	Method 
	&  Size & Avg. Gap ($\downarrow$) & Avg. LB ($\uparrow$) & Avg. Runtime & Avg. Iter. & Avg. Cuts \\ \midrule
	\multirow{9}{*}{\shortstack[l]{CVRPSEP}} 
    &    50  & \textbf{1.970\%} & \textbf{9,363.668} & 0.22 & 24 & 827.6 \\
    &    75  & \textbf{2.769\%} & \textbf{13,355.214} & 2.01 & 37 & 1,727.9 \\
    &   100  & \textbf{4.539\%} & \textbf{15,937.063} & 7.70 & 53 & 3,130.7 \\
    &   200  & \textbf{6.280\%} & \textbf{21,378.288} & 238.07 & 119 & 8,639.5 \\
    &   300  & \textbf{7.903\%} & \textbf{31,765.826} & 2,263.52 & 217 & 16,561.1 \\
    &   400  & 12.618\% & 43,595.507 & 3,624.40 & 214 & 14,637.6\\
    &   500  & 16.357\% & 50,157.146 & 3,626.66 & 214 & 11,076.6\\
    &   750  & 25.783\% & 68,248.574 & 3,652.60 & 220 & 5,041.4\\
    & 1,000  & 30.408\% & 65,667.156 & 3,661.34 & 157 & 3,538.5\\ \midrule
	\multirow{9}{*}{\shortstack[l]{NeuralSEP$_1$}} 
    &    50  & 4.151\% & 9,150.977 & 31.54 & 38 & 152.6 \\
    &    75  & 5.305\% & 13,054.606 & 66.09 & 62 & 386.2\\
    &   100  & 6.611\% & 15,599.662 & 135.37 & 94 & 642.8\\
    &   200  & 9.214\% & 20,725.490 & 419.53 & 123 & 1,222.3\\
    &   300  & 10.515\% & 31,057.145 & 1,812.96 & 151 & 2,687.8\\
    &   400  & 12.848\% & 43,824.771 & 3,018.70 & 113 & 3,221.0\\
    &   500  & 15.413\% & 50,809.302 & 3,628.01 & 88 & 2,932.3\\
    &   750  & 22.553\% & 71,140.558 & 3,694.43 & 36 & 2,357.6\\
    & 1,000  & 28.777\% & 67,278.087 & 3,707.21 & 23 & 1,658.4\\ \midrule
	\multirow{9}{*}{\shortstack[l]{NeuralSEP$_2$}}
	&    50  & 5.250\%  & 9,056.390  &  4.19   & 38 & 130.9  \\
	&    75  & 5.329\%  & 13,025.135 & 11.81   & 67  & 375.0 \\
	&   100  & 6.441\%  & 15,629.647 & 31.98   & 104 & 731.3 \\
	&   200  & 8.236\%  & 20,931.698 & 191.52  & 159 & 1,440.5 \\
	&   300  & 10.219\% & 30,955.051 & 868.84  & 195 & 2,270.9 \\
	&   400  & 13.786\% & 43,301.010 & 1,820.45 & 145 & 2,297.6  \\
	&   500  & 14.856\% & 51,444.770 & 3,270.40 & 129 & 2,169.3 \\
	&   750  & 22.159\% & 71,670.619 & 3,729.53 & 42  & 1,678.2 \\
	& 1,000  & 26.505\% & 69,439.154 & 3,780.22 & 26 & 1,289.6 \\ \midrule
	\multirow{9}{*}{\shortstack[l]{$\pi$-NeuralSEP$_2$}} 
	&    50 & 3.970\% & 9,179.732 &  5.85 & 51 & 165.8  \\
	&    75 & 5.030\% & 13,064.785 &  12.86 & 69 & 393.8 \\
	&   100 & 6.290\% & 15,652.844 & 33.30 & 108 & 733.5 \\
	&   200 & 8.102\% & 20,973.412 & 227.37  & 176 & 1,509.3 \\
	&   300 & 9.091\% & 31,463.210 & 1,655.96 & 262 & 3,180.2 \\
	&   400 & 11.192\% & 44,661.615  & 3,220.56  & 182 & 3,464.9 \\
	&   500 & 13.485\% & 52,170.290  & 3,634.32 & 102 & 2,657.6 \\
	&   750 & 20.609\% & 73,007.974 & 3,744.80 & 34 & 1,771.4 \\
	& 1,000 & 26.142\% & 69,823.796 & 3,768.77 & 25 & 1,334.9 \\ \midrule
	\multirow{9}{*}{\shortstack[l]{$\pi$-NeuralSEP$_2$ \\+ GC}} 
	&    50 &  3.647\%& 9,204.518 & 6.96 & 53 & 192.3  \\
	&    75 & 4.518\% & 13,152.891 &  18.40 & 74 & 475.2 \\
	&   100 & 5.930\% & 15,719.450 & 38.51 & 105 & 806.1 \\
	&   200 & 7.653\% & 21,077.597 & 310.16  & 201 & 1,752.6 \\
	&   300 & 8.683\% & 31,595.655 & 1,947.94 & 273 & 3,620.1 \\
	&   400 & \textbf{10.830\%} & \textbf{44,822.003}  & 3,443.87  & 192 & 3,824.0 \\
	&   500 & \textbf{13.349\%} & \textbf{52,255.555}  & 3,636.62 & 96 & 2,817.6 \\
	&   750 & \textbf{20.741\%} & \textbf{72,886.440} & 3,782.40 & 34 & 1,806.7 \\
	& 1,000 & \textbf{26.048\%} & \textbf{69,898.032} & 3,832.53 & 25 & 1,373.9 \\ \bottomrule
\end{tabular}

%% file: tables/random_runtime_1000.tex
\centering
\begin{tabular}{l|rrrrrr}
\toprule
Method 
& Avg. Gap ($\downarrow$) 
& Avg. LB ($\uparrow$) 
& Avg. Runtime 
& Avg. Iter. 
& Avg. Cuts
& Winning Ratio \\
\midrule
NeuralSEP$_2$ 
& 24.674\% 
& 71,209.695 
& 7,502.74 
& 34 
& 1,755.0
& -- \\

$\pi$-NeuralSEP$_2$ 
& 24.427\% 
& 71,475.696 
& 7,539.99 
& 33 
& 1,794.6
& 0.6 \\

$\pi$-NeuralSEP$_2$ + GC
& \textbf{24.226\%} 
& \textbf{71,647.285} 
& 7,503.12 
& 32 
& 1,837.8
& 0.8 \\
\bottomrule
\end{tabular}

%% file: supplementary.tex
\section*{Supplementary Tables}
\clearpage
\subsection*{CVRPSEP}
\begin{table}[!ht]
\caption{The results of CVRPSEP on randomly generated instances for $|V_C| < 300$ (1-hour limit).}
\begin{adjustbox}{max width=\textwidth}
	\input{ec_tables/random_time_cvrpsep_small}
\end{adjustbox}
\end{table}

\begin{table}[!ht]
\caption{The results of CVRPSEP on randomly generated instances for $|V_C| \geq 300$ (1-hour limit).}
\begin{adjustbox}{max width=\textwidth}
	\input{ec_tables/random_time_cvrpsep_large}
\end{adjustbox}
\end{table}

\clearpage
\subsection*{NeuralSEP$_1$}
\begin{table}[!ht]
\caption{The results of NeuralSEP$_1$ on randomly generated instances for $|V_C| < 300$ (1-hour limit).}
\begin{adjustbox}{max width=\textwidth}
	\input{ec_tables/random_time_neuralsep_small}
\end{adjustbox}
\end{table}

\begin{table}[!ht]
\caption{The results of NeuralSEP$_1$ on randomly generated instances for $|V_C| \geq 300$ (1-hour limit).}
\begin{adjustbox}{max width=\textwidth}
	\input{ec_tables/random_time_neuralsep_large}
\end{adjustbox}
\end{table}

\clearpage
\subsection*{NeuralSEP$_2$}
\begin{table}[!ht]
\caption{The results of NeuralSEP$_2$ on randomly generated instances for $|V_C| < 300$ (1-hour limit).}
\begin{adjustbox}{max width=\textwidth}
	\input{ec_tables/random_time_neuralsep2_small}
\end{adjustbox}
\end{table}

\begin{table}[!ht]
\caption{The results of NeuralSEP$_2$ on randomly generated instances for $|V_C| \geq 300$ (1-hour limit, with additional 2-hour runs for size-1000).}
\begin{adjustbox}{max width=\textwidth}
	\input{ec_tables/random_time_neuralsep2_large}
\end{adjustbox}
\end{table}

\clearpage
\subsection*{$\pi$-NeuralSEP$_2$}
\begin{table}[!ht]
\caption{The results of $\pi$-NeuralSEP$_2$ on randomly generated instances for $|V_C| < 300$ (1-hour limit).}
\begin{adjustbox}{max width=\textwidth}
	\input{ec_tables/random_time_neuralsep2e_small}
\end{adjustbox}
\end{table}

\begin{table}[!ht]
\caption{The results of $\pi$-NeuralSEP$_2$ on randomly generated instances for $|V_C| \geq 300$ (1-hour limit, with additional 2-hour runs for size-1000).}
\begin{adjustbox}{max width=\textwidth}
	\input{ec_tables/random_time_neuralsep2e_large}
\end{adjustbox}
\end{table}

\clearpage
\subsection*{$\pi$-NeuralSEP$_2$+GC}
\begin{table}[!ht]
\caption{The results of $\pi$-NeuralSEP$_2$+GC on randomly generated instances for $|V_C| < 300$ (1-hour limit).}
\begin{adjustbox}{max width=\textwidth}
	\input{ec_tables/random_time_neuralsep2e2_small}
\end{adjustbox}
\end{table}

\begin{table}[!ht]
\caption{The results of $\pi$-NeuralSEP$_2$+GC on randomly generated instances for $|V_C| \geq 300$ (1-hour limit, with additional 2-hour runs for size-1000).}
\begin{adjustbox}{max width=\textwidth}
	\input{ec_tables/random_time_neuralsep2e2_large}
\end{adjustbox}
\end{table}

%% file: ec_tables/random_time_cvrpsep_small.tex
\centering
\begin{tabular}{lrrrrrrrr}
\toprule
Name & Size & K & Best Known & LB & Gap (\%) & Runtime & Iterations & \# of Cuts \\
\midrule
random-24-X-n50 & 50 & 3 & 6,686.00 & 6,542.30 & 2.15 & 0.08 & 22 & 264 \\
random-4-X-n50 & 50 & 4 & 7,684.00 & 7,579.00 & 1.37 & 0.09 & 18 & 321 \\
random-28-X-n50 & 50 & 7 & 11,557.00 & 11,081.25 & 4.12 & 0.35 & 30 & 1,342 \\
random-0-X-n50 & 50 & 8 & 12,735.00 & 12,237.23 & 3.91 & 0.25 & 26 & 1,076 \\
random-16-X-n50 & 50 & 5 & 10,190.00 & 10,091.62 & 0.97 & 0.23 & 26 & 1,127 \\
random-20-X-n50 & 50 & 6 & 10,126.00 & 9,929.22 & 1.94 & 0.18 & 20 & 832 \\
random-32-X-n50 & 50 & 4 & 7,483.00 & 7,395.72 & 1.17 & 0.12 & 20 & 611 \\
random-12-X-n50 & 50 & 4 & 8,317.00 & 8,242.66 & 0.89 & 0.16 & 26 & 707 \\
random-36-X-n50 & 50 & 8 & 14,188.00 & 13,865.18 & 2.28 & 0.70 & 32 & 1,827 \\
random-8-X-n50 & 50 & 3 & 6,733.00 & 6,672.50 & 0.90 & 0.06 & 20 & 169 \\
random-17-X-n75 & 75 & 20 & 34,431.00 & 33,180.15 & 3.63 & 8.03 & 86 & 3,811 \\
random-21-X-n75 & 75 & 7 & 9,726.00 & 9,566.20 & 1.64 & 0.99 & 27 & 1,393 \\
random-25-X-n75 & 75 & 6 & 10,931.00 & 10,665.52 & 2.43 & 1.35 & 32 & 1,565 \\
random-5-X-n75 & 75 & 11 & 14,930.00 & 14,213.84 & 4.80 & 2.68 & 58 & 2,296 \\
random-9-X-n75 & 75 & 5 & 9,361.00 & 9,132.51 & 2.44 & 0.79 & 23 & 1,067 \\
random-29-X-n75 & 75 & 7 & 11,893.00 & 11,509.30 & 3.23 & 1.27 & 26 & 1,529 \\
random-1-X-n75 & 75 & 7 & 11,225.00 & 10,881.53 & 3.06 & 1.18 & 33 & 1,385 \\
random-13-X-n75 & 75 & 6 & 9,652.00 & 9,568.79 & 0.86 & 0.88 & 29 & 1,056 \\
random-37-X-n75 & 75 & 4 & 9,479.00 & 9,323.34 & 1.64 & 0.79 & 22 & 1,090 \\
random-33-X-n75 & 75 & 13 & 16,151.00 & 15,510.96 & 3.96 & 2.15 & 35 & 2,087 \\
random-2-X-n100 & 100 & 11 & 17,242.00 & 16,296.51 & 5.48 & 6.63 & 51 & 3,318 \\
random-30-X-n100 & 100 & 6 & 12,151.00 & 11,598.80 & 4.54 & 2.73 & 38 & 2,149 \\
random-14-X-n100 & 100 & 12 & 16,934.00 & 15,824.60 & 6.55 & 7.45 & 51 & 3,403 \\
random-6-X-n100 & 100 & 14 & 24,832.00 & 23,841.98 & 3.99 & 16.11 & 78 & 4,738 \\
random-34-X-n100 & 100 & 8 & 10,752.00 & 10,487.62 & 2.46 & 1.20 & 27 & 1,267 \\
random-26-X-n100 & 100 & 11 & 15,654.00 & 14,898.11 & 4.83 & 8.05 & 55 & 3,526 \\
random-18-X-n100 & 100 & 7 & 11,171.00 & 10,924.95 & 2.20 & 1.62 & 27 & 1,645 \\
random-22-X-n100 & 100 & 20 & 27,022.00 & 25,565.45 & 5.39 & 19.97 & 103 & 5,076 \\
random-38-X-n100 & 100 & 11 & 13,766.00 & 13,057.19 & 5.15 & 2.99 & 43 & 2,276 \\
random-10-X-n100 & 100 & 12 & 17,727.00 & 16,875.42 & 4.80 & 10.22 & 54 & 3,909 \\
random-19-X-n200 & 200 & 12 & 16,102.00 & 15,348.60 & 4.68 & 38.92 & 60 & 4,548 \\
random-27-X-n200 & 200 & 13 & 21,609.00 & 20,264.96 & 6.22 & 160.97 & 100 & 8,022 \\
random-23-X-n200 & 200 & 23 & 26,556.00 & 24,752.28 & 6.79 & 361.14 & 162 & 11,135 \\
random-7-X-n200 & 200 & 13 & 17,182.00 & 16,310.35 & 5.07 & 55.04 & 66 & 5,143 \\
random-11-X-n200 & 200 & 20 & 29,495.00 & 27,328.95 & 7.34 & 407.48 & 184 & 11,141 \\
random-35-X-n200 & 200 & 15 & 24,854.00 & 23,172.06 & 6.77 & 252.98 & 129 & 9,983 \\
random-15-X-n200 & 200 & 18 & 23,121.00 & 21,456.95 & 7.20 & 302.04 & 141 & 10,513 \\
random-3-X-n200 & 200 & 13 & 19,644.00 & 18,397.19 & 6.35 & 115.27 & 91 & 6,988 \\
random-31-X-n200 & 200 & 21 & 31,684.00 & 29,460.20 & 7.02 & 604.30 & 168 & 13,133 \\
random-39-X-n200 & 200 & 11 & 18,271.00 & 17,291.34 & 5.36 & 82.55 & 87 & 5,789 \\
random-7-X-n300 & 300 & 19 & 27,269.00 & 25,097.76 & 7.96 & 2,487.46 & 220 & 18,414 \\
random-1-X-n300 & 300 & 25 & 30,181.00 & 27,775.74 & 7.97 & 2,846.47 & 295 & 20,539 \\
random-5-X-n300 & 300 & 40 & 53,305.00 & 49,066.58 & 7.95 & 3,609.15 & 229 & 19,784 \\
random-0-X-n300 & 300 & 46 & 55,614.00 & 49,861.37 & 10.34 & 3,626.83 & 289 & 19,940 \\
random-8-X-n300 & 300 & 15 & 20,636.00 & 19,245.87 & 6.74 & 492.04 & 116 & 9,288 \\
random-6-X-n300 & 300 & 40 & 42,079.00 & 38,019.89 & 9.65 & 3,614.36 & 257 & 21,695 \\
random-4-X-n300 & 300 & 20 & 24,324.00 & 22,421.26 & 7.82 & 841.67 & 146 & 11,289 \\
random-2-X-n300 & 300 & 32 & 47,765.00 & 43,824.10 & 8.25 & 3,602.01 & 319 & 21,613 \\
random-3-X-n300 & 300 & 20 & 23,414.00 & 21,695.47 & 7.34 & 912.24 & 165 & 12,575 \\
random-9-X-n300 & 300 & 17 & 21,740.00 & 20,650.22 & 5.01 & 602.98 & 132 & 10,474 \\
\bottomrule
\end{tabular}

%% file: ec_tables/random_time_cvrpsep_large.tex
\centering
\begin{tabular}{lrrrrrrrr}
\toprule
Name & Size & K & Best Known & LB & Gap (\%) & Runtime & Iterations & \# of Cuts \\
\midrule
random-14-X-n400 & 400 & 46 & 61,276.00 & 52,578.22 & 14.19 & 3,649.77 & 206 & 13,875 \\
random-6-X-n400 & 400 & 54 & 52,025.00 & 45,155.96 & 13.20 & 3,631.42 & 293 & 14,235 \\
random-10-X-n400 & 400 & 47 & 59,487.00 & 51,314.70 & 13.74 & 3,607.06 & 197 & 14,069 \\
random-2-X-n400 & 400 & 42 & 61,447.00 & 52,654.09 & 14.31 & 3,611.31 & 196 & 13,643 \\
random-12-X-n400 & 400 & 30 & 37,545.00 & 33,017.38 & 12.06 & 3,645.31 & 181 & 14,525 \\
random-0-X-n400 & 400 & 61 & 77,967.00 & 66,029.98 & 15.31 & 3602.29 & 248 & 13,525 \\
random-16-X-n400 & 400 & 38 & 57,273.00 & 48,929.25 & 14.57 & 3,621.66 & 198 & 14,189 \\
random-18-X-n400 & 400 & 27 & 34,643.00 & 30,828.48 & 11.01 & 3,625.78 & 191 & 15,724 \\
random-4-X-n400 & 400 & 27 & 29,917.00 & 27,502.56 & 8.07 & 3,607.34 & 241 & 17,197 \\
random-8-X-n400 & 400 & 20 & 30,953.00 & 27,944.45 & 9.72 & 3,642.02 & 186 & 15,394 \\
random-17-X-n500 & 500 & 128 & 134,831.00 & 110,284.82 & 18.21 & 3,648.99 & 457 & 7,144 \\
random-15-X-n500 & 500 & 44 & 48,692.00 & 40,439.59 & 16.95 & 3,677.66 & 184 & 12,012 \\
random-19-X-n500 & 500 & 28 & 34,861.00 & 30,453.95 & 12.64 & 3,612.47 & 174 & 13,942 \\
random-1-X-n500 & 500 & 41 & 40,287.00 & 34,932.19 & 13.29 & 3,632.32 & 178 & 12,460 \\
random-9-X-n500 & 500 & 29 & 43,607.00 & 36,584.83 & 16.10 & 3,611.85 & 174 & 11,068 \\
random-13-X-n500 & 500 & 36 & 46,414.00 & 39,615.25 & 14.65 & 3,639.40 & 138 & 11,152 \\
random-5-X-n500 & 500 & 67 & 85,332.00 & 65,745.72 & 22.95 & 3,622.79 & 320 & 9,261 \\
random-11-X-n500 & 500 & 48 & 66,720.00 & 55,566.90 & 16.72 & 3,604.13 & 182 & 10,769 \\
random-3-X-n500 & 500 & 33 & 56,591.00 & 47,446.22 & 16.16 & 3,607.24 & 169 & 11,409 \\
random-7-X-n500 & 500 & 32 & 48,159.00 & 40,501.99 & 15.90 & 3,609.75 & 168 & 11,549 \\
random-6-X-n750 & 750 & 99 & 137,429.00 & 102,191.35 & 25.64 & 3,600.82 & 165 & 5,299 \\
random-12-X-n750 & 750 & 56 & 81,988.00 & 61,289.91 & 25.25 & 3,684.36 & 154 & 6,605 \\
random-4-X-n750 & 750 & 49 & 55,938.00 & 43,546.52 & 22.15 & 3,643.33 & 145 & 7,229 \\
random-14-X-n750 & 750 & 87 & 98,971.00 & 70,107.53 & 29.16 & 3,658.21 & 290 & 3,671 \\
random-8-X-n750 & 750 & 37 & 54,145.00 & 40,990.23 & 24.30 & 3,689.29 & 181 & 6,417 \\
random-2-X-n750 & 750 & 79 & 76,471.00 & 57,697.64 & 24.55 & 3,638.23 & 246 & 3,816 \\
random-16-X-n750 & 750 & 72 & 77,111.00 & 57,780.96 & 25.07 & 3,646.25 & 255 & 4,171 \\
random-0-X-n750 & 750 & 115 & 137,798.00 & 102,577.03 & 25.56 & 3,658.94 & 321 & 2,981 \\
random-18-X-n750 & 750 & 50 & 85,348.00 & 63,779.22 & 25.27 & 3,646.60 & 158 & 6,825 \\
random-10-X-n750 & 750 & 88 & 119,393.00 & 82,525.35 & 30.88 & 3,659.94 & 282 & 3,400 \\
random-5-X-n1000 & 1,000 & 133 & 130,385.00 & 88,205.22 & 32.35 & 3,603.63 & 200 & 1,970 \\
random-1-X-n1000 & 1,000 & 81 & 86,706.00 & 63,027.81 & 27.31 & 3,610.88 & 127 & 4,155 \\
random-7-X-n1000 & 1,000 & 64 & 81,052.00 & 54,515.83 & 32.74 & 3,733.41 & 211 & 2,869 \\
random-3-X-n1000 & 1,000 & 65 & 76,005.00 & 55,404.21 & 27.10 & 3,743.25 & 121 & 4,202 \\
random-9-X-n1000 & 1,000 & 57 & 81,448.00 & 57,267.57 & 29.69 & 3,702.48 & 152 & 4,309 \\
random-21-X-n1000 & 1,000 & 81 & 109,492.00 & 75,263.75 & 31.26 & 3,623.13 & 123 & 3,967 \\
random-19-X-n1000 & 1,000 & 56 & 73,749.00 & 52,393.52 & 28.96 & 3,620.41 & 131 & 4,271 \\
random-13-X-n1000 & 1,000 & 72 & 92,888.00 & 63,627.01 & 31.50 & 3,647.58 & 141 & 3,958 \\
random-15-X-n1000 & 1,000 & 88 & 112,383.00 & 76,238.26 & 32.16 & 3,705.68 & 130 & 3,650 \\
random-11-X-n1000 & 1,000 & 96 & 102,522.00 & 70,728.38 & 31.01 & 3,622.97 & 231 & 2,034 \\
\bottomrule
\end{tabular}

%% file: ec_tables/random_time_neuralsep_small.tex
\centering
\begin{tabular}{lrrrrrrrr}
\toprule
Name & Size & K & Best Known & LB & Gap (\%) & Runtime & Iterations & \# of Cuts \\
\midrule
random-24-X-n50 & 50 & 3 &  6,686.00  &  6,367.50  &  4.76  &  22.30  &  25 & 56\\
random-4-X-n50 & 50 & 4 &  7,684.00  &  7,374.00  &  4.03  &  25.26  &  30 & 86\\
random-28-X-n50 & 50 & 7 &  11,557.00  &  10,859.83  &  6.03  &  34.87  &  34 & 186\\
random-0-X-n50 & 50 & 8 &  12,735.00  &  11,939.07  &  6.25  &  48.94  &  49 & 234\\
random-16-X-n50 & 50 & 5 &  10,190.00  &  9,811.46  &  3.71  &  34.60  &  43 & 163\\
random-20-X-n50 & 50 & 6 &  10,126.00  &  9,751.24  &  3.70  &  35.43  &  48 & 201\\
random-32-X-n50 & 50 & 4 &  7,483.00  &  7,231.50  &  3.36  &  23.65  &  38 & 114\\
random-12-X-n50 & 50 & 4 &  8,317.00  &  8,032.14  &  3.43  &  19.23  &  26 & 81\\
random-36-X-n50 & 50 & 8 &  14,188.00  &  13,506.03  &  4.81  &  57.79  &  64 & 350\\
random-8-X-n50 & 50 & 3 &  6,733.00  &  6,637.00  &  1.43  &  13.34  &  24 & 55\\
random-17-X-n75 & 75 & 20 &  34,431.00  &  32,940.30  &  4.33  &  169.74  &  128 & 1,343\\
random-21-X-n75 & 75 & 7 &  9,726.00  &  9,249.55  &  4.90  &  43.31  &  45 & 233\\
random-25-X-n75 & 75 & 6 &  10,931.00  &  10,153.06  &  7.12  &  44.04  &  46 & 196\\
random-5-X-n75 & 75 & 11 &  14,930.00  &  13,875.14  &  7.07  &  64.81  &  60 & 440\\
random-9-X-n75 & 75 & 5 &  9,361.00  &  8,894.51  &  4.98  &  34.03  &  36 & 124\\
random-29-X-n75 & 75 & 7 &  11,893.00  &  11,198.49  &  5.84  &  56.05  &  61 & 278\\
random-1-X-n75 & 75 & 7 &  11,225.00  &  10,521.71  &  6.27  &  51.03  &  46 & 215\\
random-13-X-n75 & 75 & 6 &  9,652.00  &  9,377.54  &  2.84  &  45.33  &  53 & 198\\
random-37-X-n75 & 75 & 4 &  9,479.00  &  9,092.72  &  4.08  &  33.86  &  37 & 110\\
random-33-X-n75 & 75 & 13 &  16,151.00  &  15,243.04  &  5.62  &  118.70  &  106 & 725\\
random-2-X-n100 & 100 & 11 &  17,242.00  &  15,890.60  &  7.84  &  157.49  &  116 & 703\\
random-30-X-n100 & 100 & 6 &  12,151.00  &  11,323.09  &  6.81  &  85.60  &  67 & 250\\
random-14-X-n100 & 100 & 12 &  16,934.00  &  15,420.52  &  8.94  &  142.67  &  102 & 622\\
random-6-X-n100 & 100 & 14 &  24,832.00  &  23,431.00  &  5.64  &  207.24  &  121 & 962\\
random-34-X-n100 & 100 & 8 &  10,752.00  &  10,293.71  &  4.26  &  71.62  &  59 & 310\\
random-26-X-n100 & 100 & 11 &  15,654.00  &  14,428.86  &  7.83  &  146.08  &  99 & 676\\
random-18-X-n100 & 100 & 7 &  11,171.00  &  10,727.01  &  3.97  &  77.98  &  70 & 301\\
random-22-X-n100 & 100 & 20 &  27,022.00  &  25,202.86  &  6.73  &  185.48  &  99 & 1,315\\
random-38-X-n100 & 100 & 11 &  13,766.00  &  12,777.25  &  7.18  &  124.80  &  96 & 559\\
random-10-X-n100 & 100 & 12 &  17,727.00  &  16,501.72  &  6.91  &  154.73  &  111 & 730\\
random-19-X-n200 & 200 & 12 &  16,102.00  &  14,934.30  &  7.25  &  207.11  &  82 & 560\\
random-27-X-n200 & 200 & 13 &  21,609.00  &  19,392.85  &  10.26  &  326.37  &  111 & 941\\
random-23-X-n200 & 200 & 23 &  26,556.00  &  24,037.70  &  9.48  &  561.70  &  130 & 1,878\\
random-7-X-n200 & 200 & 13 &  17,182.00  &  15,803.73  &  8.02  &  276.49  &  101 & 827\\
random-11-X-n200 & 200 & 20 &  29,495.00  &  26,858.92  &  8.94  &  755.13  &  196 & 1,903\\
random-35-X-n200 & 200 & 15 &  24,854.00  &  22,400.15  &  9.87  &  395.79  &  117 & 1,158\\
random-15-X-n200 & 200 & 18 &  23,121.00  &  20,934.12  &  9.46  &  522.33  &  147 & 1,520\\
random-3-X-n200 & 200 & 13 &  19,644.00  &  17,695.74  &  9.92  &  317.63  &  114 & 882\\
random-31-X-n200 & 200 & 21 &  31,684.00  &  28,620.28  &  9.67  &  630.10  &  149 & 1,959\\
random-39-X-n200 & 200 & 11 &  18,271.00  &  16,577.11  &  9.27  &  202.64  &  78 & 595\\
random-7-X-n300 & 300 & 19 &  27,269.00  &  24,113.72  &  11.57  &  1,128.53  &  162 & 1,728\\
random-1-X-n300 & 300 & 25 &  30,181.00  &  26,659.81  &  11.67  &  1,429.16  &  171 & 2,249\\
random-5-X-n300 & 300 & 40 &  53,305.00  &  48,355.96  &  9.28  &  3,331.39  &  180 & 4,593\\
random-0-X-n300 & 300 & 46 &  55,614.00  &  50,040.31  &  10.02  &  3,601.51  &  156 & 5,252\\
random-8-X-n300 & 300 & 15 &  20,636.00  &  18,463.91  &  10.53  &  563.72  &  115 & 970\\
random-6-X-n300 & 300 & 40 &  42,079.00  &  37,695.35  &  10.42  &  3,376.16  &  203 & 4,749\\
random-4-X-n300 & 300 & 20 &  24,324.00  &  21,530.95  &  11.48  &  710.27  &  114 & 1,366\\
random-2-X-n300 & 300 & 32 &  47,765.00  &  43,252.23  &  9.45  &  2,832.43  &  206 & 3,638\\
random-3-X-n300 & 300 & 20 &  23,414.00  &  20,786.15  &  11.22  &  596.39  &  96 & 1,303\\
random-9-X-n300 & 300 & 17 &  21,740.00  &  19,673.06  &  9.51  &  560.01  &  107 & 1,030\\
\bottomrule
\end{tabular}

%% file: ec_tables/random_time_neuralsep_large.tex
\centering
\begin{tabular}{lrrrrrrrr}
\toprule
Name & Size & K & Best Known & LB & Gap (\%) & Runtime & Iterations  & \# of Cuts \\
\midrule
random-14-X-n400 & 400 & 46 &  61,276.00  &  53,510.30  &  12.67  &  3,628.85  &  95 & 3,811\\
random-6-X-n400 & 400 & 54 &  52,025.00  &  45,679.61  &  12.20  &  3,602.94  &  91 & 4,145\\
random-10-X-n400 & 400 & 47 &  59,487.00  &  52,061.38  &  12.48  &  3,643.68  &  95 & 3,953\\
random-2-X-n400 & 400 & 42 &  61,447.00  &  53,536.88  &  12.87  &  3,629.33  &  102 & 3,719\\
random-12-X-n400 & 400 & 30 &  37,545.00  &  32,427.73  &  13.63  &  3,003.17  &  167 & 2,671\\
random-0-X-n400 & 400 & 61 &  77,967.00  &  68,270.62  &  12.44  &  3,650.81  &  76 & 4,318\\
random-16-X-n400 & 400 & 38 &  57,273.00  &  49,628.87  &  13.35  &  3,618.76  &  110 & 3,442\\
random-18-X-n400 & 400 & 27 &  34,643.00  &  30,119.55  &  13.06  &  2,508.70  &  146 & 2,501\\
random-4-X-n400 & 400 & 27 &  29,917.00  &  26,366.19  &  11.87  &  1,403.56  &  112 & 2,048\\
random-8-X-n400 & 400 & 20 &  30,953.00  &  26,646.58  &  13.91  &  1,497.17  &  135 & 1,602\\
random-17-X-n500 & 500 & 128 &  134,831.00  &  109,742.21  &  18.61  &  3,658.02  &  36 & 4,415\\
random-15-X-n500 & 500 & 44 &  48,692.00  &  41,152.55  &  15.48  &  3,635.38  &  75 & 3,001\\
random-19-X-n500 & 500 & 28 &  34,861.00  &  30,123.59  &  13.59  &  3,604.13  &  156 & 2,412\\
random-1-X-n500 & 500 & 41 &  40,287.00  &  34,795.29  &  13.63  &  3,606.54  &  95 & 2,874\\
random-9-X-n500 & 500 & 29 &  43,607.00  &  36,798.90  &  15.61  &  3,607.19  &  111 & 2,395\\
random-13-X-n500 & 500 & 36 &  46,414.00  &  39,838.06  &  14.17  &  3,632.87  &  92 & 2,772\\
random-5-X-n500 & 500 & 67 &  85,332.00  &  69,914.03  &  18.07  &  3,657.28  &  57 & 3,509\\
random-11-X-n500 & 500 & 48 &  66,720.00  &  56,557.60  &  15.23  &  3,616.01  &  69 & 2,966\\
random-3-X-n500 & 500 & 33 &  56,591.00  &  48,132.13  &  14.95  &  3,654.21  &  94 & 2,449\\
random-7-X-n500 & 500 & 32 &  48,159.00  &  41,038.66  &  14.79  &  3,608.49  &  99 & 2,530\\
random-6-X-n750 & 750 & 99 &  137,429.00  &  102,825.85  &  25.18  &  3,607.95  &  28 & 2,673\\
random-12-X-n750 & 750 & 56 &  81,988.00  &  63,662.98  &  22.35  &  3,708.41  &  40 & 2,162\\
random-4-X-n750 & 750 & 49 &  55,938.00  &  44,895.96  &  19.74  &  3,708.21  &  44 & 2,071\\
random-14-X-n750 & 750 & 87 &  98,971.00  &  74,391.39  &  24.84  &  3,830.51  &  31 & 2,575\\
random-8-X-n750 & 750 & 37 &  54,145.00  &  43,193.78  &  20.23  &  3,719.64  &  53 & 1,845\\
random-2-X-n750 & 750 & 79 &  76,471.00  &  60,104.72  &  21.40  &  3,752.86  &  32 & 2,404\\
random-16-X-n750 & 750 & 72 &  77,111.00  &  59,956.41  &  22.25  &  3,706.97  &  33 & 2,251\\
random-0-X-n750 & 750 & 115 &  137,798.00  &  104,900.94  &  23.87  &  3,603.26  &  25 & 2,736 \\
random-18-X-n750 & 750 & 50 &  85,348.00  &  67,142.83  &  21.33  &  3,603.85  &  45 & 2,162 \\
random-10-X-n750 & 750 & 88 &  119,393.00  &  90,330.72  &  24.34  &  3,702.66  &  32 & 2,697\\
random-5-X-n1000 & 1000 & 133 &  130,385.00  &  91,445.63  &  29.86  &  3,765.14  &  15 & 1,824\\
random-1-X-n1000 & 1000 & 81 &  86,706.00  &  62,069.80  &  28.41  &  3,654.84  &  22 & 1,694\\
random-7-X-n1000 & 1000 & 64 &  81,052.00  &  57,933.21  &  28.52  &  3,705.91  &  26 & 1,571\\
random-3-X-n1000 & 1000 & 65 &  76,005.00  &  55,377.02  &  27.14  &  3,700.91  &  25 & 1,557\\
random-9-X-n1000 & 1000 & 57 &  81,448.00  &  59,559.02  &  26.87  &  3,608.10  &  27 & 1,481\\
random-21-X-n1000 & 1000 & 81 &  109,492.00  &  77,038.38  &  29.64  &  3,882.41  &  22 & 1,701\\
random-19-X-n1000 & 1000 & 56 &  73,749.00  &  52,836.71  &  28.36  &  3,675.17  &  29 & 1,567\\
random-13-X-n1000 & 1000 & 72 &  92,888.00  &  64,965.06  &  30.06  &  3,826.23  &  25 & 1,728\\
random-15-X-n1000 & 1000 & 88 &  112,383.00  &  78,649.96  &  30.02  &  3,617.78  &  21 & 1,760\\
random-11-X-n1000 & 1000 & 96 &  102,522.00  &  72,906.08  &  28.89  &  3,635.56  &  19 & 1,701\\

\bottomrule
\end{tabular}

%% file: ec_tables/random_time_neuralsep2_small.tex
\centering
\begin{tabular}{lrrrrrrrr}
\toprule
Name & Size & K & Best Known & LB & Gap (\%) & Runtime & Iterations & \# of Cuts\\
\midrule
random-24-X-n50 & 50 & 3 & 6,686.00 & 6,068.00 & 1.74 & 9.81 & 23 & 35 \\
random-4-X-n50 & 50 & 4 & 7,684.00 & 7,342.20 & 4.45 & 2.62 & 34 & 66 \\
random-28-X-n50 & 50 & 7 & 11,557.00 & 10,858.28 & 6.05 & 5.78 & 60 & 215 \\
random-0-X-n50 & 50 & 8 & 12,735.00 & 11,835.38 & 7.06 & 5.98 & 59 & 279 \\
random-16-X-n50 & 50 & 5 & 10,190.00 & 9,780.67 & 4.02 & 3.52 & 42 & 137 \\
random-20-X-n50 & 50 & 6 & 10,126.00 & 9,572.14 & 5.47 & 3.54 & 40 & 157 \\
random-32-X-n50 & 50 & 4 & 7,483.00 & 7,189.00 & 3.93 & 1.98 & 27 & 61 \\
random-12-X-n50 & 50 & 4 & 8,317.00 & 7,985.00 & 3.99 & 1.74 & 22 & 61 \\
random-36-X-n50 & 50 & 8 & 14,188.00 & 13,370.56 & 5.76 & 5.34 & 52 & 263 \\
random-8-X-n50 & 50 & 3 & 6,733.00 & 6,562.67 & 2.53 & 1.56 & 21 & 35 \\
random-17-X-n75 & 75 & 20 & 34,431.00 & 32,559.32 & 5.44 & 30.57 & 106 & 1,190 \\
random-21-X-n75 & 75 & 7 & 9,726.00 & 9,207.64 & 5.33 & 6.84 & 49 & 198 \\
random-25-X-n75 & 75 & 6 & 10,931.00 & 10,318.87 & 5.60 & 7.82 & 58 & 217 \\
random-5-X-n75 & 75 & 11 & 14,930.00 & 13,909.44 & 6.84 & 14.92 & 83 & 493 \\
random-9-X-n75 & 75 & 5 & 9,361.00 & 8,887.00 & 5.06 & 5.04 & 41 & 140 \\
random-29-X-n75 & 75 & 7 & 11,893.00 & 11,178.43 & 6.01 & 8.40 & 58 & 262 \\
random-1-X-n75 & 75 & 7 & 11,225.00 & 10,494.67 & 6.51 & 8.29 & 59 & 223 \\
random-13-X-n75 & 75 & 6 & 9,652.00 & 9,443.14 & 2.16 & 8.32 & 62 & 210 \\
random-37-X-n75 & 75 & 4 & 9,479.00 & 9,061.69 & 4.40 & 4.77 & 43 & 112 \\
random-33-X-n75 & 75 & 13 & 16,151.00 & 15,191.15 & 5.94 & 23.11 & 109 & 705 \\
random-2-X-n100 & 100 & 11 & 17,242.00 & 15,960.31 & 7.43 & 28.34 & 101 & 722 \\
random-30-X-n100 & 100 & 6 & 12,151.00 & 11,211.33 & 7.73 & 12.11 & 63 & 227 \\
random-14-X-n100 & 100 & 12 & 16,934.00 & 15,526.29 & 8.31 & 34.69 & 115 & 802 \\
random-6-X-n100 & 100 & 14 & 24,832.00 & 23,440.68 & 5.60 & 54.68 & 156 & 1,177 \\
random-34-X-n100 & 100 & 8 & 10,752.00 & 10,383.55 & 3.43 & 14.88 & 69 & 367 \\
random-26-X-n100 & 100 & 11 & 15,654.00 & 14,483.65 & 7.48 & 31.72 & 113 & 733 \\
random-18-X-n100 & 100 & 7 & 11,171.00 & 10,697.08 & 4.24 & 18.52 & 77 & 327 \\
random-22-X-n100 & 100 & 20 & 27,022.00 & 25,217.70 & 6.68 & 70.89 & 151 & 1,604 \\
random-38-X-n100 & 100 & 11 & 13,766.00 & 12,799.40 & 7.02 & 21.71 & 82 & 572 \\
random-10-X-n100 & 100 & 12 & 17,727.00 & 16,576.48 & 6.49 & 32.21 & 109 & 782 \\
random-19-X-n200 & 200 & 12 & 16,102.00 & 15,099.00 & 6.23 & 94.79 & 124 & 716 \\
random-27-X-n200 & 200 & 13 & 21,609.00 & 19,820.09 & 8.28 & 160.40 & 161 & 1,264 \\
random-23-X-n200 & 200 & 23 & 26,556.00 & 24,181.46 & 8.94 & 272.02 & 173 & 2,090 \\
random-7-X-n200 & 200 & 13 & 17,182.00 & 16,049.15 & 6.59 & 133.65 & 161 & 1,039 \\
random-11-X-n200 & 200 & 20 & 29,495.00 & 26,674.13 & 9.56 & 222.38 & 152 & 1,697 \\
random-35-X-n200 & 200 & 15 & 24,854.00 & 22,685.67 & 8.72 & 205.49 & 178 & 1,430 \\
random-15-X-n200 & 200 & 18 & 23,121.00 & 21,033.09 & 9.03 & 221.72 & 166 & 1,696 \\
random-3-X-n200 & 200 & 13 & 19,644.00 & 17,997.53 & 8.38 & 127.21 & 136 & 1,095 \\
random-31-X-n200 & 200 & 21 & 31,684.00 & 28,993.17 & 8.49 & 385.77 & 216 & 2,554 \\
random-39-X-n200 & 200 & 11 & 18,271.00 & 16,783.69 & 8.14 & 91.80 & 119 & 824 \\
random-7-X-n300 & 300 & 19 & 27,269.00 & 24,520.43 & 10.08 & 1,161.17 & 343 & 2,815 \\
random-1-X-n300 & 300 & 25 & 30,181.00 & 26,782.76 & 11.26 & 889.74 & 215 & 2,447 \\
random-5-X-n300 & 300 & 40 & 53,305.00 & 47,474.61 & 10.94 & 1,069.59 & 157 & 2,482 \\
random-0-X-n300 & 300 & 46 & 55,614.00 & 49,210.04 & 11.52 & 1,526.06 & 183 & 3,129 \\
random-8-X-n300 & 300 & 15 & 20,636.00 & 18,861.82 & 8.60 & 317.94 & 166 & 1,366 \\
random-6-X-n300 & 300 & 40 & 42,079.00 & 36,593.30 & 13.04 & 879.58 & 117 & 2,239 \\
random-4-X-n300 & 300 & 20 & 24,324.00 & 21,874.55 & 10.07 & 412.35 & 151 & 1,702 \\
random-2-X-n300 & 300 & 32 & 47,765.00 & 42,747.91 & 10.50 & 1,388.27 & 206 & 2,734 \\
random-3-X-n300 & 300 & 20 & 23,414.00 & 21,300.04 & 9.03 & 658.66 & 233 & 2,174 \\
random-9-X-n300 & 300 & 17 & 21,740.00 & 20,185.05 & 7.15 & 385.01 & 177 & 1,621 \\ \bottomrule
\end{tabular}

%% file: ec_tables/random_time_neuralsep2_large.tex
\centering
\begin{tabular}{lrrrrrrrr}
\toprule
Name & Size & K & Best Known & LB & Gap (\%) & Runtime & Iterations & \# of Cuts\\
\midrule
random-14-X-n400 & 400 & 46 & 61,276.00 & 51,871.09 & 15.35 & 1,367.79 & 91 & 1,840 \\
random-6-X-n400 & 400 & 54 & 52,025.00 & 45,392.57 & 12.75 & 2,617.61 & 171 & 3,223 \\
random-10-X-n400 & 400 & 47 & 59,487.00 & 50,869.35 & 14.49 & 1,796.49 & 96 & 2,132 \\
random-2-X-n400 & 400 & 42 & 61,447.00 & 51,997.72 & 15.38 & 1,517.66 & 92 & 1,863 \\
random-12-X-n400 & 400 & 30 & 37,545.00 & 32,144.67 & 14.38 & 2,099.95 & 218 & 2,160 \\
random-0-X-n400 & 400 & 61 & 77,967.00 & 68,988.70 & 11.52 & 3,604.01 & 131 & 4,204 \\
random-16-X-n400 & 400 & 38 & 57,273.00 & 48,704.72 & 14.96 & 1,551.29 & 112 & 1,702 \\
random-18-X-n400 & 400 & 27 & 34,643.00 & 29,842.83 & 13.86 & 1,238.89 & 144 & 1,783 \\
random-4-X-n400 & 400 & 27 & 29,917.00 & 26,446.02 & 11.60 & 1,286.12 & 219 & 2,350 \\
random-8-X-n400 & 400 & 20 & 30,953.00 & 26,752.43 & 13.57 & 1,124.71 & 180 & 1,719 \\
random-17-X-n500 & 500 & 128 & 134,831.00 & 114,365.44 & 15.18 & 3,757.21 & 47 & 3,712 \\
random-15-X-n500 & 500 & 44 & 48,692.00 & 41,207.34 & 15.37 & 2,733.40 & 103 & 1,920 \\
random-19-X-n500 & 500 & 28 & 34,861.00 & 30,317.58 & 13.03 & 3,604.79 & 263 & 2,240 \\
random-1-X-n500 & 500 & 41 & 40,287.00 & 34,608.09 & 14.10 & 2,242.87 & 115 & 1,983 \\
random-9-X-n500 & 500 & 29 & 43,607.00 & 37,060.11 & 15.01 & 3,605.82 & 179 & 1,937 \\
random-13-X-n500 & 500 & 36 & 46,414.00 & 39,387.74 & 15.14 & 2,271.32 & 106 & 1,495 \\
random-5-X-n500 & 500 & 67 & 85,332.00 & 71,623.18 & 16.07 & 3,662.16 & 70 & 2,707 \\
random-11-X-n500 & 500 & 48 & 66,720.00 & 56,535.31 & 15.26 & 3,608.10 & 109 & 2,024 \\
random-3-X-n500 & 500 & 33 & 56,591.00 & 48,226.99 & 14.78 & 3,609.89 & 148 & 1,807 \\
random-7-X-n500 & 500 & 32 & 48,159.00 & 41,115.92 & 14.62 & 3,608.47 & 154 & 1,868 \\
random-6-X-n750 & 750 & 99 & 137,429.00 & 104,718.78 & 23.80 & 3,831.86 & 36 & 1,945 \\
random-12-X-n750 & 750 & 56 & 81,988.00 & 64,078.76 & 21.84 & 3,685.60 & 44 & 1,393 \\
random-4-X-n750 & 750 & 49 & 55,938.00 & 44,197.62 & 20.99 & 3,667.28 & 50 & 1,450 \\
random-14-X-n750 & 750 & 87 & 98,971.00 & 76,862.99 & 22.34 & 3,799.39 & 33 & 1,882 \\
random-8-X-n750 & 750 & 37 & 54,145.00 & 42,537.51 & 21.44 & 3,603.11 & 63 & 1,139 \\
random-2-X-n750 & 750 & 79 & 76,471.00 & 60,617.63 & 20.73 & 3,754.86 & 36 & 2,057 \\
random-16-X-n750 & 750 & 72 & 77,111.00 & 60,910.85 & 21.01 & 3,720.22 & 38 & 1,836 \\
random-0-X-n750 & 750 & 115 & 137,798.00 & 105,912.64 & 23.14 & 3,839.84 & 32 & 1,963 \\
random-18-X-n750 & 750 & 50 & 85,348.00 & 66,786.09 & 21.75 & 3,668.55 & 52 & 1,277 \\
random-10-X-n750 & 750 & 88 & 119,393.00 & 90,083.32 & 24.55 & 3,724.57 & 34 & 1,840 \\
random-5-X-n1000 & 1,000 & 133 & 130,385.00 & 93,723.67 & 28.12 & 3,770.55 & 24 & 1,487 \\
random-1-X-n1000 & 1,000 & 81 & 86,706.00 & 65,178.11 & 24.83 & 3,615.80 & 25 & 1,213 \\
random-7-X-n1000 & 1,000 & 64 & 81,052.00 & 60,251.09 & 25.66 & 3,762.54 & 29 & 1,251 \\
random-3-X-n1000 & 1,000 & 65 & 76,005.00 & 56,470.01 & 25.70 & 3,871.65 & 28 & 1,306 \\
random-9-X-n1000 & 1,000 & 57 & 81,448.00 & 59,863.27 & 26.50 & 3,807.76 & 30 & 1,210 \\
random-21-X-n1000 & 1,000 & 81 & 109,492.00 & 79,300.21 & 27.57 & 3,782.12 & 25 & 1,222 \\
random-19-X-n1000 & 1,000 & 56 & 73,749.00 & 54,530.21 & 26.06 & 3,635.46 & 29 & 1,189 \\
random-13-X-n1000 & 1,000 & 72 & 92,888.00 & 67,860.94 & 26.94 & 3,768.42 & 26 & 1,253 \\
random-15-X-n1000 & 1,000 & 88 & 112,383.00 & 82,013.29 & 27.02 & 4,008.14 & 25 & 1,392 \\
random-11-X-n1000 & 1,000 & 96 & 102,522.00 & 75,200.74 & 26.65 & 3,779.82 & 22 & 1,373 \\
random-5-X-n1000 & 1,000 & 133 & 130,385.00 & 96,517.92 & 25.97 & 7,517.89 & 32 & 2,190 \\
random-1-X-n1000 & 1,000 & 81 & 86,706.00 & 66,588.91 & 23.20 & 7,513.26 & 33 & 1,707 \\
random-7-X-n1000 & 1,000 & 64 & 81,052.00 & 61,502.12 & 24.12 & 7,564.20 & 37 & 1,622 \\
random-3-X-n1000 & 1,000 & 65 & 76,005.00 & 57,601.44 & 24.21 & 7,481.94 & 36 & 1,725 \\
random-9-X-n1000 & 1,000 & 57 & 81,448.00 & 61,394.09 & 24.62 & 7,594.25 & 38 & 1,562 \\
random-21-X-n1000 & 1,000 & 81 & 109,492.00 & 81,691.39 & 25.39 & 7,310.10 & 32 & 1,712 \\
random-19-X-n1000 & 1,000 & 56 & 73,749.00 & 55,698.31 & 24.48 & 7,491.08 & 37 & 1,550 \\
random-13-X-n1000 & 1,000 & 72 & 92,888.00 & 69,573.90 & 25.10 & 7,722.20 & 34 & 1,735 \\
random-15-X-n1000 & 1,000 & 88 & 112,383.00 & 84,205.49 & 25.07 & 7,518.02 & 31 & 1,813 \\
random-11-X-n1000 & 1,000 & 96 & 102,522.00 & 77,323.38 & 24.58 & 7,314.50 & 29 & 1,934 \\
\bottomrule
\end{tabular}

%% file: ec_tables/random_time_neuralsep2e_small.tex
\centering
\begin{tabular}{lrrrrrrrr}
\toprule
Name & Size & K & Best Known & LB & Gap (\%) & Runtime & Iterations & \# of Cuts\\
\midrule
random-24-X-n50 & 50 & 3 & 6,686.00 & 6,288.00 & 5.95 & 2.57 & 33 & 52 \\
random-4-X-n50 & 50 & 4 & 7,684.00 & 7,431.92 & 3.28 & 2.36 & 31 & 66 \\
random-28-X-n50 & 50 & 7 & 11,557.00 & 10,831.67 & 6.28 & 6.38 & 64 & 241 \\
random-0-X-n50 & 50 & 8 & 12,735.00 & 12,024.87 & 5.58 & 7.53 & 71 & 304 \\
random-16-X-n50 & 50 & 5 & 10,190.00 & 9,958.11 & 2.28 & 6.61 & 69 & 188 \\
random-20-X-n50 & 50 & 6 & 10,126.00 & 9,758.86 & 3.63 & 5.62 & 50 & 181 \\
random-32-X-n50 & 50 & 4 & 7,483.00 & 7,257.33 & 3.02 & 4.09 & 44 & 99 \\
random-12-X-n50 & 50 & 4 & 8,317.00 & 8,038.67 & 3.35 & 2.60 & 31 & 76 \\
random-36-X-n50 & 50 & 8 & 14,188.00 & 13,641.54 & 3.85 & 11.04 & 94 & 411 \\
random-8-X-n50 & 50 & 3 & 6,733.00 & 6,566.35 & 2.48 & 1.60 & 20 & 40 \\
random-17-X-n75 & 75 & 20 & 34,431.00 & 32,644.62 & 5.19 & 38.92 & 133 & 1,289 \\
random-21-X-n75 & 75 & 7 & 9,726.00 & 9,354.55 & 3.82 & 6.15 & 43 & 198 \\
random-25-X-n75 & 75 & 6 & 10,931.00 & 10,243.72 & 6.29 & 7.42 & 53 & 198 \\
random-5-X-n75 & 75 & 11 & 14,930.00 & 13,938.33 & 6.64 & 17.19 & 86 & 518 \\
random-9-X-n75 & 75 & 5 & 9,361.00 & 8,933.50 & 4.57 & 5.53 & 42 & 134 \\
random-29-X-n75 & 75 & 7 & 11,893.00 & 11,270.54 & 5.23 & 12.04 & 81 & 312 \\
random-1-X-n75 & 75 & 7 & 11,225.00 & 10,675.95 & 4.89 & 8.57 & 61 & 245 \\
random-13-X-n75 & 75 & 6 & 9,652.00 & 9,400.40 & 2.61 & 5.44 & 41 & 164 \\
random-37-X-n75 & 75 & 4 & 9,479.00 & 8,992.71 & 5.13 & 3.94 & 35 & 99 \\
random-33-X-n75 & 75 & 13 & 16,151.00 & 15,193.53 & 5.93 & 23.43 & 115 & 781 \\
random-2-X-n100 & 100 & 11 & 17,242.00 & 15,867.95 & 7.97 & 26.31 & 95 & 712 \\
random-30-X-n100 & 100 & 6 & 12,151.00 & 11,321.35 & 6.83 & 12.05 & 64 & 243 \\
random-14-X-n100 & 100 & 12 & 16,934.00 & 15,543.66 & 8.21 & 35.81 & 121 & 846 \\
random-6-X-n100 & 100 & 14 & 24,832.00 & 23,543.04 & 5.19 & 55.96 & 166 & 1,130 \\
random-34-X-n100 & 100 & 8 & 10,752.00 & 10,409.78 & 3.18 & 16.19 & 77 & 376 \\
random-26-X-n100 & 100 & 11 & 15,654.00 & 14,563.08 & 6.97 & 34.62 & 125 & 768 \\
random-18-X-n100 & 100 & 7 & 11,171.00 & 10,650.92 & 4.66 & 10.07 & 51 & 257 \\
random-22-X-n100 & 100 & 20 & 27,022.00 & 25,213.89 & 6.69 & 76.59 & 160 & 1,614 \\
random-38-X-n100 & 100 & 11 & 13,766.00 & 12,856.43 & 6.61 & 28.97 & 108 & 616 \\
random-10-X-n100 & 100 & 12 & 17,727.00 & 16,558.34 & 6.59 & 36.40 & 116 & 773 \\
random-19-X-n200 & 200 & 12 & 16,102.00 & 15,075.63 & 6.37 & 69.64 & 96 & 604 \\
random-27-X-n200 & 200 & 13 & 21,609.00 & 19,677.80 & 8.94 & 100.17 & 100 & 938 \\
random-23-X-n200 & 200 & 23 & 26,556.00 & 24,272.93 & 8.60 & 338.33 & 207 & 2,240 \\
random-7-X-n200 & 200 & 13 & 17,182.00 & 16,050.56 & 6.59 & 110.32 & 135 & 1,043 \\
random-11-X-n200 & 200 & 20 & 29,495.00 & 26,889.40 & 8.83 & 374.99 & 251 & 2,078 \\
random-35-X-n200 & 200 & 15 & 24,854.00 & 22,774.92 & 8.37 & 252.06 & 217 & 1,716 \\
random-15-X-n200 & 200 & 18 & 23,121.00 & 21,117.63 & 8.66 & 279.34 & 206 & 1,914 \\
random-3-X-n200 & 200 & 13 & 19,644.00 & 17,956.89 & 8.59 & 113.54 & 123 & 1,017 \\
random-31-X-n200 & 200 & 21 & 31,684.00 & 29,084.01 & 8.21 & 544.90 & 311 & 2,696 \\
random-39-X-n200 & 200 & 11 & 18,271.00 & 16,834.35 & 7.86 & 90.41 & 113 & 847 \\
random-7-X-n300 & 300 & 19 & 27,269.00 & 24,519.35 & 10.08 & 918.57 & 263 & 2,414 \\
random-1-X-n300 & 300 & 25 & 30,181.00 & 27,030.94 & 10.44 & 1,019.92 & 232 & 2,833 \\
random-5-X-n300 & 300 & 40 & 53,305.00 & 48,802.66 & 8.45 & 3,506.25 & 341 & 4,996 \\
random-0-X-n300 & 300 & 46 & 55,614.00 & 50,400.71 & 9.37 & 3,457.13 & 303 & 5,536 \\
random-8-X-n300 & 300 & 15 & 20,636.00 & 18,917.23 & 8.33 & 332.16 & 177 & 1,345 \\
random-6-X-n300 & 300 & 40 & 42,079.00 & 37,810.70 & 10.14 & 2,857.30 & 314 & 4,791 \\
random-4-X-n300 & 300 & 20 & 24,324.00 & 22,029.90 & 9.43 & 684.42 & 257 & 1,910 \\
random-2-X-n300 & 300 & 32 & 47,765.00 & 43,571.40 & 8.78 & 2,880.56 & 368 & 4,455 \\
random-3-X-n300 & 300 & 20 & 23,414.00 & 21,304.76 & 9.01 & 508.07 & 180 & 1,925 \\
random-9-X-n300 & 300 & 17 & 21,740.00 & 20,244.45 & 6.88 & 395.26 & 187 & 1,597 \\
\bottomrule
\end{tabular}

%% file: ec_tables/random_time_neuralsep2e_large.tex
\centering
\begin{tabular}{lrrrrrrrr}
\toprule
Name & Size & K & Best Known & LB & Gap (\%) & Runtime & Iterations & \# of Cuts\\
\midrule
random-14-X-n400 & 400 & 46 & 61,276.00 & 54,538.27 & 11.00 & 3,633.19 & 114 & 3,508 \\
random-6-X-n400 & 400 & 54 & 52,025.00 & 46,289.52 & 11.02 & 3,606.14 & 135 & 4,143 \\
random-10-X-n400 & 400 & 47 & 59,487.00 & 53,105.74 & 10.73 & 3,603.74 & 117 & 3,656 \\
random-2-X-n400 & 400 & 42 & 61,447.00 & 54,433.59 & 11.41 & 3,611.85 & 125 & 3,601 \\
random-12-X-n400 & 400 & 30 & 37,545.00 & 33,101.34 & 11.84 & 3,603.27 & 274 & 3,766 \\
random-0-X-n400 & 400 & 61 & 77,967.00 & 69,673.06 & 10.64 & 3,634.80 & 91 & 4,143 \\
random-16-X-n400 & 400 & 38 & 57,273.00 & 50,819.14 & 11.27 & 3,633.02 & 134 & 3,401 \\
random-18-X-n400 & 400 & 27 & 34,643.00 & 30,545.17 & 11.83 & 3,086.48 & 304 & 3,268 \\
random-4-X-n400 & 400 & 27 & 29,917.00 & 26,892.13 & 10.11 & 2,252.79 & 312 & 3,130 \\
random-8-X-n400 & 400 & 20 & 30,953.00 & 27,218.19 & 12.07 & 1,540.34 & 216 & 2,033 \\
random-17-X-n500 & 500 & 128 & 134,831.00 & 114,578.06 & 15.02 & 3,728.10 & 41 & 3,820 \\
random-15-X-n500 & 500 & 44 & 48,692.00 & 41,942.59 & 13.86 & 3,633.33 & 88 & 2,644 \\
random-19-X-n500 & 500 & 28 & 34,861.00 & 30,706.80 & 11.92 & 3,608.39 & 211 & 2,695 \\
random-1-X-n500 & 500 & 41 & 40,287.00 & 35,398.72 & 12.13 & 3,611.33 & 134 & 2,943 \\
random-9-X-n500 & 500 & 29 & 43,607.00 & 37,556.51 & 13.88 & 3,629.45 & 115 & 2,264 \\
random-13-X-n500 & 500 & 36 & 46,414.00 & 40,295.21 & 13.18 & 3,619.64 & 106 & 2,419 \\
random-5-X-n500 & 500 & 67 & 85,332.00 & 72,362.65 & 15.20 & 3,638.86 & 55 & 2,829 \\
random-11-X-n500 & 500 & 48 & 66,720.00 & 58,031.99 & 13.02 & 3,601.99 & 73 & 2,442 \\
random-3-X-n500 & 500 & 33 & 56,591.00 & 49,257.37 & 12.96 & 3,633.60 & 95 & 2,240 \\
random-7-X-n500 & 500 & 32 & 48,159.00 & 41,573.00 & 13.68 & 3,638.52 & 101 & 2,280 \\
random-6-X-n750 & 750 & 99 & 137,429.00 & 105,705.76 & 23.08 & 3,842.81 & 30 & 1,962 \\
random-12-X-n750 & 750 & 56 & 81,988.00 & 65,108.58 & 20.59 & 3,696.07 & 36 & 1,556 \\
random-4-X-n750 & 750 & 49 & 55,938.00 & 45,572.16 & 18.53 & 3,605.72 & 39 & 1,630 \\
random-14-X-n750 & 750 & 87 & 98,971.00 & 78,352.07 & 20.83 & 3,743.59 & 29 & 1,865 \\
random-8-X-n750 & 750 & 37 & 54,145.00 & 43,384.13 & 19.87 & 3,753.33 & 47 & 1,406 \\
random-2-X-n750 & 750 & 79 & 76,471.00 & 61,593.58 & 19.45 & 3,831.01 & 32 & 2,026 \\
random-16-X-n750 & 750 & 72 & 77,111.00 & 62,427.84 & 19.04 & 3,839.28 & 32 & 1,862 \\
random-0-X-n750 & 750 & 115 & 137,798.00 & 107,588.88 & 21.92 & 3,610.66 & 27 & 1,983 \\
random-18-X-n750 & 750 & 50 & 85,348.00 & 68,615.07 & 19.61 & 3,689.73 & 40 & 1,592 \\
random-10-X-n750 & 750 & 88 & 119,393.00 & 91,731.67 & 23.17 & 3,835.80 & 30 & 1,832 \\
random-5-X-n1000 & 1,000 & 133 & 130,385.00 & 93,759.17 & 28.09 & 3,801.13 & 18 & 1,516 \\
random-1-X-n1000 & 1,000 & 81 & 86,706.00 & 64,835.82 & 25.22 & 3,731.27 & 22 & 1,348 \\
random-7-X-n1000 & 1,000 & 64 & 81,052.00 & 60,411.61 & 25.47 & 3,684.69 & 27 & 1,204 \\
random-3-X-n1000 & 1,000 & 65 & 76,005.00 & 56,497.40 & 25.67 & 3,641.02 & 27 & 1,314 \\
random-9-X-n1000 & 1,000 & 57 & 81,448.00 & 60,927.92 & 25.19 & 3,609.31 & 30 & 1,249 \\
random-21-X-n1000 & 1,000 & 81 & 109,492.00 & 80,771.45 & 26.23 & 3,759.21 & 24 & 1,402 \\
random-19-X-n1000 & 1,000 & 56 & 73,749.00 & 54,163.61 & 26.56 & 3,774.41 & 28 & 1,184 \\
random-13-X-n1000 & 1,000 & 72 & 92,888.00 & 67,756.03 & 27.06 & 3,675.69 & 25 & 1,303 \\
random-15-X-n1000 & 1,000 & 88 & 112,383.00 & 83,323.46 & 25.86 & 3,988.31 & 23 & 1,368 \\
random-11-X-n1000 & 1,000 & 96 & 102,522.00 & 75,791.49 & 26.07 & 4,022.64 & 21 & 1,461 \\
random-5-X-n1000 & 1,000 & 133 & 130,385.00 & 96,723.58 & 25.82 & 7,850.19 & 25 & 2,148 \\
random-1-X-n1000 & 1,000 & 81 & 86,706.00 & 66,328.35 & 23.50 & 7,723.16 & 30 & 1,850 \\
random-7-X-n1000 & 1,000 & 64 & 81,052.00 & 61,790.49 & 23.76 & 7,401.95 & 36 & 1,596 \\
random-3-X-n1000 & 1,000 & 65 & 76,005.00 & 57,599.14 & 24.22 & 7,446.73 & 37 & 1,768 \\
random-9-X-n1000 & 1,000 & 57 & 81,448.00 & 62,462.83 & 23.31 & 7,424.49 & 40 & 1,642 \\
random-21-X-n1000 & 1,000 & 81 & 109,492.00 & 82,605.98 & 24.56 & 7,205.57 & 31 & 1,837 \\
random-19-X-n1000 & 1,000 & 56 & 73,749.00 & 55,302.83 & 25.01 & 7,554.32 & 37 & 1,579 \\
random-13-X-n1000 & 1,000 & 72 & 92,888.00 & 69,034.67 & 25.68 & 7,309.52 & 33 & 1,695 \\
random-15-X-n1000 & 1,000 & 88 & 112,383.00 & 85,369.44 & 24.04 & 7,826.58 & 30 & 1,875 \\
random-11-X-n1000 & 1,000 & 96 & 102,522.00 & 77,539.65 & 24.37 & 7,657.37 & 27 & 1,956 \\
\bottomrule
\end{tabular}

%% file: ec_tables/random_time_neuralsep2e2_small.tex
\centering
\begin{tabular}{lrrrrrrrr}
\toprule
Name & Size & K & Best Known & LB & Gap (\%) & Runtime & Iterations & \# of Cuts\\
\midrule
random-24-X-n50 & 50 & 3 & 6,686.00 & 6,463.50 & 2.66 & 12.78 & 29 & 75 \\
random-4-X-n50 & 50 & 4 & 7,684.00 & 7,407.00 & 3.60 & 2.24 & 26 & 73 \\
random-28-X-n50 & 50 & 7 & 11,557.00 & 10,951.71 & 5.24 & 9.70 & 74 & 298 \\
random-0-X-n50 & 50 & 8 & 12,735.00 & 12,091.48 & 5.05 & 10.96 & 82 & 370 \\
random-16-X-n50 & 50 & 5 & 10,190.00 & 9,925.52 & 2.60 & 6.18 & 62 & 192 \\
random-20-X-n50 & 50 & 6 & 10,126.00 & 9,723.28 & 3.98 & 6.92 & 63 & 229 \\
random-32-X-n50 & 50 & 4 & 7,483.00 & 7,215.50 & 3.57 & 2.75 & 33 & 97 \\
random-12-X-n50 & 50 & 4 & 8,317.00 & 8,084.17 & 2.80 & 4.93 & 52 & 133 \\
random-36-X-n50 & 50 & 8 & 14,188.00 & 13,591.69 & 4.20 & 11.10 & 83 & 393 \\
random-8-X-n50 & 50 & 3 & 6,733.00 & 6,591.33 & 2.10 & 2.03 & 25 & 63 \\
random-17-X-n75 & 75 & 20 & 34,431.00 & 33,052.77 & 4.00 & 69.74 & 157 & 1,758 \\
random-21-X-n75 & 75 & 7 & 9,726.00 & 9,336.00 & 4.01 & 6.83 & 41 & 208 \\
random-25-X-n75 & 75 & 6 & 10,931.00 & 10,406.18 & 4.80 & 9.78 & 61 & 243 \\
random-5-X-n75 & 75 & 11 & 14,930.00 & 14,013.51 & 6.14 & 30.31 & 124 & 704 \\
random-9-X-n75 & 75 & 5 & 9,361.00 & 9,011.50 & 3.73 & 8.28 & 54 & 162 \\
random-29-X-n75 & 75 & 7 & 11,893.00 & 11,290.51 & 5.07 & 12.42 & 69 & 297 \\
random-1-X-n75 & 75 & 7 & 11,225.00 & 10,687.29 & 4.79 & 10.89 & 64 & 267 \\
random-13-X-n75 & 75 & 6 & 9,652.00 & 9,381.64 & 2.80 & 5.93 & 39 & 167 \\
random-37-X-n75 & 75 & 4 & 9,479.00 & 9,041.23 & 4.62 & 3.76 & 30 & 101 \\
random-33-X-n75 & 75 & 13 & 16,151.00 & 15,308.28 & 5.22 & 26.09 & 97 & 845 \\
random-2-X-n100 & 100 & 11 & 17,242.00 & 16,095.82 & 6.65 & 53.61 & 147 & 942 \\
random-30-X-n100 & 100 & 6 & 12,151.00 & 11,297.82 & 7.02 & 12.13 & 56 & 236 \\
random-14-X-n100 & 100 & 12 & 16,934.00 & 15,613.29 & 7.80 & 38.49 & 106 & 823 \\
random-6-X-n100 & 100 & 14 & 24,832.00 & 23,596.61 & 4.97 & 55.98 & 135 & 1,239 \\
random-34-X-n100 & 100 & 8 & 10,752.00 & 10,412.17 & 3.16 & 25.17 & 93 & 434 \\
random-26-X-n100 & 100 & 11 & 15,654.00 & 14,598.84 & 6.74 & 40.33 & 116 & 815 \\
random-18-X-n100 & 100 & 7 & 11,171.00 & 10,721.22 & 4.03 & 13.99 & 61 & 307 \\
random-22-X-n100 & 100 & 20 & 27,022.00 & 25,368.97 & 6.12 & 73.72 & 127 & 1,737 \\
random-38-X-n100 & 100 & 11 & 13,766.00 & 12,832.49 & 6.78 & 23.15 & 76 & 606 \\
random-10-X-n100 & 100 & 12 & 17,727.00 & 16,657.27 & 6.03 & 48.56 & 130 & 922 \\
random-19-X-n200 & 200 & 12 & 16,102.00 & 15,145.92 & 5.94 & 101.74 & 113 & 755 \\
random-27-X-n200 & 200 & 13 & 21,609.00 & 19,889.97 & 7.96 & 224.56 & 194 & 1,415 \\
random-23-X-n200 & 200 & 23 & 26,556.00 & 24,430.48 & 8.00 & 551.55 & 261 & 2,516 \\
random-7-X-n200 & 200 & 13 & 17,182.00 & 16,059.62 & 6.53 & 125.38 & 128 & 1,039 \\
random-11-X-n200 & 200 & 20 & 29,495.00 & 27,143.02 & 7.97 & 639.99 & 335 & 2,975 \\
random-35-X-n200 & 200 & 15 & 24,854.00 & 22,868.26 & 7.99 & 276.70 & 202 & 1,748 \\
random-15-X-n200 & 200 & 18 & 23,121.00 & 21,177.52 & 8.41 & 334.21 & 205 & 2,052 \\
random-3-X-n200 & 200 & 13 & 19,644.00 & 18,041.90 & 8.16 & 174.70 & 164 & 1,236 \\
random-31-X-n200 & 200 & 21 & 31,684.00 & 29,107.45 & 8.13 & 520.97 & 249 & 2,731 \\
random-39-X-n200 & 200 & 11 & 18,271.00 & 16,911.83 & 7.44 & 151.79 & 161 & 1,059 \\
random-7-X-n300 & 300 & 19 & 27,269.00 & 24,608.81 & 9.76 & 1,085.47 & 268 & 2,687 \\
random-1-X-n300 & 300 & 25 & 30,181.00 & 27,294.04 & 9.57 & 1,725.77 & 327 & 3,637 \\
random-5-X-n300 & 300 & 40 & 53,305.00 & 48,939.60 & 8.19 & 3,605.39 & 305 & 5,632 \\
random-0-X-n300 & 300 & 46 & 55,614.00 & 50,482.03 & 9.23 & 3,602.06 & 277 & 6,015 \\
random-8-X-n300 & 300 & 15 & 20,636.00 & 18,951.52 & 8.16 & 351.94 & 163 & 1,391 \\
random-6-X-n300 & 300 & 40 & 42,079.00 & 38,069.67 & 9.53 & 3,602.90 & 351 & 5,435 \\
random-4-X-n300 & 300 & 20 & 24,324.00 & 22,069.52 & 9.27 & 685.17 & 216 & 2,150 \\
random-2-X-n300 & 300 & 32 & 47,765.00 & 43,702.23 & 8.51 & 3,605.50 & 393 & 5,246 \\
random-3-X-n300 & 300 & 20 & 23,414.00 & 21,450.42 & 8.39 & 685.97 & 216 & 2,191 \\
random-9-X-n300 & 300 & 17 & 21,740.00 & 20,388.71 & 6.22 & 529.21 & 210 & 1,817 \\
\bottomrule
\end{tabular}

%% file: ec_tables/random_time_neuralsep2e2_large.tex
\centering
\begin{tabular}{lrrrrrrrr}
\toprule
Name & Size & K & Best Known & LB & Gap (\%) & Runtime & Iterations & \# of Cuts\\
\midrule
random-14-X-n400 & 400 & 46 & 61,276.00 & 54,591.38 & 10.91 & 3,603.85 & 110 & 3,862 \\
random-6-X-n400 & 400 & 54 & 52,025.00 & 46,617.45 & 10.39 & 3,604.90 & 127 & 4,771 \\
random-10-X-n400 & 400 & 47 & 59,487.00 & 53,255.39 & 10.48 & 3,600.95 & 108 & 3,985 \\
random-2-X-n400 & 400 & 42 & 61,447.00 & 54,640.37 & 11.08 & 3,646.73 & 116 & 3,728 \\
random-12-X-n400 & 400 & 30 & 37,545.00 & 33,214.02 & 11.54 & 3,604.00 & 231 & 3,850 \\
random-0-X-n400 & 400 & 61 & 77,967.00 & 69,871.38 & 10.38 & 3,612.65 & 85 & 4,498 \\
random-16-X-n400 & 400 & 38 & 57,273.00 & 50,847.51 & 11.22 & 3,617.65 & 130 & 3,660 \\
random-18-X-n400 & 400 & 27 & 34,643.00 & 30,632.67 & 11.58 & 3,604.72 & 300 & 3,810 \\
random-4-X-n400 & 400 & 27 & 29,917.00 & 27,178.92 & 9.15 & 3,154.05 & 405 & 3,545 \\
random-8-X-n400 & 400 & 20 & 30,953.00 & 27,370.94 & 11.57 & 2,389.24 & 304 & 2,531 \\
random-17-X-n500 & 500 & 128 & 134,831.00 & 114,718.75 & 14.92 & 3,779.42 & 41 & 3,880 \\
random-15-X-n500 & 500 & 44 & 48,692.00 & 42,093.28 & 13.55 & 3,622.27 & 83 & 2,931 \\
random-19-X-n500 & 500 & 28 & 34,861.00 & 30,698.08 & 11.94 & 3,607.78 & 190 & 2,812 \\
random-1-X-n500 & 500 & 41 & 40,287.00 & 35,658.69 & 11.49 & 3,624.73 & 125 & 3,289 \\
random-9-X-n500 & 500 & 29 & 43,607.00 & 37,490.60 & 14.03 & 3,603.03 & 111 & 2,378 \\
random-13-X-n500 & 500 & 36 & 46,414.00 & 40,257.21 & 13.26 & 3,619.38 & 98 & 2,541 \\
random-5-X-n500 & 500 & 67 & 85,332.00 & 72,712.80 & 14.79 & 3,656.28 & 54 & 2,967 \\
random-11-X-n500 & 500 & 48 & 66,720.00 & 57,985.69 & 13.09 & 3,625.08 & 71 & 2,626 \\
random-3-X-n500 & 500 & 33 & 56,591.00 & 49,307.88 & 12.87 & 3,612.99 & 92 & 2,325 \\
random-7-X-n500 & 500 & 32 & 48,159.00 & 41,632.57 & 13.55 & 3,615.19 & 100 & 2,427 \\
random-6-X-n750 & 750 & 99 & 137,429.00 & 105,342.42 & 23.35 & 3,943.46 & 31 & 1,977 \\
random-12-X-n750 & 750 & 56 & 81,988.00 & 64,990.94 & 20.73 & 3,845.55 & 36 & 1,593 \\
random-4-X-n750 & 750 & 49 & 55,938.00 & 45,386.83 & 18.86 & 3,648.96 & 39 & 1,723 \\
random-14-X-n750 & 750 & 87 & 98,971.00 & 77,186.53 & 22.01 & 3,743.50 & 29 & 1,857 \\
random-8-X-n750 & 750 & 37 & 54,145.00 & 43,528.83 & 19.61 & 3,709.97 & 47 & 1,486 \\
random-2-X-n750 & 750 & 79 & 76,471.00 & 61,710.01 & 19.30 & 3,772.68 & 31 & 1,984 \\
random-16-X-n750 & 750 & 72 & 77,111.00 & 62,301.55 & 19.21 & 3,805.07 & 32 & 1,905 \\
random-0-X-n750 & 750 & 115 & 137,798.00 & 108,003.49 & 21.62 & 3,909.81 & 28 & 2,057 \\
random-18-X-n750 & 750 & 50 & 85,348.00 & 68,620.30 & 19.60 & 3,765.12 & 40 & 1,628 \\
random-10-X-n750 & 750 & 88 & 119,393.00 & 91,793.50 & 23.12 & 3,679.89 & 30 & 1,857 \\
random-5-X-n1000 & 1,000 & 133 & 130,385.00 & 93,759.17 & 28.09 & 3,801.07 & 18 & 1,516 \\
random-1-X-n1000 & 1,000 & 81 & 86,706.00 & 64,782.58 & 25.28 & 3,867.06 & 22 & 1,405 \\
random-7-X-n1000 & 1,000 & 64 & 81,052.00 & 60,848.69 & 24.93 & 3,860.57 & 27 & 1,271 \\
random-3-X-n1000 & 1,000 & 65 & 76,005.00 & 56,644.43 & 25.47 & 3,843.28 & 27 & 1,370 \\
random-9-X-n1000 & 1,000 & 57 & 81,448.00 & 60,981.71 & 25.13 & 3,650.67 & 30 & 1,260 \\
random-21-X-n1000 & 1,000 & 81 & 109,492.00 & 80,685.10 & 26.31 & 3,770.24 & 24 & 1,490 \\
random-19-X-n1000 & 1,000 & 56 & 73,749.00 & 54,127.73 & 26.61 & 3,722.94 & 28 & 1,219 \\
random-13-X-n1000 & 1,000 & 72 & 92,888.00 & 68,128.94 & 26.65 & 3,786.03 & 25 & 1,355 \\
random-15-X-n1000 & 1,000 & 88 & 112,383.00 & 83,206.64 & 25.96 & 3,957.98 & 23 & 1,381 \\
random-11-X-n1000 & 1,000 & 96 & 102,522.00 & 75,815.33 & 26.05 & 4,065.41 & 21 & 1,472 \\
random-5-X-n1000 & 1,000 & 133 & 130,385.00 & 96,734.46 & 25.81 & 7,248.86 & 24 & 2,091 \\
random-1-X-n1000 & 1,000 & 81 & 86,706.00 & 66,856.58 & 22.89 & 7,661.90 & 29 & 1,885 \\
random-7-X-n1000 & 1,000 & 64 & 81,052.00 & 61,985.77 & 23.52 & 7,519.69 & 36 & 1,681 \\
random-3-X-n1000 & 1,000 & 65 & 76,005.00 & 57,820.54 & 23.93 & 7,270.84 & 35 & 1,796 \\
random-9-X-n1000 & 1,000 & 57 & 81,448.00 & 62,161.31 & 23.68 & 7,324.42 & 40 & 1,667 \\
random-21-X-n1000 & 1,000 & 81 & 109,492.00 & 82,908.21 & 24.28 & 7,735.73 & 32 & 1,965 \\
random-19-X-n1000 & 1,000 & 56 & 73,749.00 & 55,632.15 & 24.57 & 7,547.35 & 37 & 1,642 \\
random-13-X-n1000 & 1,000 & 72 & 92,888.00 & 69,521.63 & 25.16 & 7,458.25 & 33 & 1,809 \\
random-15-X-n1000 & 1,000 & 88 & 112,383.00 & 84,922.28 & 24.43 & 7,335.74 & 29 & 1,863 \\
random-11-X-n1000 & 1,000 & 96 & 102,522.00 & 77,929.92 & 23.99 & 7,928.46 & 27 & 1,979 \\
\bottomrule
\end{tabular}